\documentclass{article}

\PassOptionsToPackage{super,sort&compress}{natbib}
\usepackage{manuscript}

\usepackage[utf8]{inputenc} 
\usepackage[T1]{fontenc}    
\usepackage{hyperref}       
\usepackage{url}            
\usepackage{booktabs}       
\usepackage{amsfonts}       
\usepackage{nicefrac}       
\usepackage{microtype}      
\usepackage{xcolor}         
\usepackage{csquotes}
\usepackage[colorinlistoftodos]{todonotes}

\usepackage{amsmath, amssymb, amsthm}
\usepackage{bm}
\usepackage{booktabs}
\usepackage{multirow}
\usepackage{array}
\usepackage{graphicx}
\usepackage{subcaption}
\usepackage{enumitem}
\usepackage{hyperref}
\usepackage{url}
\usepackage{float}
\usepackage{adjustbox}
\usepackage{makecell}
\usepackage{tabularx}
\usepackage{threeparttable}
\usepackage{caption}
\usepackage{wrapfig}
\usepackage[ruled,vlined,linesnumbered]{algorithm2e}
\usepackage{siunitx}
\usepackage{soul}
\usepackage{dsfont}
\usepackage{pifont}
\usepackage{lineno}
\linenumbers

\newcommand{\cmark}{\ding{51}}
\newcommand{\xmark}{\ding{55}}

\usepackage{xstring}
\protected\def\suppmarker{}
\newif\ifsuppref
\newcommand{\checksuppref}[1]{%
  \global\suppreffalse
  \begingroup
    \protected\def\suppmarker{\global\suppreftrue}%
    \setbox0\hbox{\ref*{#1}}%
  \endgroup
}
\newcommand{\suppref}[2]{\hyperref[#2]{#1~\ref*{#2}}}
\NewCommandCopy{\originalautoref}{\autoref}
\DeclareRobustCommand{\autoref}[1]{%
  \checksuppref{#1}%
  \IfBeginWith{#1}{app:}%
    {\suppref{Supplementary Section}{#1}}%
    {\ifsuppref
       \IfBeginWith{#1}{fig:}{\suppref{Supplementary Fig.}{#1}}%
       {\suppref{Supplementary Table}{#1}}%
     \else\originalautoref{#1}\fi}%
}

\makeatletter
\newif\ifinsupp
\NewCommandCopy{\originalappendix}{\appendix}
\renewcommand{\appendix}{%
  \ifinsupp\else
    \global\insupptrue
    \originalappendix
    \renewcommand{\thesection}{\arabic{section}}%
    \setcounter{figure}{0}%
    \setcounter{table}{0}%
    \renewcommand{\figurename}{Supplementary Fig.}%
    \renewcommand{\tablename}{Supplementary Table}%
    \renewcommand{\p@figure}{\suppmarker}%
    \renewcommand{\p@table}{\suppmarker}%
    \renewcommand{\theHfigure}{supp.\arabic{figure}}%
    \renewcommand{\theHtable}{supp.\arabic{table}}%
    {\noindent\LARGE\bfseries Supplementary Information\par}\vskip 1em
  \fi
}
\makeatother

\newcommand{\Xul}[1]{{\setul{1pt}{.7pt}\ul{#1}}}

\hypersetup{
    colorlinks=true,
    linkcolor=blue,
    citecolor=blue,
    urlcolor=blue
}

\title{ProtoSSL: Self-Supervised Pretraining and Downstream Transfer for Projection-Based Prototype Models}

\author{%
Steven Song\textsuperscript{*,1,2,3,4}, Sahil Sethi\textsuperscript{*,4,5},
\textbf{Brett Beaulieu-Jones\textsuperscript{\textdagger,5,6}, Robert L. Grossman\textsuperscript{\textdagger,1,2,6}}\\
\textsuperscript{1} Department of Computer Science, University of Chicago\\
\textsuperscript{2} Center for Translational Data Science, University of Chicago\\
\textsuperscript{3} Medical Scientist Training Program, University of Chicago\\
\textsuperscript{4} Pritzker School of Medicine, University of Chicago\\
\textsuperscript{5} Center for Computational Medicine \& Clinical AI, University of Chicago\\
\textsuperscript{6} Section of Biomedical Data Science, Department of Medicine, University of Chicago\\
\textsuperscript{*,\textdagger} Equal Contribution.\\
Correspondence: \texttt{\{beaulieujones, rgrossman1\}@uchicago.edu}\\
}

\begin{document}

\nolinenumbers
\maketitle

\begin{abstract}
In domains where both predictive performance and interpretability are essential, deep neural networks achieve strong results but provide limited insight into how their predictions are made. Projection-based prototype networks address this limitation by grounding predictions in similarity to representative training examples, enabling case-based explanations and global prototype inspection. However, existing approaches rely on label supervision, tying prototypes to a specific task and requiring large labeled datasets. We introduce \textbf{ProtoSSL}, a framework for pretraining a foundational latent prototype bank on unlabeled data and transferring it to downstream tasks to create interpretable, projection-based prototype models. Our key idea is to separate \emph{motif discovery} from \emph{label alignment}. ProtoSSL first learns a transferrable prototype bank using a self-supervised objective applied directly to prototype activations, and then aligns these prototypes to downstream tasks through a novel efficient assignment procedure. Across six electrocardiography (ECG) datasets, ProtoSSL improves label efficiency, outperforming supervised prototype baselines in low-data regimes with as few as 256 labeled examples; with fine-tuning, ProtoSSL outperforms supervised prototype baselines at full dataset scale. In a human evaluation study, ProtoSSL produces prototypes and prototype-based explanations that are judged more favorably than those learned with direct label supervision. We further show that the framework extends to audio classification. Thus, ProtoSSL enables both learning foundational prototypes from unlabeled data before the downstream label space is known, and subsequent assignment to new tasks to create interpretable, projection-grounded prototypes.

\end{abstract}

\section{Introduction}
Deep learning models are increasingly being developed for application domains where both predictive performance and interpretability are essential, particularly across medical modalities \citep{sethi_protoecgnet_2025, barnett_improving_2024, barnett_case-based_2021, kim_xprotonet_2021, rymarczyk_protomil_2022, sun_prototype-based_2025}. While foundation models have recently achieved strong generalizable performance in these domains \citep{al-masud_benchmarking_2026, neidlinger_benchmarking_2026}, the representations that drive their predictions are difficult to inspect, and post hoc attribution methods may fail to faithfully reflect the model’s actual decision process \citep{rudin_stop_2019, barnett_case-based_2021, zhang_overlooked_2022, bai_why_2021}.

\begin{figure*}[t]
    \centering
    \includegraphics[width=0.95\textwidth]{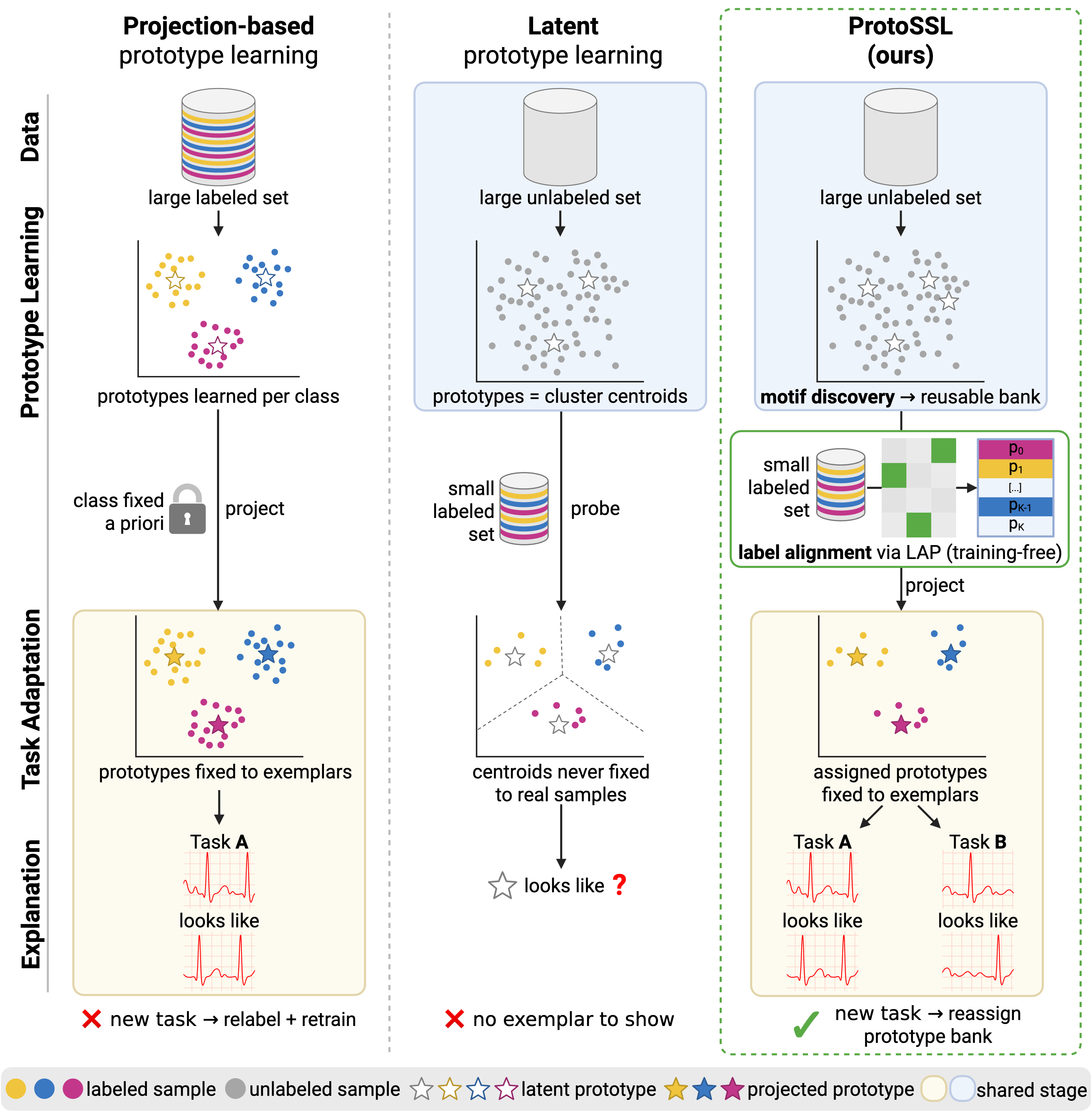}
    \caption{\textbf{ProtoSSL uniquely combines self-supervised pretraining, projection-grounded interpretability, and cross-task transfer}, a combination not achieved by prior prototype or self-supervised methods. Projection-based prototype models learn prototypes through label supervision for a single task; prototypes must therefore be relearned for new tasks, each time requiring a large, labeled dataset. Models pretrained once via self-supervision over a large, unlabeled dataset are adapted cheaply to diverse downstream tasks, but their explanations are not intrinsic to model predictions. ProtoSSL leverages the best of both paradigms to achieve pretraining and transfer of interpretable, case-based prototype models.
    }
    \label{fig:protossl_comparison}
\end{figure*}

Projection-based prototypical part neural networks offer a compelling alternative by building case-based reasoning directly into the model. In the \textit{ProtoPNet} framework \citep{chen_this_2019} and its extensions \citep{hase_interpretable_2019, wang_interpretable_2021, rymarczyk_interpretable_2022, ma_this_2023}, predictions are made by comparing inputs to a learned set of prototypes representing characteristic patterns. A defining feature of this family of methods is an explicit projection step that anchors each prototype to portions of real training examples, enabling explanations grounded in concrete, inspectable segments. This design yields intrinsically interpretable models that justify predictions via \enquote{this looks like that} reasoning. Despite these advantages, existing prototype learning approaches are fundamentally \emph{label-bound}. Prototypes are learned through label-supervised objectives that explicitly enforce alignment with class structure and tie the learned representations to a specific task \citep{de_santi_part-prototype_2024}. This introduces two key limitations: (1) prototype learning depends on large labeled datasets, which are often scarce in domains where interpretability is most critical (such as medicine \citep{carloni_applicability_2023}), and (2) learned prototypes do not readily transfer across tasks, as their semantics are defined by the original supervised objective. As a result, prototypes must be relearned when tasks, datasets, or labeling schemes change.

Self-supervised learning (SSL) addresses the complementary challenge of learning transferable structure from large amounts of unlabeled data. Methods such as SimCLR \citep{chen_simple_2020} or the DINO family \citep{oquab_dinov2_2024} learn foundational representations that transfer effectively across downstream tasks. However, these approaches operate in latent space and do not produce ante-hoc explanations for model decisions. Even when \enquote{prototypes} are introduced in self-supervised settings, they typically correspond to latent centroids rather than projected, human-inspectable exemplars \citep{caron_unsupervised_2020, li_prototypical_2020, du_prototype-guided_2023}. This gap motivates our central question: \ul{how can self-supervised methods be used to pretrain prototypes that transfer across tasks to create inherently interpretable projection-based prototype models}?

\begin{figure}[tp]
  \centering
  \includegraphics[width=0.95\textwidth]{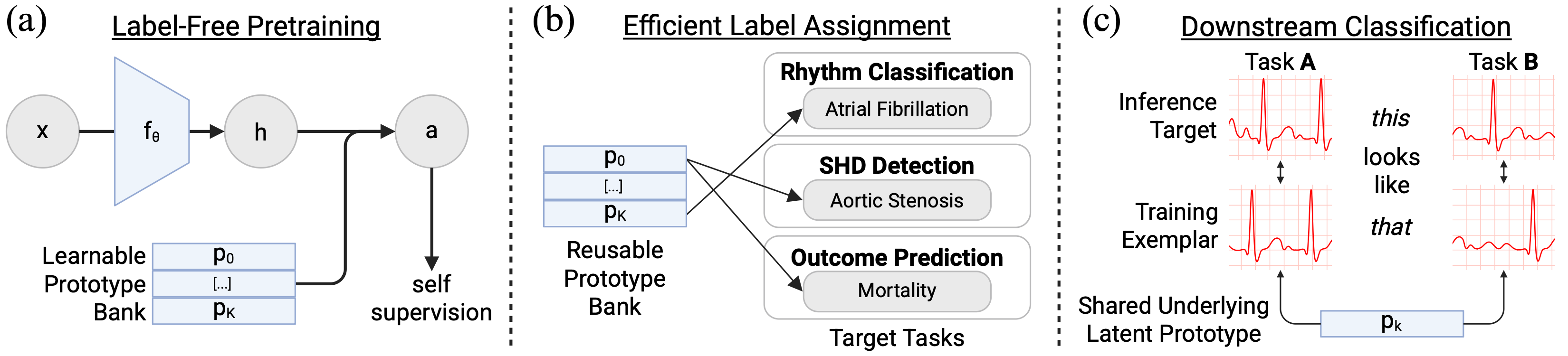}
  \caption{\textbf{Conceptual overview of ProtoSSL}, a novel framework for (a) learning reusable, foundational latent prototypes not tied to any specific task and (b) efficient, training-free label-assignment to downstream datasets, ultimately (c) enabling interpretable, case-based \textit{this} looks like \textit{that} classification across multiple domains.}
  \label{fig:conceptual-overview}
\end{figure}

We introduce \textbf{ProtoSSL}, a framework for learning foundational prototypes from unlabeled data and enabling projection-based transfer across downstream tasks (\autoref{fig:protossl_comparison}). \ul{Our central idea is to separate \emph{motif discovery} from \emph{label alignment}} (\autoref{fig:conceptual-overview}). We first learn a common latent prototype bank from unlabeled data using a self-supervised objective applied directly to prototype activations, forcing the learned structure to be expressed in prototype space. We then align these prototypes to downstream tasks using a novel training-free assignment procedure and ground them through label-supervised projection. This decomposition enables latent prototypes to be reused across datasets and label spaces without retraining the prototype bank.

\paragraph{\ul{Our key contributions are as follows}:}
\begin{itemize}[leftmargin=*]
    \item We propose \textbf{ProtoSSL}, a novel framework for projection-grounded transfer of foundational prototypes learned over unlabeled data. Using self-supervision in prototype-activation space and then assigning prototypes to downstream labels, ProtoSSL separates \emph{motif discovery} from \emph{label alignment} (\autoref{sec:proto_ssl_method}).
    \item We empirically demonstrate that ProtoSSL's separation of motif discovery and label alignment enables prototype transfer across tasks. Across six ECG datasets, ProtoSSL achieves strong full-data performance (\autoref{sec:generalization}) and substantially improves label efficiency over supervised prototype baselines, remaining effective with as few as 256 labeled examples (\autoref{sec:label_efficiency}). Complementary experiments across three audio datasets show that the framework extends beyond ECG (\autoref{sec:audio_results}). In \autoref{sec:ablations_main}, we ablate key components of ProtoSSL to better understand the framework.
    \item We confirm that ProtoSSL's prototypes, when learned without task labels yet still grounded in real exemplars, remain interpretable in a human evaluation study. ProtoSSL-derived explanations are rated higher than directly label-supervised prototype explanations, suggesting that self-supervised motif discovery can yield more representative case-based evidence than end-to-end supervised prototype learning (\autoref{sec:user_study_results}).
\end{itemize}

\section{Background}
\label{sec:related_main}

\begin{table}[tp]
\caption{\textbf{Conceptual comparison of deep learning paradigms.} ProtoSSL uniquely enables pretraining and transfer of interpretable, case-based prototype models.}
\label{tab:conceptual_comparison}
\centering
\begin{adjustbox}{max width=\textwidth}

\begin{tabular}{lcccc}
\toprule
Paradigm & \makecell{Self-supervised\\pretraining} & \makecell{Prototype-based\\prediction} & \makecell{Case-based\\interpretability} & \makecell{Transferrable\\across tasks} \\
\midrule

Contrastive SSL (e.g. SimCLR) \citep{chen_simple_2020} 
& \cmark & \xmark & \xmark & \cmark \\

Clustering-based SSL (e.g. SwAV) \citep{caron_unsupervised_2020} 
& \cmark & \xmark & \xmark & \cmark \\

Latent prototype classifiers \citep{yang_robust_2018, pan_transferrable_2019}
& \xmark & \cmark & \xmark & \xmark \\

ProtoPNet-style models \citep{chen_this_2019, barnett_improving_2024, sethi_protoecgnet_2025} 
& \xmark & \cmark & \cmark & \xmark \\

\textbf{ProtoSSL (ours)} 
& \cmark & \cmark & \cmark & \cmark \\
\bottomrule
\end{tabular}

\end{adjustbox}
\end{table}

\paragraph{Projection-based prototype part learning.}
Projection-based prototypical part networks achieve this case-based form of interpretability by grounding predictions in representative, human-inspectable segments \citep{chen_this_2019, hase_interpretable_2019, wang_interpretable_2021, rymarczyk_interpretable_2022, ma_this_2023}. In these models, prototypes correspond to real examples, or localized regions thereof, and predictions are computed directly from similarities to these input-grounded prototypes. Thus, the same objects that support prediction are also presented as explanations, yielding ante-hoc, case-based reasoning that is inherently faithful to the model's decision process \citep{li_overview_2026}. Recent work has demonstrated the feasibility of this approach in domains such as ECG, EEG, and audio \citep{sethi_protoecgnet_2025, sethi_prototype_2025, barnett_improving_2024, heinrich_audioprotopnet_2025}. However, existing projection-based prototype models are typically learned with label supervision, optimizing their prototypes for a fixed downstream task and label space. This task dependence limits their reuse when downstream tasks change or when labeled data are scarce. Despite the practical cost of obtaining labeled data in many real-world settings, limited-label projection-based prototype learning remains underexplored; to the best of our knowledge, only \citet{santos_protoal_2024} investigate prototype part networks in a limited-label regime, using active learning (though we note this is primarily because the author's dataset of interest is small at approximately 1.2K samples; in our study, we instead systematically evaluate label-efficiency by constructing subsets of varying sizes across tasks down to just 256 samples).


\paragraph{Latent prototype learning.}
A separate line of work introduces prototype-like structure in representation space without using prototypes as input-grounded explanations. In self-supervised settings, clustering-based approaches assign samples to learned centroids called prototypes, encouraging the representation to organize around a finite set of recurring patterns \citep{caron_unsupervised_2020, arteaga_why_2026}. Related ideas have also been explored in supervised settings, where class prototypes are represented as centroids in embedding space and used directly for prediction, often improving robustness under distribution shift \citep{yang_robust_2018, pan_transferrable_2019}. In both settings, \enquote{prototypes} function as latent constructs that structure the embedding space. They are not grounded in real input examples and are not intended to provide case-based explanations to the user. While \citet{pan_transferrable_2019} explicitly study transfer of latent prototype representations between domains, they do so when the pretraining data is labeled and the target data is unlabeled. This differs from our setting, where latent prototypes are learned without task labels during self-supervised pretraining, then aligned to labeled downstream tasks to present projection-grounded, interpretable evidence.

\paragraph{Self-supervision within projection-based prototype models.}
To our knowledge, only \citet{nauta_pip-net_2023} have incorporated self-supervised learning into an interpretable projection-based prototype model. Their model, PIP-Net, used self-supervised learning to improve the semantic quality of learned prototypes \citep{nauta_pip-net_2023}. However, PIP-Net remains tied to a single downstream classification task: prototypes are learned and used within one label space, and no mechanism is provided for transferring or reassigning prototypes across tasks.

\paragraph{Prototype assignment within projection-based prototype models.}
Perhaps the closest related works to ours are \citet{rymarczyk_interpretable_2022} and \citet{zhu_interpretable_2025}, both of which develop prototype-label assignment methods for when prototype-to-label mappings are unknown a priori. 

ProtoPool \citep{rymarczyk_interpretable_2022} learns a shared pool of prototypes that can be softly assigned to multiple classes within a \emph{single supervised task}. The encoder, prototypes, and assignment weights are optimized jointly using labeled data, and the assignment mechanism is designed to improve parameter efficiency within the fixed label space. As a result, prototypes remain intrinsically tied to the training task, and no mechanism is provided for transferring a learned prototype bank across datasets or reassigning it under a new label space. One user study also showed that ProtoPool's reuse of prototypes across classes led to significantly worse interpretability than standard ProtoPNet \citep{davoodi_interpretability_2023}. Even when adapted to our setting to learn assignments for frozen, pretrained prototypes, this gradient-based method yields similar performance but is computationally inefficient compared to ProtoSSL's nonparametric assignment (Supplementary Tables \ref{tab:ecg_alt_assign}--\ref{tab:assignment_runtime}).

\citet{zhu_interpretable_2025} discover prototypes by clustering latent patch features from pretrained vision transformers separately for each class. Distinct from our objective of learning a reusable prototype space, their goal is to \emph{extract} and organize task-specific prototypes from a pretrained embedding space in which part-level structure is already well-represented \citep{oquab_dinov2_2024, caron_emerging_2021}. We note several limitations of this method: 1) it depends on a strong, pretrained embedding model, 2) the derived prototypes are cluster means of sample embeddings, but are not projected back onto real training samples, and 3) the clustering algorithm requires the number of patches belonging to a class be much greater than the number of prototype slots to discover per class (see \autoref{app:protos_from_fms}). When adapted to our setting, even when paired with a strong pretrained ECG encoder \citep{li_electrocardiogram_2025}, the method performs poorly and fails to converge at low dataset scales (see Supplementary Tables \ref{tab:ecg_protos_from_fms}--\ref{tab:ecg_sk_ot_converge}).

\paragraph{Positioning of this work.}
Prior work falls primarily into two regimes: (i) projection-based prototype models that provide interpretable, case-based reasoning but in a single supervised task, and (ii) self-supervised representation learning, including clustering-based prototype methods, that learns transferable latent structure but lacks projection-grounded interpretability.

\ul{ProtoSSL lies at their intersection} (summarized conceptually in \autoref{fig:protossl_comparison} and \autoref{tab:conceptual_comparison}). It learns foundational, latent prototypes from unlabeled data and enables their reuse across downstream tasks through an explicit assignment and projection procedure that anchors them to real training examples. \ul{This combination of self-supervised learning, projection-based interpretability, and cross-task transfer is not achieved by prior prototype or self-supervised methods}.

\section{ProtoSSL Framework}
\label{sec:proto_ssl_method}

\begin{figure*}[tp]
    \centering
    \includegraphics[width=0.95\textwidth]{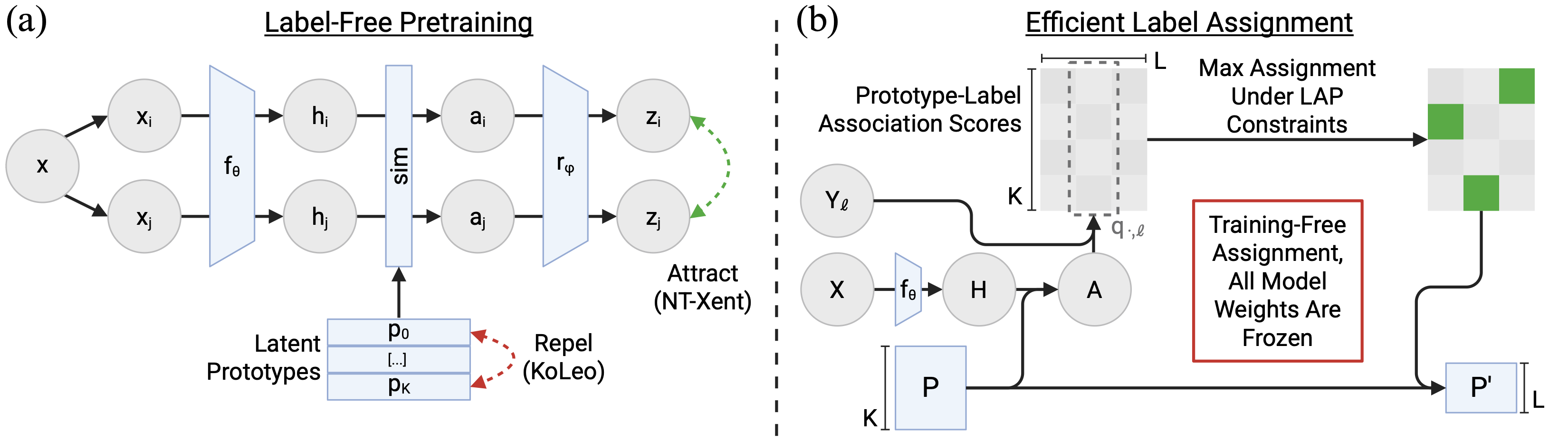}
    \caption{\textbf{ProtoSSL (a) learns a reusable latent prototype bank from unlabeled data, then (b) aligns and grounds it for downstream tasks via a training-free label assignment step that is solvable in polynomial time by reduction to the linear assignment problem (LAP).}
    Panels (a) and (b) illustrate the specific implementation of \autoref{fig:conceptual-overview}a and \ref{fig:conceptual-overview}b, respectively.
    }
    \label{fig:method_overview}
\end{figure*}

Classification tasks are generally accomplished via recognition of patterns that recur across examples of each class.
Existing supervised prototype models discover such patterns for labels defined a priori, both requiring labeled data and tying the initial latent prototypes to a specific task. ProtoSSL separates two roles that are traditionally entangled: \emph{motif discovery} and \emph{label alignment}. Motif discovery learns a foundational latent bank of recurring patterns from unlabeled data, and label alignment determines which of these patterns are relevant for a downstream task. This decomposition is central to our approach: \ul{motif discovery alone yields transferable structure, but task-aware alignment and label-supervised projection are necessary for interpretability}. Pseudocode of our framework is provided in \autoref{alg:pseudocode}; extended methods are detailed in \autoref{app:methods}.


\subsection{Learning a latent prototype bank from unlabeled data}
\label{sec:ssl_main}

We use a self-supervised contrastive objective to first learn a foundational latent prototype bank over unlabeled data. Let \(x \in \mathcal{X}\) denote an unlabeled input sample, \(h = f_\theta(x) \in \mathbb{R}^D\) an encoder representation with \(D\) dimensions, and \(P = {[p_k]_{k=1}^{K}}^\top \in \mathbb{R}^{K,D}\) the matrix of learnable prototypes. We define:
\[
a(x) = [a_k(x)]_{k=1}^K;~~~ a_k(x) = \text{sim}(f_\theta(x), p_k)
\]

\newcommand{\wrapul}[1]{\leavevmode\ul{#1}}
\begin{wrapfigure}[20]{r}{0.5\textwidth}
\vspace{5pt}
\setlength{\algomargin}{12pt} 
\begin{algorithm}[H]
\footnotesize
\caption{ProtoSSL Pseudocode}
\label{alg:pseudocode}
\SetKwBlock{KwPhase}{Phase:}{End}
\SetKwBlock{KwBlankBlock}{}{}
\KwPhase(Self-Supervised Pretraining){
    \KwIn{Unlabeled pretraining data \(\hat{\mathcal{X}}\)}
    \KwOut{Encoder \(f_\theta\), prototype pool \(P\)}
    \BlankLine
    \(\theta, P \gets \arg\underset{\theta,P}{\min}~\mathcal{L}_{\text{self-sup}}(\text{sim}(f_\theta(\hat{\mathcal{X}}),~P))\)\;
}
\BlankLine
\KwPhase(Target Adaptation){
    \vspace{2pt}
    \KwIn{Labeled target data \(\mathcal{X}, Y\)}
    \KwIn{Encoder \(f_\theta\), prototype pool \(P\)}
    \KwOut{Classifier \(g_\phi\)}
    \BlankLine
    \(P' \gets \textsc{AssignPrototypes}(P,~f_\theta(\mathcal{X}),~Y)\)\;
    \BlankLine
    \If{fine-tuning}{
        \mbox{\(\theta, P' \gets \arg\underset{\theta,P'}{\min}~\mathcal{L}_{\text{label-sup}}(f_\theta(\mathcal{X}),P',Y)\)}\;
        \vspace{-10pt}
    }
    \BlankLine
    \KwBlankBlock(\makebox[0pt][l]{\(P'' \gets \textsc{ProjectPrototypes}(P',~f_\theta(\mathcal{X}),~Y)\);}){
        \For{$p'_l \in P'$}{
            \vspace{1pt}
            $p''_l \gets \underset{h \in f_\theta(\mathcal{X}_l)}{\arg\max}~\text{sim}(h,~p'_l)$\;
        }
    }
    \vspace{-1pt}
    \mbox{\(\phi \gets \arg\underset{\phi}{\min}~\mathcal{L}_{\text{CE}}(g_\phi(\text{sim}(f_\theta(\mathcal{X}),~P'')),~Y)\)}\;
}
\end{algorithm}
\end{wrapfigure}

\noindent where \(a(x)\) is the activation vector of all prototypes in \(P\) given \(x\). For clarity of exposition, we present the activations as similarities over a single embedding of sample \(x\); in practice, we allow each \(a_k(x)\) to be derived over either global or local embeddings, following standard partial-prototype formulations \citep{chen_this_2019}. We detail the domain specific implementations of this in Supplementary Sections \ref{app:ecg_details} and \ref{app:audio_details}. ProtoSSL applies a SimCLR contrastive objective \citep{chen_simple_2020} to \(z = r_\varphi (a(x))\), where \(r_\varphi\) is a nonlinear projection head. Specifically, the projection head consists of a two-layer MLP with a ReLU activation in between, mapping from the prototype-activation space to the contrastive-projection space: Linear(\(K\)) \textrightarrow~ ReLU \textrightarrow~ Linear(\(K/2\)), where \(K\) is the prototype activation dimension (i.e., the number of prototypes) and the projection dimension is \(K/2\). \textbf{NB:} projection here refers to the contrastive learning context and \textit{not} the prototype learning context (i.e. the grounding of latent prototypes onto real exemplar samples) used elsewhere in this work.

We thus compute the NT-Xent contrastive loss \citep{chen_simple_2020} over the projected prototype activations. As the contrasted representations \(z\) are computed from the prototype activations, the model learns to represent \(x\) as an input-dependent weighting of prototypes. \ul{We refer to this process as prototype-adaptive self-selection, where all prototypes are optimized with respect to the full dataset without any label supervision}. The standard NT-Xent loss over projected prototype activations is given as follows:
\[
\mathcal{L}_{\text{NT-Xent}} = \frac{1}{2N} \sum_{n=1}^{N} \left[ \ell_{2n-1,\, 2n} + \ell_{2n,\, 2n-1} \right];~~~\ell_{i,j} = -\log \frac{\exp(\text{sim}(z_i, z_j)/\tau)}{\sum_{n \neq i} \exp(\text{sim}(z_i, z_n)/\tau)}
\]
where \(N\) is the number of samples in the batch and \((2n-1, 2n)\) denotes the positive pair formed by two views of the \(n\)th sample. To keep prototypes spread out in the latent space, we additionally apply KoLeo regularization \citep{sablayrolles_spreading_2018} to the prototype vectors in \(P\). The KoLeo loss term is given as follows:
\[
\mathcal{L}_{\text{KoLeo}} = -\frac{1}{K} \sum_{k=1}^{K} \log \min_{i \neq k} \| \hat{p}_k - \hat{p}_i \|_2;~~~\hat{p} = p / \|p\|_2
\]
The final loss for ProtoSSL's joint self-supervised pretraining of prototypes and encoder is given by:
\[
\mathcal{L}_{\text{ProtoSSL}} = \mathcal{L}_{\text{NT-Xent}} + \mathcal{L}_{\text{KoLeo}}
\]

\ul{ProtoSSL is adaptable across multiple architectures and leverages domain-specific contrastive pretraining strategies without modifying our core framework}. In ECG, we use PCLR \citep{diamant_patient_2022}, an adaptation of SimCLR that uses multiple ECGs from the same patient as the alternate views. In audio, we adopt the framework from \citet{saeed_contrastive_2020} and construct views by sampling two independent segments from the same waveform; we further do standard augmentations for contrastive audio training \citep{al-tahan_clar_2021}.

Self-supervision recovers recurring structure in data that can support downstream tasks without labels \citep{chen_simple_2020, caron_emerging_2021, oquab_dinov2_2024}; ProtoSSL explicitly represents this structure in prototype activations, enabling prototypes to be aligned with labels and grounded in representative examples via assignment and projection. We further detail in \autoref{app:ssl_interpretability} an intuition for why prototypes learned via SSL may support downstream case-based interpretability.


\subsection{Efficient assignment of prototypes to downstream labels}
\label{sec:assign_main}

After self-supervised pretraining, we assign prototypes from the learned latent prototype bank to labels in a downstream task. This step is necessary for enabling downstream interpretability using positive exemplars for predictions.

To that end, given \(K\) learned prototypes and \(L\) labels, we aim to assign \(M\) prototypes per label for a total of \(L \times M\) assignments. \ul{The key insight of our approach is a reduction of this problem to the rectangular linear assignment problem, which admits a known polynomial time solution via linear programming (LP)} \citep{kuhn_hungarian_1955}.

For ease of exposition, we let \(M = 1\) and omit it for clarity. Formally, we construct an assignment matrix \(S \in \mathbb{R}^{K,L}\), where \(s_{k,l} = \mathds{1}\)[prototype \(k\) is assigned to label-slot \(l\)] and \(1 \leq l \leq L\). We require each label-slot to be assigned exactly one prototype, while prototypes may remain unassigned; this is feasible whenever \(K \geq L\). We solve the following LP:
\[
\min \sum_k \sum_l -q_{k,l}\cdot s_{k,l} ~~~\text{s.t.}~~~ \sum_k s_{k,l} = 1, \forall l ~~~\text{and}~~~ \sum_l s_{k,l} \leq 1, \forall k
\]
where \(Q \in \mathbb{R}^{K,L}\) is an association score between each prototype \(p_k\) and label \(l\). We compute this association score using prototype activation statistics over the labeled downstream dataset. Given \(N\) samples, let \(A \in \mathbb{R}^{N,K}\) be the matrix of prototype activations for each sample, \(Y \in \{0,1\}^{N,L}\) denotes the binary multilabel matrix, where \(y_{n,l}=1\) if sample \(n\) has label \(l\) and 0 otherwise, \(y_n \in \{0,1\}^L\) denotes the \(n\)-th row of \(Y\), and \(t^+,t^- \in \mathbb{R}^L\) denotes the number of times each label occurs as positive: \(t^+ = \sum_{n=1}^N y_n\) or negative: \(t^- = \sum_{n=1}^N (1-y_n)\). The prototype-label association score matrix \(Q\) is thus given by:
\[
\begin{aligned}
Q = (\bar{Q}^+ - \bar{Q}^-) ~~~/&~~~ \sqrt{\tfrac{1}{2}(Q_{var}^+ + Q_{var}^-)} \in \mathbb{R}^{K,L}\\
\bar{Q}^+ = A^\top Y / t^+; &~~~
\bar{Q}^- = A^\top (1-Y) / t^-\\
Q_{var}^+ = ((A^2)^\top Y / t^+) - (\bar{Q}^+)^2; &~~~
Q_{var}^- = ((A^2)^\top (1-Y) / t^-) - (\bar{Q}^-)^2
\end{aligned}
\]
where squares are element-wise and vectors broadcast. Finally, given the solved assignments \(S\), we construct the label-assigned prototype matrix \(P' = S^\top P \in \mathbb{R}^{L,D}\), where each row in \(P'\) corresponds to a label-slot \(l\). The formulation generalizes naturally to \(M > 1\) slots per label, yielding \(P' \in \mathbb{R}^{LM,D}\).

\begin{table}[tp]
\centering
\caption{\textbf{List of datasets and their modeling tasks used in our study.} ProtoSSL pretraining does not use the labels defined by HEEDB or AudioSet, however SupProto pretraining does rely on the these datasets' labels.}
\label{tab:datasets}
\begin{adjustbox}{max width=\textwidth}


\footnotesize
\begin{tabular}{lllcccr}
\toprule
Domain & Dataset & Task family & \# Labels & Label type & \makecell{Label overlap\\with pretraining} & Metric \\
\midrule

\multirow[c]{7}{*}{ECG} &
HEEDB \citep{koscova_harvard-emory_2026} &
ECG interpretation &
20 &
Multilabel &
N/A & 
N/A \\

\cmidrule{2-7}

&
PTB-XL \citep{wagner_ptb-xl_2020} &
ECG interpretation &
71 &
Multilabel &
\cmark &
Macro AUROC \\

&
CinC Georgia \citep{perez_alday_classification_2020} &
ECG interpretation &
50 &
Multilabel &
\cmark &
Macro AUROC \\

&
CODE-15\% \citep{lima_deep_2021} &
ECG interpretation and outcomes &
7 &
Multilabel &
\cmark &
Macro AUROC \\

\cmidrule{2-7}

&
EchoNext \citep{poterucha_detecting_2025} &
Structural heart disease &
12 &
Multilabel &
\xmark &
Macro AUROC \\

&
MIMIC-IV-ECG \citep{gow_mimic-iv-ecg_nodate} &
Phecodes \citep{sethi_prototype_2025} \& outcomes \citep{alcaraz_enhancing_2025} \citep{strodthoff_prospects_2024} &
46 &
Multilabel &
\xmark &
Macro AUROC \\

&
ZZU pECG \citep{tan_pediatric_2025} &
Pediatric cardiac conditions &
4 &
Multilabel &
\xmark &
Macro AUROC \\

\midrule


\multirow[c]{4}{*}{Audio} &
AudioSet \citep{gemmeke_audio_2017} &
General sound tagging &
527 &
Multilabel &
N/A &
N/A \\

\cmidrule{2-7}

&
UrbanSound8K \citep{salamon_dataset_2014} &
Environmental sound classification &
10 &
Multiclass &
\cmark &
Accuracy \\

\cmidrule{2-7}

&
IEMOCAP \citep{busso_iemocap_2008} &
Emotion classification \citep{yang_superb_2021} &
4 &
Multiclass &
\xmark &
Accuracy \\

&
VoxCeleb1 ID \citep{nagrani_voxceleb_2017} &
Speaker identification &
1251 &
Multiclass &
\xmark &
Accuracy \\

\bottomrule
\end{tabular}

\end{adjustbox}
\end{table}

Intuitively, \(Q\) measures how selectively each prototype is associated with each label by comparing its activation on samples where the label is present versus absent, normalized by the variability of those activations. This corresponds to a standardized effect size (analogous to Cohen’s \(d\) \citep{lakens_calculating_2013}) which quantifies how well a prototype separates positive and negative examples for a given label. As a result, prototypes that consistently activate more strongly for one class than others receive higher scores, independent of their overall activation magnitude. Solving the assignment \(S\) using these scores introduces task-specific semantics by selecting prototypes that best distinguish each label, without requiring the latent prototype bank itself to be relearned. In practice, we balance the positive and negative class when computing \(Q\) through repeated sampling of the majority class (we refer to our code for the precise implementation, see \autoref{app:code_avail}).

\ul{Importantly, label assignment is training-free and non-parametric}. The pretrained encoder and prototypes remain frozen, only computing embeddings with prototype activations and solving a linear program. After label assignment, we consider two classifier training regimes:
\begin{itemize}[leftmargin=*]
    \item \textbf{Frozen probing transfer.} The encoder and prototypes from self-supervised pretraining remain frozen. The assigned prototypes are projected under label-supervision before a logistic-regression classifier is trained on frozen, projected prototype activations (\autoref{sec:projection}).
    \item \textbf{Fine-tuning transfer.} The encoder and label-assigned prototypes are initialized from self-supervised pretraining and fine-tuned on labeled data following the standard prototype part objectives \citep{chen_this_2019, sethi_protoecgnet_2025} (\autoref{sec:label-sup}). Subsequent label-supervised projection and logistic classifier training over frozen, projected prototype activations as before (\autoref{sec:projection}).
\end{itemize}

\subsection{Label-supervised Projection and Classifier Training}
\label{sec:projection}

\begin{table}[tp]
\centering
\caption{\textbf{Prototypes learned via ProtoSSL are transferrable to multiple ECG domains}. Fine-tuning ProtoSSL surpasses baselines, while frozen linear-probing of ProtoSSL is competitive against end-to-end training from scratch. Macro AUROC (bootstrapped 95\% CI), averaged across 5 random seeds. Best model \textbf{bolded}, next best model \Xul{underlined}.}
\label{tab:ecg_full_scale}
{
\setlength{\tabcolsep}{4pt}
\begin{adjustbox}{max width=\textwidth}

\begin{tabular}{@{}l@{\hspace{2pt}}l@{\hspace{4pt}}lccccc@{}}
\toprule
& & \multirow{2}{*}[-2pt]{Dataset} & \multicolumn{2}{c}{Tuned} & \multicolumn{2}{c}{Probed} & \multirow{2}{*}[-2pt]{SupProto Direct} \\
\cmidrule(lr){4-5}\cmidrule(lr){6-7}
& &  & ProtoSSL HEEDB & SupProto HEEDB & ProtoSSL HEEDB & SupProto HEEDB & \\
\midrule
\multirow{6}{*}{\raisebox{-0.75\height}{\rotatebox{90}{\makecell{Label Overlap\\\Xul{with HEEDB}}}}} & \multirow{3}{*}{\rotatebox{90}{No}} & EchoNext & \textbf{0.819} [0.818-0.820] & 0.782 [0.780-0.783] & \Xul{0.810} [0.809-0.811] & 0.763 [0.762-0.764] & 0.805 [0.804-0.806] \\
& & MIMIC-IV-ECG & \textbf{0.813} [0.811-0.814] & 0.788 [0.786-0.789] & \Xul{0.806} [0.804-0.807] & 0.774 [0.772-0.776] & 0.805 [0.804-0.806] \\
& & ZZU pECG & \textbf{0.815} [0.813-0.817] & 0.751 [0.748-0.753] & 0.776 [0.775-0.778] & 0.696 [0.694-0.699] & \Xul{0.790} [0.788-0.792] \\
\cmidrule{2-8}
& \multirow{3}{*}{\rotatebox{90}{Yes}} & PTB-XL & \textbf{0.911} [0.910-0.913] & 0.899 [0.897-0.901] & 0.881 [0.879-0.883] & 0.836 [0.834-0.838] & \Xul{0.900} [0.899-0.902] \\
& & CinC Georgia & \textbf{0.883} [0.881-0.885] & 0.858 [0.855-0.860] & 0.845 [0.842-0.847] & 0.831 [0.829-0.834] & \Xul{0.862} [0.859-0.864] \\
& & CODE-15\% & \textbf{0.966} [0.966-0.966] & 0.960 [0.960-0.961] & 0.949 [0.949-0.950] & 0.948 [0.948-0.948] & \Xul{0.962} [0.962-0.962] \\
\bottomrule
\end{tabular}

\end{adjustbox}
}
\end{table}

Latent prototypes initially learned without label-supervision must first be assigned to labels (as described in \autoref{sec:assign_main}). With a prototype-to-label mapping (either via assignment or classical label-supervised construction), each prototype is projected onto a representative sample (which may be a localized segment) from the downstream training set. \ul{This preserves the defining step which grounds projection-based prototype models in real cases}. A classifier trained on activation vectors over these projected prototypes is thus ante-hoc interpretable, both globally (per-label prototypes) and by case (prototypes supporting a specific prediction).

\paragraph{Label-supervised projection onto real exemplars:}
Projection is accomplished under label-supervision (as proposed in the original ProtoPNet \citep{chen_this_2019}), where a learned, latent, and labeled prototype is replaced with the encoder representation of the training case which is most similar to the prototype and has the prototype's label. Formally, given the encoder embeddings of all training samples \(H = f_\theta(\mathcal{X})\in \mathbb{R}^{N,D}\), the activation matrix \(A' = \text{sim}(H,P') \in \mathbb{R}^{N,L}\) relative to the learned and assigned prototypes \(P'\), and the corresponding binary multilabels \(Y \in \mathbb{R}^{N,L}\), the projected prototype matrix is given by:
\[
P'' = [p''_l]_{l=1}^{L};~~~p''_l = f_\theta(x^*_l);~~~x^*_l=\arg\max_{x_n \in \mathcal{X}_l}~a'_{n,l};~~~\mathcal{X}_l = \{x_n\in\mathcal{X}~|~y_{n,l}=1\}
\]
where \(x^*_l\) is the sample with the greatest activation for prototype \(p'_l\), selected from the subset of training samples \(\mathcal{X}_l\) belonging to label \(l\). In practice, we use patch-level prototype activations (detailed in Supplementary Sections \ref{app:ecg_details} and \ref{app:audio_details}).

\paragraph{Classifier training:}
To train the final classifier, prototype activations \(A''\) are recomputed relative to the projected prototypes \(P''\). These are used to train per-label logistic regressions \(g_l = \text{LogReg}(A'', y_{\cdot,l})\) in the multilabel setting or a multiclass logistic regression in the multiclass setting.

We note that our classifier training differs from prior work \citep{sethi_protoecgnet_2025, chen_this_2019, barnett_improving_2024} in that we do not initialize deep classifier weights with prototype connections followed by fine-tuning with masked regularization (which pushes coefficients to zero for off-target prototype activations). While these techniques help encourage interpretable predictions when using a deep classifier, we find that simply using a logistic regression classifier results in the model still learning to use positive prototype connections to drive predictions (\autoref{tab:ecg_proto_coef}). Further, our simplification allows prototype activations to be directly probed using out-of-the-box software packages, such as scikit-learn.

\subsection{Label-supervised Prototype Learning}
\label{sec:label-sup}

Label-supervised prototype learning trains each prototype to be specific to its assigned label. This requires a known prototype-to-label mapping, which may be derived via our efficient prototype assignment method (\autoref{sec:assign_main}), though traditionally this mapping is derived by construction (prototypes are randomly initialized for each label). Thus label-supervision can be used for fine-tuning pretrained-and-assigned prototypes or to learn prototypes de novo.

\begin{figure*}[tp]
    \centering
    \includegraphics[width=0.95\linewidth]{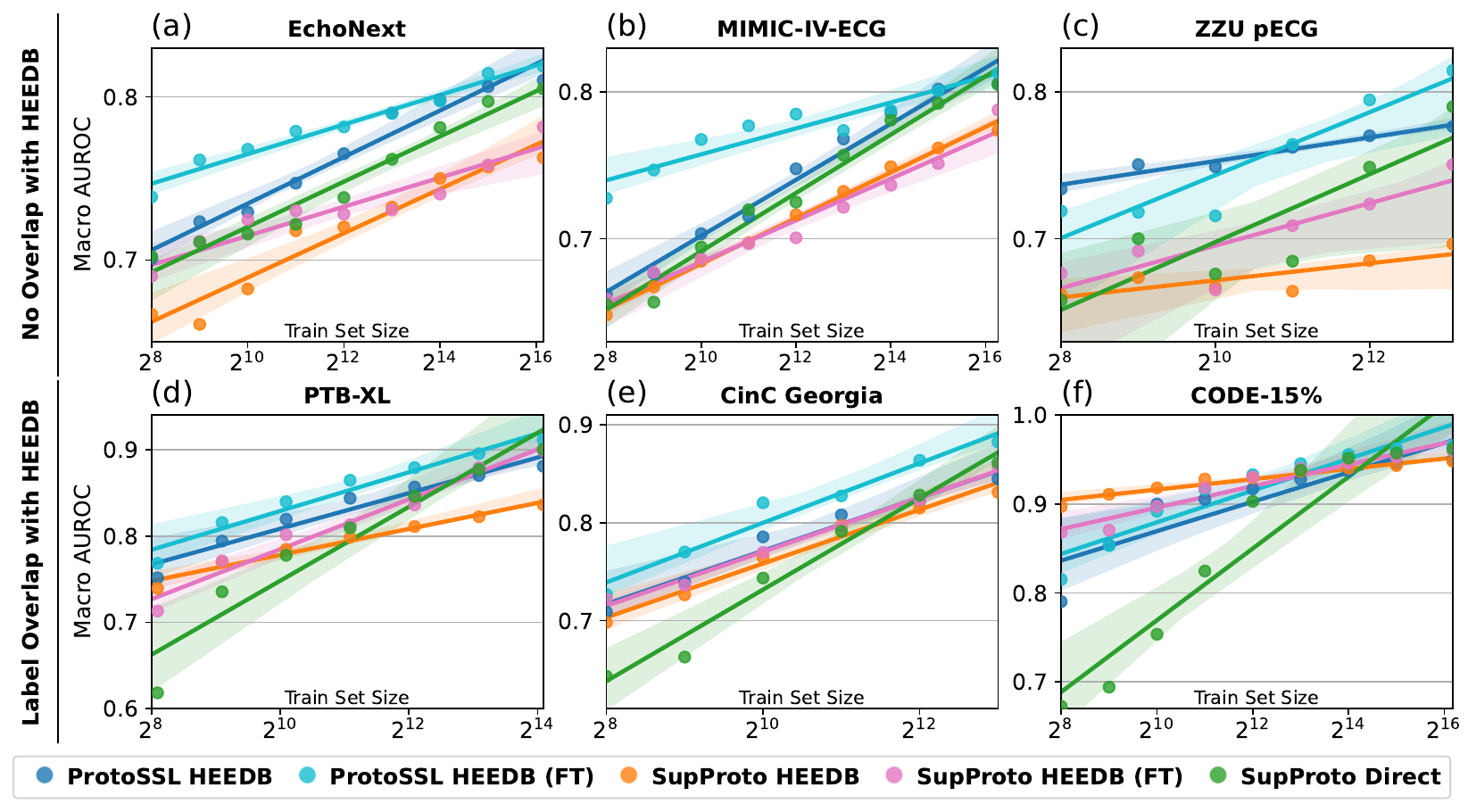}
    \caption{\textbf{ProtoSSL requires very few samples to successfully adapt pretrained prototypes to downstream tasks}, outperforming label-supervised baselines in both the frozen linear probing and fine-tuning (FT) settings. Results are averaged across 5 random seeds, prior to fitting a line of best fit (shaded region denotes 95\% CI for the line of best fit).}
    \label{fig:ecg_efficiency}
\end{figure*}

As we largely follow the method originally proposed by \citet{chen_this_2019}, along with adaptations by \citet{wang_interpretable_2021}, \citet{barnett_improving_2024}, and \citet{sethi_protoecgnet_2025}, we briefly summarize here the training objective:
\[
\mathcal{L}_{\text{SupProto}} = \mathcal{L}_{\text{CE}} + \lambda_{\text{Clst}}\mathcal{L}_{\text{Clst}} + \lambda_{\text{Sep}}\mathcal{L}_{\text{Sep}} + \lambda_{\text{Div}}\mathcal{L}_{\text{Div}} + \lambda_{\text{Cntrst}}\mathcal{L}_{\text{Cntrst}}
\]
where \(\mathcal{L}_{\text{CE}}\) is the standard cross entropy loss of logits from a final classifier head over prototype activations, \(\mathcal{L}_{\text{Clst}}\) pulls prototypes for each label closer to the latent representations of samples which are positive for the label, \(\mathcal{L}_{\text{Sep}}\) has the opposite effect to \(\mathcal{L}_{\text{Clst}}\), \(\mathcal{L}_{\text{Div}}\) encourages prototypes to be orthogonal in latent space, and \(\mathcal{L}_{\text{Cntrst}}\) is a contrastive term that encourages prototypes for co-occurring labels to be closer together in latent space. The \(\lambda\) coefficients control the relative strength of each of these terms. Under this objective, both prototypes and encoder are trained jointly.

For our experiments in ECG, we use \(\lambda_{\text{Clst}} = 0.004\), \(\lambda_{\text{Sep}} = 0.0004\), \(\lambda_{\text{Div}} = 250\), and \(\lambda_{\text{Cntrst}} = 300\), taken from \citet{sethi_protoecgnet_2025}. In our experiments in audio, we use \(\lambda_{\text{Clst}} = 0.8\), \(\lambda_{\text{Sep}} = 0.08\), \(\lambda_{\text{Div}} = 100\), and \(\lambda_{\text{Cntrst}} = 0\) from \citet{barnett_improving_2024}. As described in \autoref{app:audio_details}, we do not use the contrastive term for audio, as co-occurring labels are less meaningful in that context than in ECG.

\section{Results}
\label{sec:results_main}

We primarily study ProtoSSL in ECG classification, where model interpretability is particularly important and downstream tasks vary widely in semantics. Experimental details and baselines are provided in \autoref{sec:ecg_setup}. We first show that ProtoSSL produces transferrable ECG prototypes across six datasets (\autoref{tab:datasets}): frozen transfer is competitive across tasks, and fine-tuning yields the strongest prototype-based performance overall (\autoref{sec:generalization}). We then evaluate label efficiency, showing that ProtoSSL remains effective in low-data regimes with as few as 256 labeled examples (\autoref{sec:label_efficiency}). To assess whether prototypes learned without task labels remain interpretable, we conduct a controlled human evaluation of ECG explanations (\autoref{sec:user_study_results}). We test whether the same framework extends beyond ECG through audio transfer experiments (\autoref{sec:audio_results}). Finally, we conduct extensive ablations to understand the components of ProtoSSL (\autoref{sec:ablations_main}). A summary of datasets used in our experiments is presented in \autoref{tab:datasets}.

\begin{table}[tp]
\centering
\setlength{\tabcolsep}{3.5pt}
\renewcommand{\arraystretch}{0.95}
\caption{\textbf{In a user study, frozen prototypes transferred by ProtoSSL are rated higher than those learned directly on PTB-XL via label supervision.}
Acceptability denotes the proportion of responses rated as \enquote{good.} Differences are participant-level paired mean differences. \(p\)-values are from exact two-sided Wilcoxon signed-rank tests.}
\label{tab:user_study}

\footnotesize
\begin{tabular}{lcccc}
\toprule
Metric & \textbf{ProtoSSL HEEDB} & SupProto Direct & Difference (95\% CI) & $p$ \\
\midrule
Prototype acceptability (\%) & \textbf{91.4} & 66.4 & +25.0 [14.3, 35.7] & 0.0156 \\
Explanation acceptability (\%) & \textbf{82.9} & 67.9 & +15.0 [5.0, 25.0] & 0.0156 \\
\bottomrule
\end{tabular}

\end{table}
\subsection{ECG Experimental Setup}
\label{sec:ecg_setup}

We pretrain ProtoSSL on HEEDB \citep{koscova_harvard-emory_2026} and evaluate the resulting model (\textbf{ProtoSSL HEEDB}) on six downstream ECG datasets (\autoref{tab:datasets}), spanning both ECG interpretation and broader clinical prediction tasks. For ECG experiments, we choose 1000 prototypes in pretraining to create a large prototype bank and 14 prototypes per label in all downstream tasks (see \autoref{app:num_proto}). We average results across five random seeds over nested stratified subsets of training data. See \autoref{app:ecg_details} for more details.

\ul{As our goal is to understand pretraining and transfer of \emph{intrinsically interpretable} models, our direct comparisons are to projection-based prototype models}. We first compare to \textbf{SupProto Direct}, a label-supervised prototype model trained from scratch on each downstream dataset (following \citet{sethi_protoecgnet_2025}), as the direct interpretable baseline. \textbf{SupProto HEEDB} instead uses label-supervised prototype pretraining on the 20 HEEDB ECG interpretation labels (derived and clinically validated by \citet{li_electrocardiogram_2025}), then applies the same downstream assignment and projection procedure as ProtoSSL. Thus, \emph{SupProto HEEDB isolates the effect of the self-supervised pretraining objective, but both methods use our assignment mechanism}.

While not the primary focus of our work, we compare to \enquote{blackbox} ECG baselines to contextualize the tradeoff of prototype-based modeling (\autoref{app:blackbox}). Specifically, we compare to a deep classifier which shares the same encoder architecture as ProtoSSL, either trained directly on the downstream classification task or probed after contrastive pretraining with PCLR. In addition, we probe ECGFounder \citep{li_electrocardiogram_2025} and ST-MEM \citep{na_guiding_2023}, following \citet{al-masud_benchmarking_2026}. \ul{We emphasize that other prototype part models are the primary comparison, as prototype model explanations are more faithful to model reasoning and interpretable than post-hoc explanations} \citep{li_deep_2018, rudin_stop_2019, chen_this_2019, sethi_prototype_2025, heinrich_audioprotopnet_2025, barnett_improving_2024, barnett_case-based_2021, de_santi_part-prototype_2024, turbe_evaluation_2023}.

\subsection{ProtoSSL learns foundational ECG prototypes}
\label{sec:generalization}

We report in \autoref{tab:ecg_full_scale} the macro AUROC of ProtoSSL HEEDB on six diverse downstream ECG datasets, compared to label-supervised baselines under both fine-tuning and frozen linear probing settings. We find that fine-tuning \ul{ProtoSSL consistently results in the most performant prototype-based model, despite its initial latent prototypes not being tailored to any specific predictive task}. More strikingly, using these same initial prototypes, training-free prototype adaptation with projection followed by linear probing of ProtoSSL HEEDB remains competitive with full end-to-end training of a deep, prototype-based model from scratch (SupProto Direct). Computationally, \emph{the adaptation and linear probing of ProtoSSL does not require a GPU and finishes in mere seconds} (quantified in \autoref{tab:assignment_runtime}). Critically, frozen probing of ProtoSSL is performant and efficient while preserving user interpretability (\autoref{sec:user_study_results}).

Additionally, we establish that ProtoSSL's gains are not simply due to extensive pretraining: SupProto HEEDB was pretrained over ECG interpretation tasks that have a high degree of overlap with the label-overlap datasets enumerated in \autoref{tab:ecg_full_scale}. Consequently, we find that SupProto HEEDB performs substantially worse on out-of-domain datasets whose labels encompass tasks beyond simple ECG interpretation. Further, despite being in-domain for SupProto HEEDB, we find that ProtoSSL HEEDB generally outperforms it on datasets spanning ECG interpretation as well, under both fine-tuned and frozen settings. Together, these results indicate that \ul{ProtoSSL's task agnostic motif discovery learns a latent prototype bank that can serve as a foundation for diverse downstream tasks, while label-supervised prototype pretraining diminishes out-of-domain generalization}.

\begin{table}[tp]
\centering
\caption{\textbf{ProtoSSL enables prototype transfer in audio.} Accuracy (bootstrapped 95\% CI), averaged over provided cross-validation splits for UrbanSound8K and IEMOCAP. For each dataset, best model \textbf{bolded}, second best model \Xul{underlined}.}
\label{tab:audio_results}
{
\setlength{\tabcolsep}{4pt}
\begin{adjustbox}{max width=\textwidth}

\begin{tabular}{ccccccc}
\toprule
\multirow{2}{*}[-2pt]{\makecell{Label Overlap\\w/ AudioSet}} & \multirow{2}{*}[-2pt]{Dataset} & \multicolumn{2}{c}{Tuned} & \multicolumn{2}{c}{Probed} & \multirow{2}{*}[-2pt]{SupProto Direct} \\
\cmidrule(lr){3-4}\cmidrule(lr){5-6}
& & ProtoSSL AudioSet & SupProto AudioSet & ProtoSSL AudioSet & SupProto AudioSet & \\
\midrule
\multirow{2}{*}{No} & VoxCeleb1 ID & \textbf{0.438} [0.438-0.438] & \Xul{0.426} [0.426-0.427] & 0.042 [0.042-0.042] & 0.003 [0.003-0.004] & 0.399 [0.399-0.399] \\
& IEMOCAP & \Xul{0.591} [0.590-0.592] & \Xul{0.591} [0.590-0.592] & 0.541 [0.540-0.542] & 0.531 [0.530-0.532] & \textbf{0.606} [0.605-0.607] \\
\midrule
Yes & UrbanSound8K & \Xul{0.820} [0.819-0.820] & \textbf{0.840} [0.839-0.841] & 0.723 [0.722-0.724] & 0.783 [0.782-0.783] & 0.756 [0.755-0.756] \\
\bottomrule
\end{tabular}

\end{adjustbox}
}
\end{table}
\subsection{ProtoSSL improves label efficiency across ECG tasks}
\label{sec:label_efficiency}

We establish ProtoSSL's label efficiency across all six ECG datasets using nested stratified training subsets down to just 256 samples (\autoref{fig:ecg_efficiency}, \autoref{tab:ecg_full}). \ul{ProtoSSL HEEDB not only outperforms label-supervised baselines at full dataset scale, but also remains performant across dataset scales}. The strongest gains appear over datasets where label-supervised baselines cannot rely on overlap with pretraining labels (\autoref{fig:ecg_efficiency}a--c): frozen probing of ProtoSSL HEEDB consistently outperforms both SupProto Direct and SupProto HEEDB across the low-data regime. Fine-tuning can further enhance ProtoSSL's performance, remaining beneficial even under extreme data scarcity (\autoref{fig:ecg_efficiency}a,b). Remarkably, under fine-tuning, \emph{ProtoSSL HEEDB with only \emph{4K} samples matches SupProto HEEDB performance at full dataset scale (over \emph{64K} samples)} on EchoNext and MIMIC (\autoref{tab:ecg_full}). To our knowledge, no prior studies have systematically evaluated projection-based prototype models across data scales and in low-label settings. Our results suggest that supervised prototype learning (SupProto Direct) is as difficult as training blackbox classifiers when labeled examples are scarce (\autoref{tab:ecg_blackbox}), whereas a pretrained, reusable latent prototype bank may work with as few as 256 labeled samples.

However, \ul{pretraining alone is insufficient to learn foundational prototypes for low-data regimes}. As in \autoref{sec:generalization}, we find that label-supervised prototype pretraining (SupProto HEEDB) performs well primarily on downstream tasks that share labels with the pretraining dataset. \emph{We hypothesize that label-supervision tailors the learned prototypes to the pretraining domain's task, hindering downstream adaptability}. Conversely, ProtoSSL performs well across all studied ECG datasets, suggesting that \ul{self-supervision is crucial for diverse generalization across downstream ECG tasks}. Lastly, we observe that fine-tuning pretrained prototypes (both ProtoSSL HEEDB and SupProto HEEDB) may not always be suitable across all scales (\autoref{fig:ecg_efficiency}c,f). We attribute this to the extreme label imbalance in these datasets, where many labels only have a single instance in the smaller subsets (see \autoref{app:data}), suggesting that additional considerations are necessary when adapting prototype models in extreme data scarcity and label rarity settings. Nonetheless, taken together, \ul{ProtoSSL yields repeatable gains in label-scarce settings, positioning it as a promising foundation for studies otherwise limited by data availability}.

\subsection{ProtoSSL is human-interpretable}
\label{sec:user_study_results}

Through a blinded, IRB-approved user study, \ul{we demonstrate that frozen ECG prototypes learned by ProtoSSL support human-interpretable explanations}.
While projection-based prototype models intrinsically rely on prototype similarity as the basis for prediction, whether a human would evaluate a prototype to be similar to an inference target is an empirical property that must be measured per-model \citep{davoodi_interpretability_2023}. Following \citet{davoodi_interpretability_2023}, we adapt their design to study human-model similarity coherence in the context of ECG interpretation. Full study design is detailed in \autoref{app:user_study}.

Over 20 PTB-XL \citep{wagner_ptb-xl_2020} test cases spanning four ECG interpretation labels (AMI, LBBB, RBBB, PVC), we compared the interpretability of frozen probing of ProtoSSL HEEDB to SupProto Direct, for which a similar model has been previously demonstrated to be interpretable \citep{sethi_protoecgnet_2025}. Seven participants (final-year medical students with formal ECG interpretation training and clinical experience) evaluated (i) prototype quality in isolation, defined as whether the prototype was a good example of the predicted diagnosis, and (ii) prototype-based case explanations, defined as whether the prototype-test ECG pair formed a convincing match. \ul{Across both tasks, ProtoSSL was rated more favorably than SupProto Direct} (\autoref{tab:user_study}). Notably, SupProto Direct achieves higher predictive performance on the same dataset (PTB-XL, see \autoref{tab:ecg_full_scale}), yet produces prototypes that are less consistently judged as good examples of the underlying diagnosis. All study cases were restricted to correct predictions for both models, isolating explanation quality from classification accuracy. This suggests that stronger discriminative performance does not necessarily imply human-preferred prototypes. We interpret this result as evidence that the objective used to learn prototypes influences the type of examples they recover. End-to-end label supervision may select patterns that are useful for classification but less consistently perceived as clear examples of a diagnosis by human evaluators. In contrast, learning prototypes from the unlabeled data distribution appears to preserve patterns that are more often judged as representative. Thus, \ul{separating motif discovery from label alignment not only enables transfer, but can also improve the perceived quality of case-based explanations}. We examine qualitative examples in \autoref{app:user_study_analysis}.

\subsection{ProtoSSL learns transferrable audio prototypes}
\label{sec:audio_results}

We find that ProtoSSL yields gains in prototype learning for audio. We pretrain ProtoSSL on AudioSet \citep{gemmeke_audio_2017} (\textbf{ProtoSSL AudioSet}) and transfer to three downstream tasks (\autoref{tab:datasets}): UrbanSound8K \citep{salamon_dataset_2014}, whose environmental sound labels substantially overlap with AudioSet, and IEMOCAP \citep{busso_iemocap_2008} and VoxCeleb1 identification \citep{nagrani_voxceleb_2017}, which have minimal label overlap. As in ECG, we compare to learning prototypes directly on the target task (\textbf{SupProto Direct}) and transferring label-supervised prototypes pretrained on AudioSet (\textbf{SupProto AudioSet}) using our assignment method. We provide further implementation details in \autoref{app:audio_details}.

The results in \autoref{tab:audio_results} demonstrate that fine-tuned ProtoSSL outperforms SupProto Direct on VoxCeleb1 and UrbanSound8K. ProtoSSL AudioSet also outperforms SupProto AudioSet on VoxCeleb1 and matches on IEMOCAP, two downstream datasets with minimal label overlap relative to label-supervised pretraining. Conversely, SupProto AudioSet performs best on UrbanSound8K, where 9 of 10 labels are direct AudioSet labels and the remaining UrbanSound8K label (street music) is closely related to multiple AudioSet labels. These results mirror those observed in ECG: \ul{label-supervised prototypes transfer well when the target task preserves the source label semantics, while ProtoSSL may generalize better when the downstream label space differs}.

The audio results broaden the empirical scope of ProtoSSL. Compared with ECG, the audio experiments significantly change the input representation, architecture, and self-supervised signal (\autoref{app:audio_details}). The three audio tasks are also substantially different from one another, spanning environmental audio classification, emotion recognition, and speaker identification. This task diversity is greater than in the ECG experiments and helps explain why audio transfer is less uniform, particularly in the frozen setting. The key result is that, despite these changes, \emph{the ProtoSSL framework continues to produce transferable prototype models}. Together, these results show that \ul{the separation of motif discovery and label alignment extends beyond ECG, enabling projection-grounded prototype models to be transferred from unlabeled data to downstream tasks with different label spaces}.


\subsection{Ablations isolate the contribution of each ProtoSSL component}
\label{sec:ablations_main}

We ablate each stage of ProtoSSL (self-supervised prototype pretraining, prototype-label assignment, and the prototype bottleneck itself), primarily on EchoNext, given its large training set, moderate number of labels, and lack of label overlap with HEEDB. Full results are provided in \autoref{app:more_results}.

\paragraph{Transferable structure comes from self-supervised prototype pretraining.}
ProtoSSL outperforms SupProto HEEDB even when the supervised baseline transfers unprojected prototypes (\autoref{app:supproto_heedb_pila}, \autoref{tab:supproto_noproj}) or is pretrained on a 7.5\(\times\) larger label space (\autoref{app:heedb_150}, \autoref{tab:heedb_150}), so \ul{the limitation of label supervision is prototype semantics fixed to a source label space, not a lack of signal}. A strong foundation model is no substitute: prototypes clustered \citep{zhu_interpretable_2025} from frozen ECGFounder embeddings underperform ProtoSSL and are no better than random prototypes in the same space (\autoref{app:protos_from_fms}, Supplementary Tables \ref{tab:ecg_protos_from_fms}--\ref{tab:ecg_sk_ot_converge}). ProtoSSL also improves with larger downstream prototype budgets where the supervised baseline does not (\autoref{app:num_proto}, \autoref{tab:ppl}); relatedly, a larger upstream bank helps most under distribution shift, e.g. between adult to pediatric ECG (\autoref{app:num_proto_upstream}, \autoref{tab:zzu_capacity}).

\paragraph{Assignment makes this structure interpretable at little cost.}
Training-free LAP assignment is competitive with a learned ProtoPool-derived alternative \citep{rymarczyk_interpretable_2022} while running 8.5--76\(\times\) faster (\autoref{app:alt_assign}, Supplementary Tables \ref{tab:ecg_alt_assign}--\ref{tab:assignment_runtime}), and it closely matches non-interpretable variants that skip assignment entirely, except in the smallest data regimes (\autoref{tab:ecg_no_assign}). Even without enforced masking, the classifier consistently weights each label's assigned prototypes above all others (\autoref{app:classifier_explanation}, \autoref{tab:ecg_proto_coef}), so \ul{assignments are reflected in how predictions are made}.

\paragraph{Pretrained prototypes encode consistent diagnostic morphology.}
Independent assignments on PTB-XL, CinC Georgia, and CODE-15\% repeatedly select the same pretrained prototypes for the same diagnosis, far more often than chance (\autoref{app:proto_coherence}, \autoref{fig:venn}). For the sake of visualization, label-free projection of a latent prototype (shared by multiple downstream RBBB assignments) onto HEEDB pretraining samples retrieves an RBBB ECG with the characteristic morphology (\autoref{fig:heedb_rbbb}): \ul{a diagnostic motif discovered without diagnostic label supervision during pretraining}.

\paragraph{The prototype bottleneck carries a modest cost that a larger prototype budget can reduce.}
Probing ProtoSSL HEEDB with our default 14 prototypes per label is weaker on four of six datasets than probing PCLR HEEDB, the same encoder and pretraining objective without the prototype layer; the gap is modest at most scales but widens at the smallest training sizes on CODE-15\% (\autoref{app:blackbox}, \autoref{tab:ecg_blackbox}). With 83 prototypes per label, ProtoSSL HEEDB matches PCLR HEEDB on EchoNext at 8K samples and above (\autoref{tab:ecg_no_assign}).

Together, these ablations support the central design principle of ProtoSSL: \ul{self-supervised motif discovery supplies transferable structure that neither label-supervised pretraining nor foundation-model clustering recovers, while efficient label alignment and projection convert that structure into case-based explanations}.

\section{Discussion}
\paragraph{Limitations and future work.} ProtoSSL's latent prototype bank can only reuse structure that is sufficiently represented and learnable in the pretraining corpus, so its prototypes cannot transfer to waveform morphologies that are novel relative to pretraining; this is a known general limitation of self-supervised methods \citep{chen_intriguing_2021, tian_what_2020}. For simplicity, we use one-second windows for our ECG prototype modeling; however these windows may not capture multi-beat rhythms well; extending ProtoSSL to a multi-branch design as in ProtoECGNet \citep{sethi_protoecgnet_2025} is a natural next step. Finally, our human evaluation was restricted to jointly correct cases and measures explanation acceptability rather than clinical utility; evaluating incorrect and ambiguous cases in a decision-support setting \citep{barnett_improving_2024} remains future work. Further limitations are discussed in \autoref{app:limitations}.

\paragraph{Conclusion.} ProtoSSL demonstrates that learning latent, foundational prototypes, which support projection-based interpretability across multiple tasks, can be decoupled from task-specific label supervision. By learning prototypes from unlabeled data, assigning them to downstream labels, and grounding them through projection, ProtoSSL enables prototype transfer across label spaces. Across six ECG datasets, ProtoSSL improves label efficiency and, after fine-tuning, achieves the strongest prototype-based performance at full scale. A human evaluation study demonstrates that task-label-free prototype discovery yields interpretable case-based evidence. Finally, audio experiments demonstrate that ProtoSSL's separation of motif discovery and label alignment can support prototype transfer in a distinct application domain. Together, these results show that a transferrable latent prototype bank can be learned before the downstream label space is known; downstream assignment and target projection then produce a task-specific, projection-grounded, interpretable prototype model.

\clearpage


\bibliographystyle{unsrtnat}
\bibliography{proto_references_2026_09_29} 
\clearpage


\appendix
\appendix

\section{Code Availability}
\label{app:code_avail}

Our code is publicly available here: \url{https://github.com/StevenSong/ProtoSSL}

\section{Extended Methods}
\label{app:methods}

\subsection{ECG Experimental Details}
\label{app:ecg_details}

\paragraph{Input representation and prototype architecture:}
10-second, 12-lead ECG inputs are resampled to 100 Hz and normalized using per-lead statistics derived over each dataset's training set. Specifically, we clip values at the 99.9 and 0.1 percentiles, followed by z-score normalization. We ensure the leads are ordered following the standard lead ordering. Following \citet{sethi_protoecgnet_2025}, we adopt a 2D ResNet18-style encoder to embed ECGs by treating each ECG as a 2D input with a single channel, modifying the first convolutional layer to use a 2D kernel of height 12 to span all 12 leads, resulting in 512-dimensional embeddings.

A key feature of the prototype part models proposed by \citet{chen_this_2019} is the mapping of prototypes to individual parts of an input (patch embeddings of an image in the original computer vision formulation), referred to as ``partial'' prototypes. While we generally follow the adaptations of the method proposed by ProtoECGNet \citep{sethi_protoecgnet_2025} for ECG, we make a key modification to the way partial prototypes are derived. ProtoECGNet directly adapts the partial prototype mechanism from ProtoPNet, both of which map patch embeddings back to regions of the input without accounting for the true receptive field of each latent feature, which in reality can be much wider and increases in width as the network gets deeper \citep{araujo_computing_2019}. While the effective receptive field is likely a narrow region \citep{luo_understanding_2016} and prior prototype models work well empirically, this potentially confounds the interpretability of the prototypes and potentially limits the depth of encoders that are usable with prototype-based methods. In this work, we leverage the well-established approach in time-series modeling of sliding windows and compute a single ``patch'' embedding per window. Furthermore, we do not pool prototype activations across contiguous patches (as in ProtoECGNet); partial prototype activations precisely correspond only to a discrete window of the input signal. Together, these ensure that when a prototype-based explanation is presented visually to a user, the highlighted region of the input is exactly what the model is using to drive its prediction.

Formally, given an input ECG \(x\), let \(x_t\) denote a window of \(x\) and \(h_t = f_\theta(x_t)\) the ``patch'' embedding of the window. We thus compute prototype activations as:
\[
a(x) = [a_k(x)]_{k=1}^K;~~~ a_k(x) = \max_t\text{sim}(f_\theta(x_t), p_k)
\]
where the activation of each prototype \(p_k\) relative to \(x\) is the most similar window of \(x\). Specifically, we use windows of 1-second length with 50\% overlap between consecutive windows, resulting in 19 total windows for a 10-second ECG.

\paragraph{Self-supervised pretraining:}
As detailed in \autoref{sec:ecg_setup}, ProtoSSL for ECG is pretrained on HEEDB using 1000 prototypes, each 512-dimensional. This number of prototypes was chosen to create a large, reusable latent prototype bank. For the contrastive views of the ECG, we adopt PCLR \citep{diamant_patient_2022}, where multiple ECGs from the same patient are contrasted. We do not use any additional data augmentations. We refer to the resulting model as ProtoSSL HEEDB.

\paragraph{Supervised pretraining:}
To understand if a traditional, label-supervised prototype model can generalize to other tasks, we do label-supervised pretraining over HEEDB. This additionally allows us to understand the effect of self-supervision when learning prototypes. To achieve the same number of prototypes from self-supervised pretraining, we use 50 prototypes per label for the 20 labels derived from HEEDB by \citet{li_electrocardiogram_2025}. The final step of traditional label-supervised prototype modeling is to project learned prototypes onto training samples under label-supervision, as described in \autoref{sec:projection}. We refer to the resulting model as SupProto HEEDB.

\paragraph{Downstream transfer:}
After pretraining, ProtoSSL and SupProto HEEDB are transferred to downstream tasks. In both cases, we use the method described in \autoref{sec:assign_main} to assign prototypes to downstream labels (even though SupProto HEEDB's prototypes are initially learned under label-supervision, the downstream labels are never a strict subset of the pretraining labels and thus require assignment). If the prototype model is being fine-tuned, we jointly train the prototypes and encoder on the downstream dataset in a label-supervised prototype learning setting, as described in \autoref{sec:label-sup}. Finally, prototypes are projected onto downstream samples and a classifier is trained over prototype activations, following the method described in \autoref{sec:projection}. Unless otherwise stated, we use \(M=14\) prototypes per label for downstream ECG datasets, the maximum number of prototypes per label that can be achieved using 1000 prototypes in the pretrained set of prototypes across all of our downstream tasks (PTB-XL has 71 labels).


\subsection{Audio Experimental Details}
\label{app:audio_details}

\paragraph{Input representation and prototype architecture:}
Audio inputs are resampled to 32 kHz and converted to log-mel spectrograms. We adopt the 2D partial-prototype formulation used in AudioProtoPNet \citep{heinrich_audioprotopnet_2025}, which extends ProtoPNet \citep{chen_this_2019} to time-frequency representations by matching localized patches in a latent feature map.

Each input clip is processed by a Cnn14 encoder \citep{kong_panns_2020}, which operates on log-mel spectrograms and produces a 2D feature map. Prototypes correspond to small spatial patches in this representation and are compared to all locations using cosine similarity.

Spectrograms are computed using a 1024-point FFT, hop length 320, and 64 mel bins, with a frequency range of 50-14{,}000 Hz. During training, the encoder applies SpecAugment-style masking with two time masks (width 64) and two frequency masks (width 8).

\paragraph{Self-supervised pretraining:}
ProtoSSL for audio is pretrained on the full AudioSet unbalanced training set. Self supervised audio pretraining used 6324 prototypes to accommodate downstream datasets with many labels/classes.

We construct positive pairs by sampling two independent 2-second segments from the same audio waveform, following COLA \citep{saeed_contrastive_2020}. Each sampled waveform is augmented with random gain (applied with probability 0.2, with gain sampled uniformly from $-6$ to $+6$ dB) and additive Gaussian noise (applied with probability 0.2, with standard deviation sampled uniformly from $10^{-4}$ to $2\times10^{-3}$), and is then peak-normalized before conversion to a log-mel spectrogram. These waveform-space perturbations are consistent with audio-specific augmentation strategies studied in \citet{al-tahan_clar_2021}.

The resulting spectrograms are processed by the Cnn14 frontend \citep{kong_panns_2020}, which applies the same time- and frequency-masking used in that architecture. As in ECG, contrastive learning is applied directly to prototype activations, and we optimize the NT-Xent loss together with the KoLeo regularization term. We refer to the resulting model as ProtoSSL AudioSet.

\paragraph{Supervised pretraining:}
For label-supervised audio pretraining, we match encoder architecture and use 12 prototypes per label for AudioSet's 527 labels. To be comparable with self-supervised pretraining in audio, the same data augmentations are applied during supervised pretraining. Unlike ECG, we do not include a co-occurrence regularization term, as AudioSet label co-occurrence does not correspond to interpretable relationships between classes. Instead, we use the formulation from \citet{wang_interpretable_2021} and \citet{barnett_improving_2024}, consisting of classification, clustering, separation, and diversity losses, though we note the classification pretraining objective over AudioSet remains multilabel. As in ECG, we project prototypes onto AudioSet samples under label-supervision. We refer to the resulting model as SupProto AudioSet.

\paragraph{Downstream transfer:}
After pretraining, ProtoSSL and SupProto AudioSet are transferred to downstream tasks, first deriving prototype-label assignments (described in \autoref{sec:assign_main}), optionally fine-tuned under label supervision (described in \autoref{sec:label-sup}), and finally projected onto downstream samples with classifier training (described in \autoref{sec:projection}). Distinct from the ECG setting and from AudioSet pretraining, we train a multiclass classifier for all downstream audio classification tasks. We evaluate on three downstream datasets (\autoref{tab:datasets}): UrbanSound8K \citep{salamon_dataset_2014}, IEMOCAP \citep{busso_iemocap_2008}, and VoxCeleb1 identification \citep{nagrani_voxceleb_2017}. For all downstream datasets, we use 5 prototypes per label. As our prototypes operate over partial segments of the input via patches, audio samples across tasks can have different durations. Informed by each respective dataset, we use 4 second clips for UrbanSound8K, 4.5 second clips for IEMOCAP, and 3 second clips for VoxCeleb1. No augmentations are applied in the downstream datasets.


\subsection{Baseline Comparisons}
\label{app:baselines}

\paragraph{Label-supervised prototype models:}
We compare to two settings for label-supervised prototype model baselines: either training the model from scratch directly on the target dataset or transferring a pretrained model. We describe the pretrained transfer setting in \autoref{app:ecg_details} (SupProto HEEDB) and \autoref{app:audio_details} (SupProto AudioSet). To compare to a label-supervised prototype model trained directly on the target dataset without any pretraining (SupProto Direct), we adopt the same encoder backbone as described for ECG or audio. Additionally, we initialize the same number of prototypes per label as in the pretrained downstream transfer setting, i.e. 14 for ECG or 5 for audio datasets. The prototypes and encoder are trained directly under the label-supervised prototype learning objectives (\autoref{sec:label-sup}) prior to projection and classifier training (\autoref{sec:projection}). These SupProto Direct comparisons allow us to isolate the effect of pretraining.

\paragraph{Blackbox classifier:}
To contextualize the tradeoff of prototype learning with traditional deep classifiers, we additionally train a blackbox baseline which leverages the same encoder backbone as the prototype-based models in our experiments. In these models, the encoder is trained jointly with a linear classification head from scratch, directly on the target dataset. We train these models by optimizing over the standard cross entropy loss objective. We refer to these models as Blackbox Direct.

\paragraph{Foundation models:}
To contextualize the performance of linear probing of other pretrained models, particularly in terms of label efficiency, we evaluate notable ECG foundation models that performed well in a recent benchmarking study of ECG FMs \citep{al-masud_benchmarking_2026}. Specifically, we compute 1024-dimensional frozen embeddings from ECGFounder \citep{li_electrocardiogram_2025} and 768-dimensional frozen embeddings from ST-MEM \citep{na_guiding_2023}. We note that we consider these models blackboxes given interpretability for both relies on post-hoc explanations. ECGFounder serves as our label-supervised, pretrained blackbox comparison, while ST-MEM serves as our self-supervised, pretrained blackbox comparison. We probe the frozen embeddings of these models using logistic regressions.

\subsection{Training Details and Hyperparameters}
\label{app:training}

Across all experiments, we use cosine similarity as the similarity function when computing prototype activations. For training or fine-tuning deep architectures, we use an AdamW optimizer with learning rate $0.001$ and weight decay $0.01$, learning rate reduction by a factor of $0.1$ when validation loss does not improve (patience of $3$ epochs for ECG experiments and $10$ epochs for audio), early stopping when validation loss does not improve (patience of $10$ epochs for ECG and $20$ for audio), batch size 512 for ECG and 128 for audio, and a maximum of 1000 epochs for target datasets. For logistic regressions, we input z-score normalized prototype activations (normalized using training set statistics) to scikit-learn's logistic regression implementation with L2 regularization (\(C=0.0005\)) with the saga solver and 100 maximum iterations. Models were evaluated using 1000 bootstrapped resamples (with replacement and stratification) of the test set, where each resample was the size of the original test set. All experiments were conducted on a slurm cluster where each job was allocated 24x CPU cores, 200 GB of RAM, and 1x NVIDIA H200 GPU.


\subsection{Datasets and Splits}
\label{app:data}

An overview of the datasets and downstream tasks used in our study are provided in \autoref{tab:datasets}. We present in Supplementary Tables \ref{tab:ecg_dataset_heedb}--\ref{tab:ecg_dataset_code15} characteristics for data splits or subsets not directly defined in the literature (this primarily covers ECG data subsets which we construct to evaluate ProtoSSL's label efficiency).

\paragraph{ECG datasets:} We use the Harvard-Emory ECG Database (HEEDB) \citep{koscova_harvard-emory_2026} as our ECG pretraining dataset. We use human-corrected, machine-derived labels for tasks requiring HEEDB labels, specifically using the 20 labels derived and clinically validated by \citet{li_electrocardiogram_2025}. We derive per-site (Harvard vs Emory) normalization statistics. We filter to adult ECGs and use all ECGs prior to 2021 as training data, ECGs from 2021 as the validation set, and ECGs from 2022 as the test set.

For downstream datasets, we rely on dataset author-provided training/validation/test splits for EchoNext \citep{poterucha_detecting_2025}, PTB-XL \citep{wagner_ptb-xl_2020}, and the splits of MIMIC-IV-ECG \citep{gow_mimic-iv-ecg_nodate} defined by MIMIC-IV-Ext-MDS-ED \citep{alcaraz_enhancing_2025}. We construct initial splits at full-data scale for the CinC Georgia subset \citep{perez_alday_classification_2020}, CODE-15\% \citep{lima_deep_2021}, and ZZU pECG \citep{tan_pediatric_2025}, ensuring that patients are entirely contained within one split.

We utilize dataset author-provided definitions of labels and note the few instances where we derive additional labels. For MIMIC-IV-ECG, while we utilize the outcomes defined by \citet{alcaraz_enhancing_2025} and \citet{strodthoff_prospects_2024}, we additionally construct Phecode labels defined by \citet{sethi_prototype_2025}. For CODE-15\%, in addition to the ECG interpretation labels provided by \citet{lima_deep_2021}, we additionally compute a mortality label by binarizing survival data about the median follow-up time (approximately 3.5 years), requiring that patients either had observed mortality prior to this threshold or observed follow-up or mortality after this threshold. For ZZU pECG, we bin fine-grained labels into the 4 disease categories described by \citet{tan_pediatric_2025}.

For label-efficiency experiments over ECG datasets, we create nested and stratified subsets of each dataset's training set. When presented with a large number of labels, we rely on the stratification algorithm implemented by \citet{wagner_ptb-xl_2020} to create approximately stratified subsets. Even so, rare labels are often entirely absent in the smallest training subset derived by this algorithm. To preserve label presence across all subsets, we inject one randomly sampled ECG positive for each label. The characteristics of these subsets are presented in Supplementary Tables \ref{tab:ecg_dataset_mimic}--\ref{tab:ecg_dataset_code15}.

\paragraph{Audio datasets:} We use the AudioSet \citep{gemmeke_audio_2017} dataset for pretraining, using the unbalanced training set as our train split, balanced train as the validation split, and the eval set as the test split. In label-supervised experiments with AudioSet, we use its 527 binary labels.

For downstream audio datasets, we use dataset author-provided splits for all datasets. For UrbanSound8K \citep{salamon_dataset_2014} and IEMOCAP \citep{busso_iemocap_2008}, we cross validate results with the provided folds; in each iteration, we reserve 1 fold for test and 1 fold for validation and use all remaining folds for training. For VoxCeleb1 identification \citep{nagrani_voxceleb_2017}, we use the provided train/dev/test splits.

All audio waveforms are converted to mono-channel by averaging and peak-normalized. Samples are clipped and padded to the target length (dependent on the dataset); when not doing data augmentation, clips are left-aligned and truncated or padded to the target length. Specifically, following each dataset's authors, we use 10 second clips for AudioSet, 4 second clips for UrbanSound8K, 4.5 second clips for IEMOCAP, and 3 second clips for VoxCeleb1. No augmentations are applied in the downstream datasets. Further data augmentations for AudioSet are described in \autoref{app:audio_details}.

\section{Why self-supervised prototypes can support interpretable reasoning}
\label{app:ssl_interpretability}

A central question in our framework is why prototypes learned without task labels can still yield meaningful explanations. In supervised prototype models, interpretability is typically attributed to class-specific training objectives that explicitly align prototypes with labels. ProtoSSL removes this supervision during prototype learning, so any resulting interpretability must arise from the structure of the data and the learning objective itself.

Our explanation builds on a well-established property of modern self-supervised learning. Methods such as SimCLR \citep{chen_simple_2020}, DINO \citep{caron_emerging_2021}, and DINOv2 \citep{oquab_dinov2_2024} learn representations by enforcing consistency across different views of the same input. Empirically, these methods produce representations that transfer well across downstream tasks \citep{chen_simple_2020, caron_emerging_2021, oquab_dinov2_2024}. In some settings, the structure learned by these models is sufficiently organized that downstream concepts can be recovered directly from the representation. For example, \citet{zhu_interpretable_2025} show that class-specific prototypes can be obtained by clustering features from a frozen self-supervised vision model. This demonstrates that self-supervised learning can recover features aligned with downstream semantics without access to labels during training.

ProtoSSL applies this principle directly in prototype space. Instead of learning a global embedding and introducing prototypes afterward separately for each task, the model is trained so that each input is represented by its similarities to a shared prototype bank. The self-supervised objective enforces consistency of these activation patterns across views. This encourages each prototype to respond to patterns that recur across views, since only such patterns can be activated consistently. Importantly, this specialization emerges without reference to any label space.

The resulting prototype bank therefore encodes a set of reusable latent patterns that reflect the structure of the data. These patterns are not yet associated with specific downstream labels, but they form a basis from which label-relevant structure can be recovered. This leads to a concrete implication: if the prototype bank captures transferable structure, then it should already support downstream tasks before any label-aware alignment. We test this using two label-free projection settings (described in \autoref{app:alt_assign}): in \emph{project-in-pretrain} (PIP), prototypes are projected onto nearest neighbors in the pretraining dataset; in \emph{project-in-target} (PIT), projection is performed in the downstream dataset. We also test directly using unprojected, latent prototypes. In each of these cases, a classifier trained on frozen prototype activations achieves excellent performance across dataset scales (\autoref{tab:ecg_no_assign}), indicating that the pretrained prototypes encode structure that readily generalizes to downstream datasets.

However, this structure is not yet tied to any specific label, so these prototypes do not support case-based interpretability (see further discussion in \autoref{app:alt_assign}). ProtoSSL therefore introduces a second stage consisting of label-aware assignment followed by label-supervised projection. Assignment identifies which prototypes are most associated with each label, and projection anchors those prototypes to representative, class-consistent signal segments. This step is crucial to produce coherent, task-specific explanations.

Taken together, these observations support a two-stage view of interpretability in ProtoSSL. Self-supervised learning recovers a reusable set of latent patterns that reflect the structure of the data, and downstream supervision aligns and grounds those patterns for a specific task. Interpretability therefore arises from the combination of structure learned during pretraining and grounding through projection, rather than from label supervision during prototype learning itself.

\section{Extended Results}
\label{app:more_results}

We present further results from experiments studying ProtoSSL. In \autoref{app:blackbox}, we compare ProtoSSL against blackbox ECG models for predictive context across all six ECG datasets. In Supplementary Sections \ref{app:alt_assign}, \ref{app:supproto_heedb_pila}--\ref{app:heedb_150}, and \ref{app:num_proto}--\ref{app:classifier_explanation}, we conduct ablation experiments over the EchoNext ECG dataset given its large number of samples, moderate number of labels, label space distinctness from HEEDB, and strong performance of models from the literature.

\begin{table}[tp]
\centering
\caption{\textbf{Full ECG results presented in \autoref{fig:ecg_efficiency}.} Macro AUROC (bootstrapped 95\% CI), averaged across 5 randomly seeded replicates. Best model \textbf{bolded}, next best model \Xul{underlined}.}
\label{tab:ecg_full}
{
\setlength{\tabcolsep}{4pt}
\begin{adjustbox}{max width=\textwidth}

\begin{tabular}{lrccccc}
\toprule
\multirow{2}{*}[-2pt]{Dataset} & \multirow{2}{*}[-2pt]{\makecell{Train\\Size}} & \multicolumn{2}{c}{Tuned} & \multicolumn{2}{c}{Probed} & \multirow{2}{*}[-2pt]{SupProto Direct} \\
\cmidrule(lr){3-4}\cmidrule(lr){5-6}
 &  & ProtoSSL HEEDB & SupProto HEEDB & ProtoSSL HEEDB & SupProto HEEDB &  \\
\midrule
\multirow[t]{9}{*}{EchoNext} & 72475 & \textbf{0.819} [0.818-0.820] & 0.782 [0.780-0.783] & \Xul{0.810} [0.809-0.811] & 0.763 [0.762-0.764] & 0.805 [0.804-0.806] \\
 & 32768 & \textbf{0.814} [0.813-0.815] & 0.758 [0.757-0.759] & \Xul{0.806} [0.805-0.807] & 0.757 [0.756-0.758] & 0.797 [0.796-0.798] \\
 & 16384 & \textbf{0.798} [0.797-0.799] & 0.740 [0.739-0.742] & \textbf{0.798} [0.797-0.799] & 0.750 [0.749-0.751] & \Xul{0.781} [0.780-0.782] \\
 & 8192 & \textbf{0.790} [0.789-0.791] & 0.731 [0.729-0.732] & \textbf{0.790} [0.789-0.791] & 0.732 [0.731-0.733] & \Xul{0.762} [0.761-0.763] \\
 & 4096 & \textbf{0.782} [0.780-0.783] & 0.728 [0.727-0.729] & \Xul{0.765} [0.764-0.766] & 0.720 [0.719-0.721] & 0.738 [0.737-0.739] \\
 & 2048 & \textbf{0.779} [0.778-0.780] & 0.730 [0.729-0.731] & \Xul{0.747} [0.746-0.748] & 0.718 [0.717-0.719] & 0.722 [0.721-0.723] \\
 & 1024 & \textbf{0.768} [0.767-0.769] & 0.725 [0.724-0.726] & \Xul{0.729} [0.728-0.731] & 0.682 [0.681-0.684] & 0.716 [0.715-0.717] \\
 & 512 & \textbf{0.761} [0.760-0.763] & 0.711 [0.710-0.713] & \Xul{0.723} [0.722-0.725] & 0.660 [0.659-0.662] & 0.711 [0.710-0.712] \\
 & 256 & \textbf{0.739} [0.737-0.740] & 0.690 [0.689-0.691] & 0.701 [0.700-0.702] & 0.667 [0.665-0.668] & \Xul{0.702} [0.701-0.703] \\
\midrule
\multirow[t]{9}{*}{MIMIC-IV-ECG} & 78470 & \textbf{0.813} [0.811-0.814] & 0.788 [0.786-0.789] & \Xul{0.806} [0.804-0.807] & 0.774 [0.772-0.776] & 0.805 [0.804-0.806] \\
 & 32768 & \Xul{0.801} [0.799-0.802] & 0.751 [0.749-0.753] & \textbf{0.802} [0.800-0.804] & 0.762 [0.760-0.764] & 0.792 [0.791-0.794] \\
 & 16384 & \textbf{0.788} [0.786-0.789] & 0.737 [0.735-0.738] & \Xul{0.786} [0.784-0.788] & 0.749 [0.747-0.751] & 0.781 [0.779-0.782] \\
 & 8192 & \textbf{0.774} [0.772-0.775] & 0.721 [0.719-0.723] & \Xul{0.768} [0.766-0.770] & 0.732 [0.730-0.734] & 0.757 [0.755-0.759] \\
 & 4096 & \textbf{0.785} [0.784-0.787] & 0.701 [0.699-0.703] & \Xul{0.748} [0.746-0.749] & 0.716 [0.714-0.718] & 0.725 [0.723-0.727] \\
 & 2048 & \textbf{0.777} [0.776-0.778] & 0.697 [0.695-0.699] & 0.715 [0.713-0.717] & 0.697 [0.695-0.699] & \Xul{0.720} [0.718-0.722] \\
 & 1024 & \textbf{0.768} [0.766-0.769] & 0.686 [0.684-0.688] & \Xul{0.703} [0.702-0.705] & 0.685 [0.683-0.687] & 0.694 [0.692-0.696] \\
 & 512 & \textbf{0.747} [0.745-0.749] & \Xul{0.677} [0.675-0.679] & 0.676 [0.675-0.678] & 0.667 [0.665-0.669] & 0.657 [0.655-0.659] \\
 & 256 & \textbf{0.728} [0.726-0.729] & 0.659 [0.657-0.661] & \Xul{0.662} [0.660-0.664] & 0.648 [0.646-0.650] & 0.655 [0.653-0.657] \\
\midrule
\multirow[t]{6}{*}{ZZU pECG} & 8658 & \textbf{0.815} [0.813-0.817] & 0.751 [0.748-0.753] & 0.776 [0.775-0.778] & 0.696 [0.694-0.699] & \Xul{0.790} [0.788-0.792] \\
 & 4096 & \textbf{0.795} [0.792-0.797] & 0.724 [0.721-0.726] & \Xul{0.770} [0.768-0.772] & 0.685 [0.682-0.688] & 0.749 [0.746-0.751] \\
 & 2048 & \textbf{0.764} [0.762-0.767] & 0.709 [0.706-0.712] & \Xul{0.762} [0.760-0.764] & 0.664 [0.662-0.667] & 0.685 [0.682-0.688] \\
 & 1024 & \Xul{0.716} [0.713-0.718] & 0.665 [0.662-0.668] & \textbf{0.749} [0.747-0.751] & 0.667 [0.664-0.670] & 0.676 [0.673-0.679] \\
 & 512 & \Xul{0.718} [0.716-0.721] & 0.692 [0.689-0.694] & \textbf{0.751} [0.748-0.753] & 0.673 [0.671-0.676] & 0.700 [0.697-0.703] \\
 & 256 & \Xul{0.719} [0.716-0.721] & 0.676 [0.674-0.679] & \textbf{0.734} [0.732-0.736] & 0.662 [0.659-0.664] & 0.658 [0.655-0.661] \\
\midrule
\multirow[t]{7}{*}{PTB-XL} & 17418 & \textbf{0.911} [0.910-0.913] & 0.899 [0.897-0.901] & 0.881 [0.879-0.883] & 0.836 [0.834-0.838] & \Xul{0.900} [0.899-0.902] \\
 & 8722 & \textbf{0.895} [0.894-0.897] & \Xul{0.879} [0.877-0.881] & 0.870 [0.868-0.872] & 0.822 [0.820-0.825] & 0.877 [0.875-0.879] \\
 & 4356 & \textbf{0.879} [0.877-0.881] & 0.836 [0.833-0.839] & \Xul{0.857} [0.855-0.859] & 0.811 [0.808-0.814] & 0.846 [0.844-0.848] \\
 & 2175 & \textbf{0.865} [0.863-0.867] & 0.813 [0.810-0.816] & \Xul{0.844} [0.842-0.846] & 0.799 [0.796-0.801] & 0.809 [0.807-0.812] \\
 & 1091 & \textbf{0.840} [0.837-0.842] & 0.801 [0.799-0.804] & \Xul{0.819} [0.817-0.822] & 0.785 [0.782-0.787] & 0.778 [0.775-0.780] \\
 & 547 & \textbf{0.816} [0.813-0.818] & 0.771 [0.768-0.774] & \Xul{0.794} [0.792-0.797] & 0.770 [0.767-0.773] & 0.735 [0.732-0.738] \\
 & 273 & \textbf{0.768} [0.766-0.771] & 0.713 [0.710-0.716] & \Xul{0.752} [0.749-0.755] & 0.739 [0.736-0.742] & 0.618 [0.615-0.622] \\
\midrule
\multirow[t]{6}{*}{CinC Georgia} & 8192 & \textbf{0.883} [0.881-0.885] & 0.858 [0.855-0.860] & 0.845 [0.842-0.847] & 0.831 [0.829-0.834] & \Xul{0.862} [0.859-0.864] \\
 & 4096 & \textbf{0.864} [0.862-0.866] & 0.823 [0.821-0.826] & 0.822 [0.820-0.825] & 0.815 [0.813-0.818] & \Xul{0.828} [0.826-0.831] \\
 & 2048 & \textbf{0.827} [0.825-0.830] & 0.796 [0.793-0.799] & \Xul{0.808} [0.805-0.811] & 0.797 [0.795-0.800] & 0.791 [0.789-0.794] \\
 & 1024 & \textbf{0.820} [0.818-0.823] & 0.770 [0.767-0.773] & \Xul{0.786} [0.783-0.789] & 0.765 [0.762-0.767] & 0.744 [0.740-0.747] \\
 & 512 & \textbf{0.770} [0.767-0.773] & 0.736 [0.733-0.740] & \Xul{0.739} [0.736-0.743] & 0.726 [0.723-0.730] & 0.663 [0.659-0.666] \\
 & 256 & \textbf{0.727} [0.724-0.730] & \Xul{0.722} [0.719-0.725] & 0.709 [0.706-0.712] & 0.698 [0.695-0.701] & 0.643 [0.639-0.647] \\
\midrule
\multirow[t]{9}{*}{CODE-15\%} & 74112 & \textbf{0.966} [0.966-0.966] & 0.960 [0.960-0.961] & 0.949 [0.949-0.950] & 0.948 [0.948-0.948] & \Xul{0.962} [0.962-0.962] \\
 & 32768 & \textbf{0.963} [0.962-0.963] & 0.953 [0.953-0.953] & 0.946 [0.945-0.946] & 0.943 [0.943-0.944] & \Xul{0.958} [0.957-0.958] \\
 & 16384 & \textbf{0.956} [0.956-0.956] & 0.946 [0.946-0.946] & 0.937 [0.937-0.937] & 0.940 [0.940-0.941] & \Xul{0.952} [0.952-0.952] \\
 & 8192 & \textbf{0.946} [0.945-0.946] & \Xul{0.940} [0.940-0.940] & 0.928 [0.928-0.928] & 0.937 [0.937-0.937] & 0.938 [0.938-0.938] \\
 & 4096 & \textbf{0.933} [0.933-0.933] & \Xul{0.931} [0.930-0.931] & 0.917 [0.917-0.917] & 0.929 [0.929-0.929] & 0.903 [0.903-0.904] \\
 & 2048 & 0.917 [0.916-0.917] & \Xul{0.918} [0.918-0.919] & 0.905 [0.905-0.906] & \textbf{0.928} [0.928-0.928] & 0.825 [0.824-0.825] \\
 & 1024 & 0.892 [0.892-0.892] & 0.897 [0.897-0.898] & \Xul{0.900} [0.900-0.900] & \textbf{0.918} [0.918-0.919] & 0.754 [0.753-0.754] \\
 & 512 & 0.853 [0.853-0.854] & \Xul{0.871} [0.870-0.871] & 0.854 [0.853-0.854] & \textbf{0.911} [0.911-0.911] & 0.694 [0.693-0.695] \\
 & 256 & 0.815 [0.815-0.816] & \Xul{0.868} [0.868-0.869] & 0.790 [0.790-0.791] & \textbf{0.897} [0.897-0.898] & 0.673 [0.672-0.673] \\
\bottomrule
\end{tabular}

\end{adjustbox}
}
\end{table}


\subsection{Contextual Comparison to Blackbox ECG Models}
\label{app:blackbox}

Our primary comparisons are to projection-based prototype models, as these baselines share ProtoSSL's case-based prediction mechanism and provide projection-grounded explanations. Nevertheless, we include \autoref{tab:ecg_blackbox} to contextualize the predictive tradeoff relative to strong blackbox ECG models. Specifically, we compare frozen probing of ProtoSSL HEEDB against two ECG foundation models identified as strong in a recent benchmark by \citet{al-masud_benchmarking_2026}: ECGFounder, a label-supervised ECG foundation model trained on HEEDB \citep{li_electrocardiogram_2025}, and ST-MEM, a strong self-supervised ECG model \citep{na_guiding_2023}. Additionally, we compare ProtoSSL HEEDB to two matched encoder architectures, either trained directly on the downstream classification task (Blackbox Direct) or probed after PCLR contrastive pretraining over HEEDB (essentially removing ProtoSSL HEEDB's prototype requirement). None of these models mechanistically provide explanations for their predictions, but they provide useful reference points for assessing whether ProtoSSL sacrifices predictive performance to obtain interpretability.

\begin{table}[tp]
\centering
\caption{\textbf{ProtoSSL compared to blackbox models across all data scales.} ProtoSSL HEEDB, PCLR HEEDB, ECGFounder, and ST-MEM are used as frozen encoders and probed; Blackbox Direct is trained end-to-end using a random initialization of the same backbone as ProtoSSL. Macro AUROC (bootstrapped 95\% CI), averaged across 5 randomly seeded replicates. Best model \textbf{bolded}, next best model \Xul{underlined}.}
\label{tab:ecg_blackbox}
{
\setlength{\tabcolsep}{4pt}
\begin{adjustbox}{max width=\textwidth}

\begin{tabular}{lrccccc}
\toprule
\multirow{2}{*}[-2pt]{Dataset} & \multirow{2}{*}[-2pt]{\makecell{Train\\Size}} & \multicolumn{4}{c}{Probed} & \multirow{2}{*}[-2pt]{Blackbox Direct} \\
\cmidrule{3-6}
& & ProtoSSL HEEDB & PCLR \citep{diamant_patient_2022} HEEDB & ECGFounder \citep{li_electrocardiogram_2025} & ST-MEM \citep{na_guiding_2023} & \\
\midrule
\multirow[t]{9}{*}{EchoNext} & 72475 & 0.810 [0.809-0.811] & \textbf{0.824} [0.823-0.825] & 0.802 [0.801-0.803] & \Xul{0.813} [0.812-0.814] & 0.783 [0.782-0.784] \\
 & 32768 & 0.806 [0.805-0.807] & \textbf{0.824} [0.823-0.825] & 0.798 [0.797-0.799] & \Xul{0.808} [0.807-0.809] & 0.782 [0.781-0.783] \\
 & 16384 & 0.798 [0.797-0.799] & \textbf{0.820} [0.819-0.821] & 0.791 [0.790-0.792] & \Xul{0.800} [0.799-0.801] & 0.782 [0.781-0.784] \\
 & 8192 & 0.790 [0.789-0.791] & \textbf{0.814} [0.813-0.815] & 0.781 [0.780-0.782] & \Xul{0.792} [0.791-0.793] & 0.768 [0.767-0.770] \\
 & 4096 & 0.765 [0.764-0.766] & \textbf{0.802} [0.801-0.803] & 0.766 [0.765-0.767] & \Xul{0.773} [0.772-0.774] & 0.749 [0.747-0.750] \\
 & 2048 & 0.747 [0.746-0.748] & \textbf{0.789} [0.787-0.790] & 0.753 [0.752-0.754] & \Xul{0.755} [0.754-0.756] & 0.732 [0.731-0.733] \\
 & 1024 & 0.729 [0.728-0.731] & \textbf{0.772} [0.771-0.774] & \Xul{0.736} [0.735-0.737] & 0.726 [0.724-0.727] & 0.714 [0.713-0.715] \\
 & 512 & \Xul{0.723} [0.722-0.725] & \textbf{0.745} [0.743-0.746] & 0.720 [0.719-0.721] & 0.713 [0.712-0.715] & 0.710 [0.709-0.711] \\
 & 256 & \Xul{0.701} [0.700-0.702] & \textbf{0.719} [0.717-0.720] & 0.700 [0.698-0.701] & 0.693 [0.692-0.695] & 0.667 [0.666-0.668] \\
\midrule
\multirow[t]{9}{*}{MIMIC-IV-ECG} & 78470 & \Xul{0.806} [0.804-0.807] & 0.805 [0.803-0.806] & 0.801 [0.800-0.803] & \textbf{0.809} [0.808-0.811] & 0.802 [0.800-0.803] \\
 & 32768 & \Xul{0.802} [0.800-0.804] & 0.795 [0.794-0.797] & 0.800 [0.798-0.801] & \textbf{0.803} [0.801-0.804] & 0.792 [0.791-0.794] \\
 & 16384 & 0.786 [0.784-0.788] & \Xul{0.787} [0.785-0.788] & 0.780 [0.778-0.781] & \textbf{0.797} [0.795-0.798] & 0.773 [0.771-0.774] \\
 & 8192 & \Xul{0.768} [0.766-0.770] & 0.767 [0.765-0.769] & 0.757 [0.756-0.759] & \textbf{0.780} [0.778-0.782] & 0.767 [0.765-0.769] \\
 & 4096 & 0.748 [0.746-0.749] & 0.744 [0.743-0.746] & 0.743 [0.741-0.745] & \textbf{0.765} [0.763-0.767] & \Xul{0.750} [0.748-0.751] \\
 & 2048 & 0.715 [0.713-0.717] & 0.725 [0.723-0.726] & 0.723 [0.722-0.725] & \textbf{0.755} [0.753-0.756] & \Xul{0.747} [0.746-0.749] \\
 & 1024 & 0.703 [0.702-0.705] & 0.713 [0.711-0.715] & 0.717 [0.715-0.719] & \textbf{0.734} [0.732-0.735] & \Xul{0.722} [0.721-0.724] \\
 & 512 & 0.676 [0.675-0.678] & 0.687 [0.686-0.689] & 0.701 [0.699-0.703] & \Xul{0.713} [0.711-0.715] & \textbf{0.714} [0.712-0.715] \\
 & 256 & 0.662 [0.660-0.664] & 0.668 [0.667-0.670] & 0.683 [0.681-0.684] & \textbf{0.698} [0.696-0.700] & \Xul{0.685} [0.683-0.687] \\
\midrule
\multirow[t]{6}{*}{ZZU pECG} & 8658 & 0.776 [0.775-0.778] & 0.810 [0.809-0.812] & \textbf{0.837} [0.836-0.839] & \Xul{0.834} [0.832-0.836] & 0.780 [0.778-0.783] \\
 & 4096 & 0.770 [0.768-0.772] & 0.805 [0.804-0.807] & \textbf{0.816} [0.814-0.818] & \Xul{0.814} [0.812-0.816] & 0.754 [0.751-0.756] \\
 & 2048 & 0.762 [0.760-0.764] & \Xul{0.793} [0.792-0.795] & \textbf{0.796} [0.793-0.798] & \Xul{0.793} [0.791-0.795] & 0.707 [0.704-0.710] \\
 & 1024 & 0.749 [0.747-0.751] & \Xul{0.765} [0.763-0.767] & 0.764 [0.761-0.767] & \textbf{0.768} [0.766-0.770] & 0.669 [0.667-0.672] \\
 & 512 & \Xul{0.751} [0.748-0.753] & \textbf{0.756} [0.754-0.758] & 0.749 [0.746-0.751] & 0.745 [0.743-0.747] & 0.674 [0.671-0.677] \\
 & 256 & \Xul{0.734} [0.732-0.736] & \textbf{0.739} [0.736-0.741] & 0.732 [0.730-0.735] & 0.715 [0.713-0.717] & 0.621 [0.618-0.623] \\
\midrule
\multirow[t]{7}{*}{PTB-XL} & 17418 & 0.881 [0.879-0.883] & 0.878 [0.876-0.880] & \textbf{0.930} [0.929-0.931] & 0.896 [0.895-0.898] & \Xul{0.922} [0.920-0.923] \\
 & 8722 & 0.870 [0.868-0.872] & 0.865 [0.863-0.867] & \textbf{0.920} [0.919-0.922] & 0.875 [0.873-0.877] & \Xul{0.904} [0.902-0.905] \\
 & 4356 & 0.857 [0.855-0.859] & 0.843 [0.841-0.846] & \textbf{0.906} [0.904-0.907] & 0.862 [0.860-0.864] & \Xul{0.874} [0.872-0.876] \\
 & 2175 & 0.844 [0.842-0.846] & 0.827 [0.825-0.830] & \textbf{0.895} [0.893-0.896] & 0.842 [0.840-0.845] & \Xul{0.850} [0.848-0.852] \\
 & 1091 & 0.819 [0.817-0.822] & 0.803 [0.801-0.805] & \textbf{0.877} [0.875-0.879] & \Xul{0.824} [0.822-0.827] & 0.801 [0.799-0.803] \\
 & 547 & \Xul{0.794} [0.792-0.797] & 0.777 [0.774-0.779] & \textbf{0.857} [0.855-0.859] & 0.790 [0.787-0.792] & 0.755 [0.752-0.757] \\
 & 273 & \Xul{0.752} [0.749-0.755] & 0.737 [0.734-0.740] & \textbf{0.824} [0.821-0.826] & 0.746 [0.743-0.749] & 0.715 [0.713-0.718] \\
\midrule
\multirow[t]{6}{*}{CinC Georgia} & 8192 & 0.845 [0.842-0.847] & 0.851 [0.849-0.854] & \textbf{0.915} [0.913-0.917] & 0.843 [0.841-0.845] & \Xul{0.875} [0.873-0.877] \\
 & 4096 & 0.822 [0.820-0.825] & 0.830 [0.828-0.833] & \textbf{0.912} [0.910-0.913] & \Xul{0.842} [0.840-0.844] & 0.837 [0.834-0.840] \\
 & 2048 & 0.808 [0.805-0.811] & 0.814 [0.811-0.816] & \textbf{0.906} [0.904-0.908] & \Xul{0.833} [0.830-0.835] & 0.805 [0.803-0.808] \\
 & 1024 & 0.786 [0.783-0.789] & 0.799 [0.796-0.802] & \textbf{0.890} [0.888-0.892] & \Xul{0.806} [0.803-0.809] & 0.740 [0.736-0.743] \\
 & 512 & 0.739 [0.736-0.743] & 0.759 [0.756-0.762] & \textbf{0.863} [0.861-0.865] & \Xul{0.776} [0.773-0.778] & 0.694 [0.691-0.698] \\
 & 256 & 0.709 [0.706-0.712] & 0.729 [0.726-0.732] & \textbf{0.839} [0.836-0.841] & \Xul{0.755} [0.753-0.758] & 0.605 [0.601-0.608] \\
\midrule
\multirow[t]{9}{*}{CODE-15\%} & 74112 & 0.949 [0.949-0.950] & \Xul{0.964} [0.964-0.964] & \Xul{0.964} [0.964-0.964] & \textbf{0.970} [0.970-0.970] & \Xul{0.964} [0.963-0.964] \\
 & 32768 & 0.946 [0.945-0.946] & 0.961 [0.961-0.961] & \Xul{0.962} [0.962-0.962] & \textbf{0.969} [0.968-0.969] & 0.957 [0.957-0.957] \\
 & 16384 & 0.937 [0.937-0.937] & 0.957 [0.957-0.957] & \Xul{0.959} [0.959-0.960] & \textbf{0.967} [0.966-0.967] & 0.949 [0.948-0.949] \\
 & 8192 & 0.928 [0.928-0.928] & 0.951 [0.951-0.951] & \Xul{0.957} [0.956-0.957] & \textbf{0.965} [0.965-0.965] & 0.909 [0.909-0.909] \\
 & 4096 & 0.917 [0.917-0.917] & 0.943 [0.942-0.943] & \Xul{0.952} [0.952-0.952] & \textbf{0.961} [0.961-0.961] & 0.815 [0.815-0.816] \\
 & 2048 & 0.905 [0.905-0.906] & 0.937 [0.937-0.937] & \Xul{0.944} [0.944-0.945] & \textbf{0.959} [0.959-0.959] & 0.744 [0.744-0.745] \\
 & 1024 & 0.900 [0.900-0.900] & 0.923 [0.923-0.924] & \Xul{0.929} [0.928-0.929] & \textbf{0.956} [0.956-0.956] & 0.664 [0.663-0.665] \\
 & 512 & 0.854 [0.853-0.854] & 0.909 [0.908-0.909] & \Xul{0.924} [0.924-0.924] & \textbf{0.950} [0.950-0.950] & 0.602 [0.601-0.603] \\
 & 256 & 0.790 [0.790-0.791] & 0.887 [0.887-0.888] & \Xul{0.919} [0.918-0.919] & \textbf{0.943} [0.943-0.943] & 0.573 [0.572-0.574] \\
\bottomrule
\end{tabular}

\end{adjustbox}
}
\end{table}

ProtoSSL remains competitive with blackbox ECG models in several settings, although the relative ordering is task- and scale-dependent. On EchoNext, ProtoSSL is consistently better than Blackbox Direct and close to both foundation models: ST-MEM is slightly higher at larger training set sizes, ECGFounder is comparable at intermediate sizes, and ProtoSSL exceeds both at the two smallest training sizes. On MIMIC-IV-ECG, ProtoSSL is close to ST-MEM and stronger than ECGFounder and Blackbox Direct at full scale, but ST-MEM is the strongest probe across data scales and the advantage of the other blackbox references increases in the smallest-label regimes. On ZZU pECG, ECGFounder and ST-MEM are stronger at full and intermediate training sizes, but the gap narrows as labels decrease and ProtoSSL exceeds both at the two smallest training sizes; it also exceeds Blackbox Direct at every reduced training size.

The three standard ECG interpretation datasets show clearer advantages for the pretrained blackbox foundation models. On PTB-XL, ECGFounder is strongest across all data scales, while ProtoSSL is generally close to ST-MEM and exceeds Blackbox Direct in the smallest training regimes. On CinC Georgia, ECGFounder is substantially stronger across all scales, while ProtoSSL is generally below ST-MEM but exceeds Blackbox Direct at smaller scales. On CODE-15\%, ST-MEM and ECGFounder are consistently stronger than ProtoSSL, particularly in low-label settings; however, ProtoSSL remains substantially stronger than Blackbox Direct once the training set is reduced to 8K examples or fewer.


PCLR HEEDB shares ProtoSSL HEEDB's encoder, pretraining data, and contrastive objective but has no prototype layer, so it isolates the predictive cost of the prototype bottleneck. With our default 14 prototypes per label, frozen ProtoSSL HEEDB is below PCLR HEEDB at every training size on EchoNext, ZZU pECG, CinC Georgia, and CODE-15\%; ProtoSSL HEEDB exceeds PCLR HEEDB at every training size on PTB-XL. Notably, on EchoNext, the gap may partially be recovered by expanding the prototype budget: probing ProtoSSL HEEDB with 83 prototypes per label (\autoref{tab:ecg_no_assign}) matches PCLR HEEDB at 8K training samples and above (0.824 vs.\ 0.824 at full scale), although PCLR HEEDB retains an advantage below 8K. PCLR HEEDB is itself not uniformly strongest: it leads on EchoNext but trails ECGFounder or ST-MEM on the other datasets at most scales, so contrastive pretraining alone does not yield the most transferable structure across diverse datasets.

Taken together, \autoref{tab:ecg_blackbox} shows that frozen ProtoSSL is not consistently below strong blackbox ECG models. It is close to or stronger than foundation-model probes on some tasks and data regimes, behind them in others, and often more label-efficient than training the same backbone as a blackbox model from scratch. Relative to its matched unconstrained encoder, the prototype bottleneck carries a cost that is small at most scales and that a larger prototype budget can close. Thus, the blackbox results indicate that ProtoSSL preserves much of the predictive utility of strong unrestricted ECG encoders while supporting case-based interpretability through projected, label-aligned ECG exemplars.


\subsection{Varying and Ablating the Assignment of Pretrained Prototypes}
\label{app:alt_assign}

\begin{table}[tp]
\centering
\caption{\textbf{Variation of ProtoSSL's prototype-label assignment mechanism over EchoNext.} With fine-tuning, LAP is close in performance to Pool. LAP: assignment via reduction to linear assignment problem (ours), followed by label-guided projection; Pool: assignment via adaptation of ProtoPool \citep{rymarczyk_interpretable_2022}, followed by label-guided projection. Macro AUROC (bootstrapped 95\% CI), averaged across 5 randomly seeded replicates. Best model \textbf{bolded}, next best model \Xul{underlined}.}
\label{tab:ecg_alt_assign}
{
\setlength{\tabcolsep}{4pt}
\begin{adjustbox}{max width=\textwidth}

\begin{tabular}{rcccc}
\toprule
\multirow{3}{*}[-4pt]{\makecell[l]{EchoNext\\Train Size}} & \multicolumn{4}{c}{ProtoSSL HEEDB Assignment Strategy} \\
\cmidrule(lr){2-5}
& \multicolumn{2}{c}{Tuned} & \multicolumn{2}{c}{Probed} \\
\cmidrule(lr){2-3}\cmidrule(lr){4-5}
& LAP & Pool & LAP & Pool \\
\midrule
72475 & \Xul{0.819} [0.818-0.820] & 0.818 [0.817-0.819] & 0.810 [0.809-0.811] & \textbf{0.820} [0.819-0.821] \\
32768 & \Xul{0.814} [0.813-0.815] & 0.802 [0.801-0.803] & 0.806 [0.805-0.807] & \textbf{0.816} [0.815-0.817] \\
16384 & \Xul{0.798} [0.797-0.799] & 0.795 [0.794-0.796] & \Xul{0.798} [0.797-0.799] & \textbf{0.809} [0.808-0.810] \\
8192 & \Xul{0.790} [0.789-0.791] & 0.784 [0.783-0.785] & \Xul{0.790} [0.789-0.791] & \textbf{0.803} [0.803-0.804] \\
4096 & \Xul{0.782} [0.780-0.783] & 0.775 [0.774-0.776] & 0.765 [0.764-0.766] & \textbf{0.783} [0.782-0.784] \\
2048 & \textbf{0.779} [0.778-0.780] & \Xul{0.772} [0.771-0.773] & 0.747 [0.746-0.748] & 0.765 [0.763-0.766] \\
1024 & \Xul{0.768} [0.767-0.769] & \textbf{0.771} [0.770-0.772] & 0.729 [0.728-0.731] & 0.746 [0.744-0.747] \\
512 & \textbf{0.761} [0.760-0.763] & \Xul{0.759} [0.758-0.760] & 0.723 [0.722-0.725] & 0.729 [0.728-0.730] \\
256 & \Xul{0.739} [0.737-0.740] & \textbf{0.742} [0.741-0.744] & 0.701 [0.700-0.702] & 0.702 [0.701-0.703] \\
\bottomrule
\end{tabular}

\end{adjustbox}
}
\end{table}
\begin{table}[tp]
\centering
\caption{\textbf{Runtime comparison of prototype-label assignment mechanism over varying EchoNext dataset scales.} LAP is consistently an order of magnitude faster than Pool. While the specific runtime will vary depending on hardware, we note that we leveraged relatively moderate compute resources in our experiments (see \autoref{app:training}). LAP: reduction to linear assignment problem (ours); Pool: adaptation of ProtoPool \citep{rymarczyk_interpretable_2022}. Runtime in seconds averaged across 5 randomly seeded replicates ± 1 std dev.}
\label{tab:assignment_runtime}
{
\setlength{\tabcolsep}{4pt}
\begin{adjustbox}{max width=\textwidth}

\begin{tabular}{lrrrrrrrrr}
\toprule
\multirow{2}{*}[-2pt]{\makecell[l]{Assignment\\Method}} & \multicolumn{9}{c}{EchoNext Train Subset Size} \\
\cmidrule{2-10}
& 72475 & 32768 & 16384 & 8192 & 4096 & 2048 & 1024 & 512 & 256 \\
\midrule
LAP & 70.2 ± 1.6 & 37.8 ± 0.8 & 25.2 ± 1.1 & 20.6 ± 0.5 & 17.6 ± 0.5 & 16.0 ± 0.0 & 15.0 ± 0.0 & 14.2 ± 0.4 & 14.0 ± 0.0 \\
Pool & 599.4 ± 66.9 & 446.0 ± 42.2 & 412.4 ± 70.3 & 450.0 ± 33.4 & 539.2 ± 45.7 & 679.0 ± 56.8 & 844.2 ± 124.9 & 1049.8 ± 225.5 & 1063.2 ± 179.2 \\
\bottomrule
\end{tabular}

\end{adjustbox}
}
\end{table}

\paragraph{Ablation of assignment mechanism:} By first using self-supervised pretraining to discover data motifs (prototypes), we introduce the need to do label alignment when adapting prototypes to downstream tasks. This critically preserves the intrinsic interpretability mechanism of prototype-based models via label-supervised projection (see \autoref{sec:projection}), enabling \emph{this} looks like \emph{that} predictions. We introduce a novel and efficient mechanism to do such alignment (\autoref{sec:assign_main}). However, we note that one prior work from the literature may be adapted to perform this assignment step: ProtoPool \citep{rymarczyk_interpretable_2022}. Specifically, while holding prototypes and the encoder frozen, we construct an intermediate mapping layer between the pool of pretrained prototypes and \(M\) prototype slots per label for \(L\) labels. The weights of this layer govern the assignment of prototypes to different labels and are optimized under in-label prototype orthogonality and cross-entropy objectives (where the cross-entropy term is computed using logits output by a final classifier layer over assigned prototype activations). Following assignment with this adaptation of the ProtoPool method, we apply the same (optional) fine-tuning, label-supervised projection, and classifier training steps as in ProtoSSL; we refer to these results in \autoref{tab:ecg_alt_assign} as Pool. We find that on EchoNext, Pool-based assignment results in moderate gains over LAP-based assignment under frozen probing when data is abundant (however, \emph{we note that the Pool results are not strictly probed, as the Pool-based assignment is parametric}). Moreover, when assigned prototypes are fine-tuned, LAP and Pool are comparable in performance.

We note further that the ProtoPool-adapted assignment mechanism is an order of magnitude slower than ProtoSSL's reduction to the linear assignment problem over EchoNext (\autoref{tab:assignment_runtime}).
While the specific runtimes will vary depending on hardware and hyperparameters (see \autoref{app:training} for the settings of our experiments), we note the relative benefit of LAP in terms of computational efficiency.

\begin{table}[tp]
\centering
\caption{\textbf{Ablation of ProtoSSL's prototype-label assignment mechanism over EchoNext, where we perform no downstream assignment and instead do unguided projection (PIT and PIP) or no projection at all (Latent).} Importantly, with no label assignment, the prototype activation vectors input into the final classifier have length \(K = 1000\) (i.e. the number of prototypes in the pretrained pool), whereas assigned-then-projected-prototype activations (as in LAP) have length \(M \times L\). To reduce confounding with the length of the prototype activation vectors, we use \(M=83\) prototypes per label in this experiment over EchoNext's \(L=12\) labels. As assignment is required to fine-tune under label-supervision, we only report frozen-probing results here. Under these settings, we find that LAP performs comparably to PIT, PIP, and Latent from 4K samples and above. \textbf{Note:} PIT and PIP invalidate prototype interpretability, as projected exemplars may not demonstrate the predicted label. Latent is also not interpretable as learned prototypes are never projected onto real cases. LAP: assignment via reduction to linear assignment problem (ours), followed by label-guided projection; PIT: label-unguided projection onto target dataset samples; PIP: label-unguided projection onto pretraining dataset samples. Latent: unprojected, latent prototypes learned during self supervised pretraining. Macro AUROC (bootstrapped 95\% CI), averaged across 5 randomly seeded replicates. Best model \textbf{bolded}, next best model \Xul{underlined}.}
\label{tab:ecg_no_assign}
{
\setlength{\tabcolsep}{4pt}
\begin{adjustbox}{max width=\textwidth}

\begin{tabular}{rcccc}
\toprule
\multirow{3}{*}[-4pt]{\makecell[l]{EchoNext\\Train Size}} & \multicolumn{4}{c}{ProtoSSL HEEDB Assignment Strategy} \\
\cmidrule(lr){2-5}
& Assigned & \multicolumn{3}{c}{Unassigned} \\
\cmidrule(lr){2-2}\cmidrule(lr){3-5}
& LAP & PIT & PIP & Latent \\
\midrule
72475 & 0.824 [0.823-0.825] & \Xul{0.826} [0.825-0.827] & \textbf{0.827} [0.826-0.828] & 0.823 [0.822-0.824] \\
32768 & 0.824 [0.823-0.825] & \Xul{0.826} [0.825-0.826] & \textbf{0.827} [0.826-0.828] & 0.824 [0.823-0.825] \\
16384 & 0.820 [0.819-0.821] & \Xul{0.822} [0.821-0.823] & \textbf{0.824} [0.823-0.825] & \Xul{0.822} [0.821-0.823] \\
8192 & \Xul{0.814} [0.813-0.815] & \textbf{0.817} [0.816-0.818] & \textbf{0.817} [0.816-0.818] & 0.812 [0.811-0.813] \\
4096 & 0.797 [0.796-0.798] & \Xul{0.800} [0.799-0.801] & \textbf{0.802} [0.800-0.803] & \Xul{0.800} [0.799-0.801] \\
2048 & 0.781 [0.780-0.782] & \textbf{0.791} [0.790-0.792] & \Xul{0.790} [0.789-0.791] & 0.786 [0.784-0.787] \\
1024 & 0.761 [0.760-0.762] & \Xul{0.774} [0.773-0.776] & \textbf{0.775} [0.774-0.776] & 0.772 [0.771-0.773] \\
512 & 0.741 [0.739-0.742] & \Xul{0.757} [0.756-0.758] & \textbf{0.758} [0.757-0.760] & 0.754 [0.753-0.755] \\
256 & 0.720 [0.719-0.721] & \textbf{0.741} [0.740-0.743] & \textbf{0.741} [0.740-0.742] & \Xul{0.731} [0.730-0.732] \\
\bottomrule
\end{tabular}

\end{adjustbox}
}
\end{table}

\paragraph{Unguided or no projection:} While ProtoPool presents an alternate, less efficient assignment mechanism, we additionally conduct two experiments entirely removing the assignment mechanism: Project-in-Target (PIT) and Project-in-Pretrain (PIP). In these experiments, we do not assign the pretrained prototypes to any target labels. Instead, we simply project the \(K\) learned prototypes onto the most similar training example, regardless of the labels of those training examples (as opposed to requiring projected examples to have a specific label under our default label-supervised projection, detailed in \autoref{sec:projection}). In PIT, the projected samples come from the downstream, target dataset; in PIP, the projected samples come from the pretraining dataset. We additionally directly assess the latent prototypes (Latent) learned in ProtoSSL's pretraining stage.

The classifiers are then trained over the \(K\) projected-prototype activations. As PIT, PIP, and Latent-derived prototype activations have length \(K=1000\), to better isolate the effect of our assignment step and reduce confounding with incomparable feature sizes, we use \(M=83\) prototypes per label for assignment with LAP. For this experiment over EchoNext with \(L=12\) labels, this results in prototype activations of length \(M \times L = 996\). As the prototypes remain unassociated with any label, we cannot do label-supervised fine-tuning over PIT, PIP, or Latent derived models, so we present only frozen-probing results over these methods in \autoref{tab:ecg_no_assign}.

We note that, without assignment, \emph{PIT and PIP invalidate the interpretability of the prototype-based model} as the exemplars may not actually demonstrate the positive label of the predicted class. This is particularly true in PIP if the pretraining dataset is unlabeled or does not share the same label space as the downstream task. Latent is also not case-based interpretable as its prototypes are never fixed to real samples. As such, these experiments interrogate a key aspect of prototype-based models: when prototype projection is purely data-driven and governed by fewer prototype-interpretability constraints, or when prototypes are allowed to be arbitrary latent representations, are resulting classifications more accurate? We find that LAP-assigned-then-projected (under label-supervision) prototype models are comparable to PIT, PIP, and Latent at 4K samples and above (\autoref{tab:ecg_no_assign}). At 2K samples and below, we hypothesize that label-supervised projection is constrained by sparse exemplars to choose from for any given rare label, as opposed to unguided projection where exemplars need not demonstrate any specific positive class (we discuss this limitation in \autoref{app:limitations}). However, \ul{these results suggest that the pretrained, self-supervised prototypes already encode downstream dataset structure without requiring label semantics}, supporting the discussion in \autoref{app:ssl_interpretability}.


\subsection{Pretrained latent prototype coherence across downstream datasets}
\label{app:proto_coherence}

One of our core findings is that ProtoSSL learns foundational latent prototypes that may be reused across downstream tasks after training-free assignment, either under projection with probing or as initialization for fine tuning. Beyond the empirical demonstration of strong classification results (\autoref{sec:generalization}), label efficiency (\autoref{sec:label_efficiency}), and human interpretability (\autoref{sec:user_study_results}), we study the specific claim that the latent prototypes capture foundational representation by examining the prototypes' coherence across pretraining and downstream datasets. To that end, we examine prototypes for ECG interpretation labels because 1) our pretraining dataset labels fall under this category, 2) these labels are shared with three downstream datasets, and 3) it is unknown how to manually verify model learned patterns for non-ECG interpretation tasks.

Briefly, we reestablish our label assignment setting (detailed in \autoref{sec:assign_main}). Given 1000 pretrained latent prototypes in ProtoSSL HEEDB, we assign 14 prototypes per downstream label (as described in \autoref{app:ecg_details}). Importantly, \emph{prototype-to-label assignments across downstream datasets are independent}. Thus, a latent prototype which is assigned to the same label across multiple datasets must strongly be associated with that label (and not other labels), precisely the semantic we aim to achieve through our novel assignment mechanism.

\begin{figure*}[tp]
    \centering
    \includegraphics[width=5.5in]{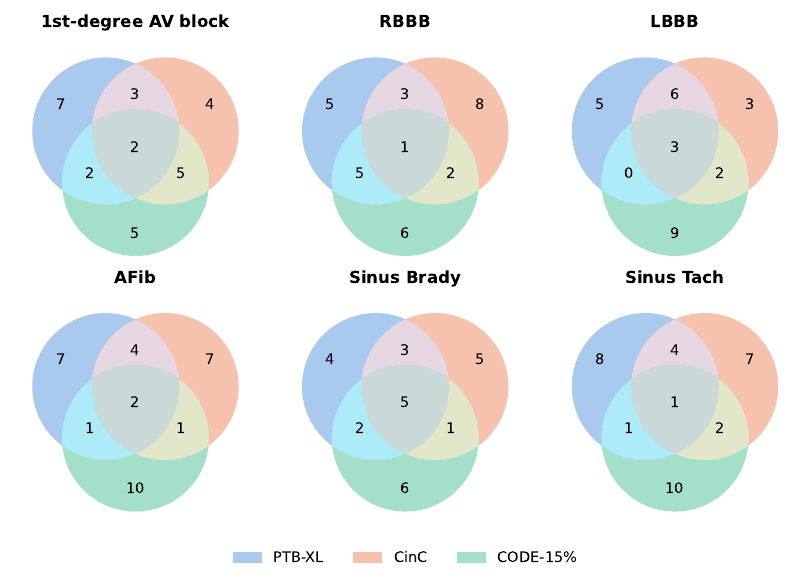}
    \caption{\textbf{The same ProtoSSL pretrained latent prototypes recur across downstream datasets}. Prototypes are learned once, without labels, during pretraining on HEEDB. For each diagnostic label, each downstream dataset (PTB-XL, CinC, CODE-15\%) \emph{independently} selects 14 out of 1000 pretrained prototypes to represent that label. Each circle is one dataset's selection of 14 prototypes, and each number counts the prototypes in that region; the center region counts prototypes that all three datasets chose for the label. An overlap means that datasets with distinct patients, recording setups, and label definitions independently assigned the same latent prototype for the same diagnosis. Presented prototype assignments were conducted under a single experiment seed.}
    \label{fig:venn}
\end{figure*}

We analyze six ECG interpretation labels (1st degree AV block, RBBB, LBBB, AFib, Sinus Bradycardia, and Sinus Tachycardia) across three downstream datasets (PTB-XL, CinC-Georgia, and CODE-15\%). For a single experiment seed, we examine the number of times a latent prototype recurs across the datasets. We present a Venn diagram in \autoref{fig:venn} quantifying the latent prototype recurrence; we find that many prototype-label assignments are shared between two or more datasets, with at least one assignment shared across all three downstream datasets. This suggests that these latent prototypes are capturing an underlying morphological pattern that describes each of these labels.

\begin{figure*}[tp]
    \centering
    \includegraphics[width=0.85\linewidth]{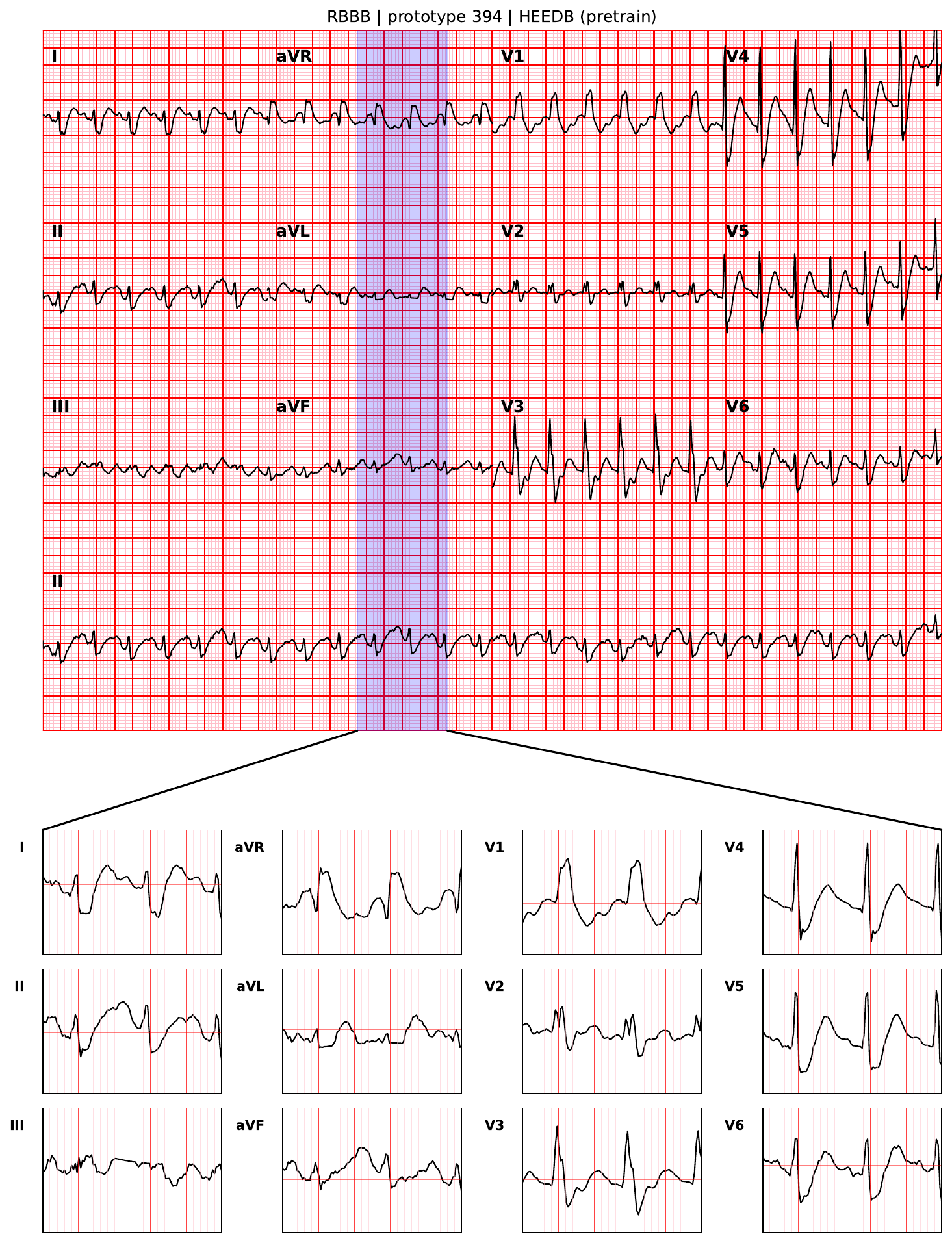}
    \caption{\textbf{Unguided, label-free projection of a recurrent prototype in the pretraining dataset recovers the downstream assigned label.} The RBBB-assigned prototype shared by all three datasets in \protect\autoref{fig:venn} (ProtoSSL HEEDB latent prototype 394) is projected onto the nearest HEEDB pretraining sample in embedding space. Despite unguided, label-free projection, the sample has a ground-truth label for RBBB and demonstrates the characteristic morphology for RBBB. Full 12-lead, 10-second ECG rendered in diagnostic layout with callout for precise 1-second highlighted window captured by the prototype.}
    \label{fig:heedb_rbbb}
\end{figure*}

To further confirm this finding, we present a case study (\autoref{fig:heedb_rbbb}) on the RBBB-assigned latent prototype shared by all downstream datasets (prototype 394). Remarkably, when we do unguided (label-free) projection of this latent prototype onto the HEEDB pretraining dataset (i.e. taking the sample closest in embedding space without requiring any label presence), we find that the projection recovers a sample with a ground-truth label of RBBB. Examining the morphology captured by the projected prototype, we observe the characteristic RSR' pattern in leads V1-3 and slurred S wave in the lateral leads. \emph{Together, these results demonstrate that ProtoSSL learns coherent foundational prototypes that support reuse in downstream applications}. Nonetheless, we acknowledge that this is a single case study and does not demonstrate downstream model interpretability, as in our user study in \autoref{sec:user_study_results}.


\subsection{Label-Supervised Pretraining without Projection}
\label{app:supproto_heedb_pila}

\begin{table}[tp]
\centering
\caption{\textbf{We primarily investigate whether a traditional, label-supervised prototype-based model which includes projection in its source dataset can generalize to downstream tasks (SupProto HEEDB).} However, we present here a variation where instead the latent prototypes learned under label-supervision are transferred using the same techniques (SupProto HEEDB (NoProj)). While SupProto HEEDB (NoProj) performs better than SupProto HEEDB, ProtoSSL is still superior across both fine-tuned and probed settings. Best model \textbf{bolded}, next best model \Xul{underlined}.}
\label{tab:supproto_noproj}
{
\setlength{\tabcolsep}{4pt}
\begin{adjustbox}{max width=\textwidth}

\begin{tabular}{rcccccc}
\toprule
\multirow{2}{*}[-2pt]{\makecell{EchoNext\\Train Size}} & \multicolumn{3}{c}{Tuned} & \multicolumn{3}{c}{Probed} \\
\cmidrule(lr){2-4}\cmidrule(lr){5-7}
 & ProtoSSL HEEDB & SupProto HEEDB (NoProj) & SupProto HEEDB & ProtoSSL HEEDB & SupProto HEEDB (NoProj) & SupProto HEEDB \\
\midrule
72475 & \textbf{0.819} [0.818-0.820] & 0.807 [0.806-0.808] & 0.782 [0.780-0.783] & \Xul{0.810} [0.809-0.811] & 0.773 [0.772-0.774] & 0.763 [0.762-0.764] \\
32768 & \textbf{0.814} [0.813-0.815] & 0.802 [0.801-0.803] & 0.758 [0.757-0.759] & \Xul{0.806} [0.805-0.807] & 0.761 [0.760-0.762] & 0.757 [0.756-0.758] \\
16384 & \textbf{0.798} [0.797-0.799] & \Xul{0.793} [0.792-0.794] & 0.740 [0.739-0.742] & \textbf{0.798} [0.797-0.799] & 0.757 [0.756-0.758] & 0.750 [0.749-0.751] \\
8192 & \textbf{0.790} [0.789-0.791] & \Xul{0.775} [0.774-0.776] & 0.731 [0.729-0.732] & \textbf{0.790} [0.789-0.791] & 0.742 [0.741-0.743] & 0.732 [0.731-0.733] \\
4096 & \textbf{0.782} [0.780-0.783] & 0.756 [0.755-0.757] & 0.728 [0.727-0.729] & \Xul{0.765} [0.764-0.766] & 0.721 [0.720-0.723] & 0.720 [0.719-0.721] \\
2048 & \textbf{0.779} [0.778-0.780] & \Xul{0.757} [0.756-0.758] & 0.730 [0.729-0.731] & 0.747 [0.746-0.748] & 0.720 [0.719-0.722] & 0.718 [0.717-0.719] \\
1024 & \textbf{0.768} [0.767-0.769] & \Xul{0.756} [0.755-0.758] & 0.725 [0.724-0.726] & 0.729 [0.728-0.731] & 0.672 [0.671-0.674] & 0.682 [0.681-0.684] \\
512 & \textbf{0.761} [0.760-0.763] & \Xul{0.750} [0.749-0.751] & 0.711 [0.710-0.713] & 0.723 [0.722-0.725] & 0.655 [0.654-0.657] & 0.660 [0.659-0.662] \\
256 & \textbf{0.739} [0.737-0.740] & \Xul{0.738} [0.737-0.740] & 0.690 [0.689-0.691] & 0.701 [0.700-0.702] & 0.654 [0.652-0.655] & 0.667 [0.665-0.668] \\
\bottomrule
\end{tabular}

\end{adjustbox}
}
\end{table}

SupProto HEEDB is intended to test whether a conventional label-supervised, projection-based prototype model learned on one source task can be reused under a new downstream label space when using our novel assignment mechanism. We therefore apply the standard label-supervised projection step (\autoref{sec:projection}) in HEEDB before transfer, because this is the source-domain prototype set produced by supervised projection-based training: the learned prototype vectors have been grounded in real HEEDB examples with their source labels. Transferring the pre-projection latent vectors would answer a different question, namely whether label-supervised latent prototype parameters provide a useful initialization for downstream projection. While this is a valid ablation, it is not the same as transferring a completed source-domain projection-based prototype model.

During pretraining, ProtoSSL has no source label space and is not trained as a source-task classifier, so there is no label-supervised source projection to perform. Its pretrained prototypes remain an unlabeled motif bank until downstream adaptation, where they are assigned to target labels and projected in the target domain. In all downstream comparisons, including SupProto HEEDB and ProtoSSL, the final prototypes used for target-task prediction and explanation are target-projected after assignment.

For completeness, we also evaluate a no-source-projection variant of SupProto HEEDB: the label-supervised, HEEDB-pretrained, latent prototypes are transferred to a target dataset, following the same assignment, optional fine-tuning, projection, and classifier training stages as before. We refer to this variant as SupProto HEEDB (NoProj). We find that this variant improves over source-projected SupProto HEEDB, showing that HEEDB projection may hinder transfer (\autoref{tab:supproto_noproj}). However, ProtoSSL consistently outperforms no-source-projection SupProto HEEDB (NoProj) under both frozen probing and fine-tuning. Thus, ProtoSSL's advantage is not an artifact of comparing against a pretraining-projected supervised baseline; pretraining-projected SupProto HEEDB remains the primary baseline because it corresponds to transfer of a standard supervised projection-based prototype model after traditional label-supervised prototype training.


\subsection{Label-Supervised Pretraining with More Labels}
\label{app:heedb_150}

\begin{table}[tp]
\centering
\caption{\textbf{We ablate the label-supervised prototype label-space over HEEDB to understand if label-supervised pretrained transferrability is determined by label diversity.} Specifically, rather than 20 clinician-curated labels used elsewhere in our study, we test a 150 label set which is more diverse, yet lower quality. Notably, these 150 labels remain under the broad category of ECG interpretation. We test either 1050 or 7500 pretraining prototypes (see \protect\autoref{app:heedb_150} for explanation on comparability to other results) under both probing and tuning on EchoNext with 14 prototypes per downstream label. Best model \textbf{bolded}, next best model \Xul{underlined}.}
\label{tab:heedb_150}
{
\setlength{\tabcolsep}{4pt}
\begin{adjustbox}{max width=\textwidth}

\begin{tabular}{rrccccc}
\toprule
\multirow{2}{*}[-2pt]{\makecell[r]{Num Pretrained\\Prototypes}} & \multirow{2}{*}[-2pt]{\makecell[r]{EchoNext\\Train Size}} & \multicolumn{2}{c}{Tuned} & \multicolumn{2}{c}{Probed} & \multirow{2}{*}[-2pt]{SupProto Direct} \\
\cmidrule(lr){3-4}\cmidrule(lr){5-6}
 &  & ProtoSSL HEEDB & SupProto HEEDB & ProtoSSL HEEDB & SupProto HEEDB &  \\
\midrule
\multirow[t]{9}{*}{1050} & 72475 & \textbf{0.819} [0.818-0.820] & 0.801 [0.800-0.802] & \Xul{0.814} [0.813-0.815] & 0.784 [0.783-0.785] & 0.805 [0.804-0.806] \\
 & 32768 & \Xul{0.809} [0.808-0.810] & 0.777 [0.776-0.778] & \textbf{0.811} [0.810-0.812] & 0.781 [0.780-0.782] & 0.797 [0.796-0.798] \\
 & 16384 & \Xul{0.795} [0.794-0.796] & 0.762 [0.761-0.764] & \textbf{0.803} [0.802-0.804] & 0.778 [0.777-0.779] & 0.781 [0.780-0.782] \\
 & 8192 & \Xul{0.784} [0.783-0.785] & 0.752 [0.751-0.753] & \textbf{0.795} [0.794-0.796] & 0.769 [0.767-0.770] & 0.762 [0.761-0.763] \\
 & 4096 & \textbf{0.776} [0.775-0.777] & 0.747 [0.746-0.748] & \Xul{0.772} [0.771-0.773] & 0.754 [0.753-0.755] & 0.738 [0.737-0.739] \\
 & 2048 & \textbf{0.772} [0.771-0.774] & 0.743 [0.742-0.744] & \Xul{0.755} [0.753-0.756] & 0.741 [0.740-0.742] & 0.722 [0.721-0.723] \\
 & 1024 & \textbf{0.761} [0.760-0.762] & \Xul{0.739} [0.738-0.740] & 0.734 [0.733-0.735] & 0.715 [0.713-0.716] & 0.716 [0.715-0.717] \\
 & 512 & \textbf{0.752} [0.751-0.754] & 0.730 [0.729-0.731] & \Xul{0.731} [0.730-0.732] & 0.680 [0.679-0.682] & 0.711 [0.710-0.712] \\
 & 256 & \textbf{0.733} [0.731-0.734] & \Xul{0.722} [0.721-0.723] & 0.705 [0.704-0.707] & 0.679 [0.678-0.681] & 0.702 [0.701-0.703] \\
\midrule
\multirow[t]{9}{*}{7500} & 72475 & \textbf{0.820} [0.819-0.821] & 0.774 [0.772-0.775] & \Xul{0.811} [0.810-0.812] & 0.769 [0.768-0.770] & 0.805 [0.804-0.806] \\
 & 32768 & \textbf{0.809} [0.808-0.810] & 0.757 [0.755-0.758] & \Xul{0.806} [0.805-0.807] & 0.769 [0.768-0.770] & 0.797 [0.796-0.798] \\
 & 16384 & \Xul{0.798} [0.797-0.799] & 0.749 [0.747-0.750] & \textbf{0.799} [0.798-0.800] & 0.758 [0.757-0.759] & 0.781 [0.780-0.782] \\
 & 8192 & \Xul{0.785} [0.784-0.786] & 0.738 [0.737-0.740] & \textbf{0.792} [0.791-0.793] & 0.737 [0.736-0.738] & 0.762 [0.761-0.763] \\
 & 4096 & \Xul{0.778} [0.777-0.779] & 0.744 [0.742-0.745] & \textbf{0.780} [0.779-0.781] & 0.720 [0.719-0.722] & 0.738 [0.737-0.739] \\
 & 2048 & \textbf{0.769} [0.767-0.770] & 0.750 [0.749-0.752] & \Xul{0.758} [0.757-0.760] & 0.715 [0.714-0.716] & 0.722 [0.721-0.723] \\
 & 1024 & \textbf{0.763} [0.762-0.764] & \Xul{0.737} [0.735-0.738] & 0.734 [0.733-0.735] & 0.660 [0.658-0.661] & 0.716 [0.715-0.717] \\
 & 512 & \textbf{0.752} [0.751-0.753] & 0.726 [0.725-0.728] & \Xul{0.727} [0.725-0.728] & 0.644 [0.642-0.645] & 0.711 [0.710-0.712] \\
 & 256 & \textbf{0.716} [0.715-0.717] & \Xul{0.712} [0.711-0.714] & 0.702 [0.701-0.703] & 0.587 [0.585-0.588] & 0.702 [0.701-0.703] \\
\bottomrule
\end{tabular}

\end{adjustbox}
}
\end{table}

In our experiments, we compare ProtoSSL HEEDB to a label-supervised pretrained prototype baseline SupProto HEEDB. To enable this comparison, we use the 20 HEEDB labels curated by \citet{li_electrocardiogram_2025} via clinician-consensus. However, the authors' own model (ECGFounder) was trained over a wider, but lower quality, set of 150 labels they selected. As context, the labels in HEEDB are derived from templated text statements \citep{koscova_harvard-emory_2026}, meaning nonspecific text tokens may be treated as \enquote{labels}. For example, in the 150 label set, labels such as \enquote{acute} or \enquote{with undetermined rhythm irregularity} are present. This data quality concern motivated our use of the HEEDB-20 label set for SupProto HEEDB.

Nonetheless, a natural critique of only utilizing 20 labels for label-supervised pretraining is unfairly constraining SupProto HEEDB to work with less supervisory signal than is available, potentially resulting in subpar downstream generalizability. However, we note that this limitation does not apply to our audio experiements, where SupProto AudioSet trains with 527 labels. Here, we ablate the HEEDB label set decision for ECG. Notably, we test two variations for SupProto HEEDB using 150 labels: 1) holding approximately 1000 prototypes (as elsewhere in our study) in the latent bank by using 7 prototypes per label, yielding 1050 pretrained prototypes, or 2) holding 50 prototypes per pretraining label (as elsewhere in our study) in the latent bank, yielding 7500 pretrained prototypes.

These ablation results are presented in \autoref{tab:heedb_150} over EchoNext and its subsets. Using a fixed 14 prototypes per label in the downstream dataset adaptation, we observe that ProtoSSL HEEDB with a matched number of pretraining prototypes consistently outperforms SupProto HEEDB pretrained over the HEEDB-150 label set in both fine-tuned and probed settings. These results mirror our findings in comparing to SupProto HEEDB pretrained over the HEEDB-20 label set. We emphasize that \emph{even with the expanded pretraining label set, we still consider HEEDB's 150 ECG interpretation labels to have no overlap with the downstream EchoNext target labels}. Thus, as in Sections \ref{sec:generalization} and \ref{sec:label_efficiency}, we hypothesize that the poor transferrability of SupProto HEEDB pretrained prototypes is not simply due to lack of diverse supervisory signal, but due to misaligned and fixed prototype-label semantics learned during pretraining.


\subsection{Varying the Number of Upstream Prototypes}
\label{app:num_proto_upstream}

\begin{table}[tp]
\centering
\caption{\textbf{Varying the size of ProtoSSL's pretrained prototype bank may better capture patterns that are relatively rare in the pretraining data, yet downstream tasks rely on.} Probing ProtoSSL HEEDB on ZZU pECG at full dataset scale when pretrained with 7500 prototypes recovers SupProto Direct performance.}
\label{tab:zzu_capacity}
{
\setlength{\tabcolsep}{4pt}
\begin{adjustbox}{max width=\textwidth}

\begin{tabular}{ccc}
\toprule
\multicolumn{2}{c}{Probed ProtoSSL HEEDB} & \multirow{2}{*}[-2pt]{SupProto Direct} \\
\cmidrule{1-2}
1000 pretrained prototypes & 7500 pretrained prototypes & \\
\midrule
0.776 [0.775-0.778] & 0.790 [0.788-0.792] & 0.790 [0.788-0.792] \\
\bottomrule
\end{tabular}

\end{adjustbox}
}
\end{table}

One additional result derived from \autoref{app:heedb_150} is the effect of varying the number of pretrained prototypes for ProtoSSL HEEDB. When adapting ProtoSSL HEEDB's pretrained prototypes to EchoNext using a fixed 14 prototypes per label, we do not observe much difference between frozen probing of a pretrained bank of 1000 vs 7500 prototypes (Supplementary Tables \ref{tab:ecg_full} vs \ref{tab:heedb_150}, respectively). Yet, we observe substantive differences when probing ProtoSSL HEEDB on ZZU pECG with 14 prototypes per label, where probing ProtoSSL HEEDB with 7500 pretrained prototypes matches SupProto Direct at full dataset scale, while ProtoSSL HEEDB with 1000 pretrained prototypes lags behind (\autoref{tab:zzu_capacity}). This suggests that training-free adaptation to ZZU pECG's task crosses a wider distributional shift than adaptation to EchoNext, as coverage of ZZU's prototypes seems constrained by prototype bank capacity. Intuitively, a downstream task which relies on patterns rarely seen in the pretraining dataset will not be frequently learned by the self supervised pretraining objective, requiring a larger pretraining bank to adequately capture the rare patterns in a sufficient number of prototypes; indeed, HEEDB to ZZU adaptation encompasses an adult-to-pediatric shift where we might expect such relatively rare waveform morphologies in HEEDB.


\subsection{Varying the Number of Downstream Prototypes}
\label{app:num_proto}

\begin{table}[tp]
\centering
\caption{\textbf{We ablate the number of prototypes per label.} Throughout our ECG experiments, we use 14 prototypes per label as the default (described in \autoref{app:ecg_details}). However, over EchoNext, we can ablate the number of prototypes per label, given only 12 labels and 1000 prototypes in the pretrained prototype bank. We present here experiments using 7 and 28 prototypes per label, half and twice the number of prototypes in our default configuration, respectively. While we find that prototype-based models are sensitive to the number of prototypes per label (consistent with the literature \citep{sethi_protoecgnet_2025, chen_this_2019, rymarczyk_interpretable_2022}), ProtoSSL consistently beats label-supervised prototype baselines using the same number of prototypes per label. Best model \textbf{bolded}, next best model \Xul{underlined}.}
\label{tab:ppl}
{
\setlength{\tabcolsep}{4pt}
\begin{adjustbox}{max width=\textwidth}

\begin{tabular}{rrccccc}
\toprule
\multirow{2}{*}[-2pt]{\makecell[r]{Prototypes\\Per Label}} & \multirow{2}{*}[-2pt]{\makecell[r]{EchoNext\\Train Size}} & \multicolumn{2}{c}{Tuned} & \multicolumn{2}{c}{Probed} & \multirow{2}{*}[-2pt]{SupProto Direct} \\
\cmidrule(lr){3-4}\cmidrule(lr){5-6}
 &  & ProtoSSL HEEDB & SupProto HEEDB & ProtoSSL HEEDB & SupProto HEEDB &  \\
\midrule
\multirow[t]{9}{*}{7} & 72475 & \textbf{0.813} [0.812-0.815] & 0.783 [0.782-0.784] & 0.796 [0.795-0.798] & 0.754 [0.753-0.755] & \Xul{0.805} [0.804-0.806] \\
 & 32768 & \textbf{0.805} [0.804-0.806] & 0.765 [0.764-0.766] & 0.788 [0.787-0.789] & 0.751 [0.750-0.752] & \Xul{0.793} [0.792-0.794] \\
 & 16384 & \textbf{0.791} [0.790-0.792] & 0.752 [0.750-0.753] & \Xul{0.779} [0.778-0.780] & 0.742 [0.741-0.743] & 0.778 [0.777-0.779] \\
 & 8192 & \textbf{0.783} [0.782-0.784] & 0.735 [0.734-0.736] & \Xul{0.775} [0.774-0.776] & 0.728 [0.727-0.730] & 0.752 [0.750-0.753] \\
 & 4096 & \textbf{0.771} [0.769-0.772] & 0.731 [0.730-0.733] & \Xul{0.745} [0.744-0.746] & 0.719 [0.718-0.720] & 0.738 [0.737-0.739] \\
 & 2048 & \textbf{0.768} [0.767-0.769] & 0.729 [0.727-0.730] & \Xul{0.732} [0.731-0.733] & 0.715 [0.714-0.716] & 0.730 [0.729-0.732] \\
 & 1024 & \textbf{0.767} [0.766-0.768] & \Xul{0.730} [0.729-0.731] & 0.721 [0.720-0.722] & 0.679 [0.677-0.680] & 0.716 [0.714-0.717] \\
 & 512 & \textbf{0.759} [0.757-0.760] & 0.712 [0.711-0.713] & 0.718 [0.717-0.720] & 0.660 [0.659-0.662] & \Xul{0.721} [0.720-0.722] \\
 & 256 & \textbf{0.736} [0.734-0.737] & \Xul{0.703} [0.702-0.704] & 0.690 [0.689-0.692] & 0.670 [0.669-0.671] & 0.701 [0.700-0.702] \\
\midrule
\multirow[t]{9}{*}{14} & 72475 & \textbf{0.819} [0.818-0.820] & 0.782 [0.780-0.783] & \Xul{0.810} [0.809-0.811] & 0.763 [0.762-0.764] & 0.805 [0.804-0.806] \\
 & 32768 & \textbf{0.814} [0.813-0.815] & 0.758 [0.757-0.759] & \Xul{0.806} [0.805-0.807] & 0.757 [0.756-0.758] & 0.797 [0.796-0.798] \\
 & 16384 & \textbf{0.798} [0.797-0.799] & 0.740 [0.739-0.742] & \textbf{0.798} [0.797-0.799] & 0.750 [0.749-0.751] & \Xul{0.781} [0.780-0.782] \\
 & 8192 & \textbf{0.790} [0.789-0.791] & 0.731 [0.729-0.732] & \textbf{0.790} [0.789-0.791] & 0.732 [0.731-0.733] & \Xul{0.762} [0.761-0.763] \\
 & 4096 & \textbf{0.782} [0.780-0.783] & 0.728 [0.727-0.729] & \Xul{0.765} [0.764-0.766] & 0.720 [0.719-0.721] & 0.738 [0.737-0.739] \\
 & 2048 & \textbf{0.779} [0.778-0.780] & 0.730 [0.729-0.731] & \Xul{0.747} [0.746-0.748] & 0.718 [0.717-0.719] & 0.722 [0.721-0.723] \\
 & 1024 & \textbf{0.768} [0.767-0.769] & 0.725 [0.724-0.726] & \Xul{0.729} [0.728-0.731] & 0.682 [0.681-0.684] & 0.716 [0.715-0.717] \\
 & 512 & \textbf{0.761} [0.760-0.763] & 0.711 [0.710-0.713] & \Xul{0.723} [0.722-0.725] & 0.660 [0.659-0.662] & 0.711 [0.710-0.712] \\
 & 256 & \textbf{0.739} [0.737-0.740] & 0.690 [0.689-0.691] & 0.701 [0.700-0.702] & 0.667 [0.665-0.668] & \Xul{0.702} [0.701-0.703] \\
\midrule
\multirow[t]{9}{*}{28} & 72475 & \textbf{0.820} [0.819-0.821] & 0.775 [0.774-0.776] & \Xul{0.817} [0.816-0.818] & 0.774 [0.773-0.775] & 0.804 [0.803-0.805] \\
 & 32768 & \Xul{0.813} [0.812-0.814] & 0.757 [0.755-0.758] & \textbf{0.815} [0.814-0.816] & 0.766 [0.765-0.767] & 0.791 [0.790-0.792] \\
 & 16384 & \Xul{0.804} [0.803-0.805] & 0.734 [0.732-0.735] & \textbf{0.809} [0.808-0.810] & 0.759 [0.758-0.760] & 0.779 [0.778-0.780] \\
 & 8192 & \Xul{0.797} [0.796-0.798] & 0.721 [0.720-0.722] & \textbf{0.803} [0.802-0.804] & 0.743 [0.742-0.744] & 0.763 [0.762-0.764] \\
 & 4096 & \textbf{0.785} [0.783-0.786] & 0.718 [0.717-0.720] & \textbf{0.785} [0.784-0.786] & 0.727 [0.726-0.728] & \Xul{0.745} [0.744-0.746] \\
 & 2048 & \textbf{0.783} [0.782-0.784] & 0.713 [0.712-0.715] & \Xul{0.762} [0.761-0.763] & 0.721 [0.720-0.722] & 0.719 [0.718-0.720] \\
 & 1024 & \textbf{0.773} [0.772-0.775] & 0.707 [0.706-0.708] & \Xul{0.743} [0.742-0.745] & 0.684 [0.683-0.685] & 0.713 [0.712-0.715] \\
 & 512 & \textbf{0.768} [0.767-0.770] & 0.700 [0.699-0.702] & \Xul{0.729} [0.728-0.730] & 0.659 [0.658-0.661] & 0.703 [0.702-0.705] \\
 & 256 & \textbf{0.749} [0.748-0.750] & 0.684 [0.683-0.686] & \Xul{0.710} [0.709-0.712] & 0.666 [0.664-0.667] & 0.699 [0.698-0.700] \\
\bottomrule
\end{tabular}

\end{adjustbox}
}
\end{table}

As described in \autoref{app:ecg_details}, we use 14 prototypes per label in downstream tasks across our ECG experiments. We fixed this hyperparameter to control the effect of prototypes per label, which prototype-based models are known to be sensitive to \citep{sethi_protoecgnet_2025, chen_this_2019, rymarczyk_interpretable_2022}. Briefly, we find that ProtoSSL outperforms label-supervised baselines at this fixed number of prototypes per label for domain generalization and label efficiency (presented in Sections \ref{sec:generalization} and \ref{sec:label_efficiency}, respectively). We demonstrate in \autoref{tab:ppl} that these trends hold across varying configurations of prototypes per label, with either 7 or 28 prototypes per label. Indeed, we recapitulate that projected-based prototype models are sensitive to this hyperparameter, yet ProtoSSL yields consistent gains regardless, surpassing label-supervised prototype baselines with the same number of prototypes-per-label.


\subsection{Deriving Target Domain Prototypes from Foundation Models}
\label{app:protos_from_fms}

\begin{table}[tp]
\centering
\caption{\textbf{Ablation of ProtoSSL's prototype pretraining.} Prototypes are derived purely in the downstream dataset, EchoNext, using frozen embeddings derived by ECGFounder. All de novo prototype discovery approaches do worse than simple linear probing over ECGFounder embeddings. Prototypes derived by SK-OT clustering is not meaningfully different from random initialization and subsequent assignment/projection of prototypes. All encoders are frozen and any prototype updates are purely non-parametric. Any prototype-based models use 14 prototypes per label and are projected onto real training samples prior to linear probing over prototype activations. ECGFounder (SK-OT): prototypes are discovered using the algorithm proposed by \citet{zhu_interpretable_2025}. ECGFounder (LAP): 1000 prototypes are randomly initialized before being assigned by ProtoSSL's assignment method. ECGFounder (Rand Assign): 1000 prototypes are randomly initialized and randomly assigned. Best model \textbf{bolded}, next best model \Xul{underlined}.}
\label{tab:ecg_protos_from_fms}
{
\setlength{\tabcolsep}{4pt}
\begin{adjustbox}{max width=\textwidth}

\begin{tabular}{rccccc}
\toprule
\multirow{2}{*}[-2pt]{\makecell[l]{EchoNext\\Train Size}} & \multicolumn{5}{c}{Probed} \\
\cmidrule{2-6}
& ProtoSSL HEEDB & ECGFounder (SK-OT) & ECGFounder (LAP) & ECGFounder (Rand Assign) & ECGFounder \\
\midrule
72475 & \textbf{0.810} [0.809-0.811] & 0.785 [0.784-0.786] & 0.789 [0.788-0.790] & 0.788 [0.787-0.789] & \Xul{0.802} [0.801-0.803] \\
32768 & \textbf{0.806} [0.805-0.807] & 0.780 [0.779-0.781] & 0.782 [0.781-0.783] & 0.783 [0.782-0.784] & \Xul{0.798} [0.797-0.799] \\
16384 & \textbf{0.798} [0.797-0.799] & 0.774 [0.773-0.775] & 0.778 [0.777-0.779] & 0.779 [0.778-0.780] & \Xul{0.791} [0.790-0.792] \\
8192 & \textbf{0.790} [0.789-0.791] & 0.759 [0.758-0.760] & 0.765 [0.764-0.766] & 0.767 [0.766-0.768] & \Xul{0.781} [0.780-0.782] \\
4096 & \Xul{0.765} [0.764-0.766] & 0.747 [0.746-0.749] & 0.752 [0.751-0.753] & 0.756 [0.755-0.757] & \textbf{0.766} [0.765-0.767] \\
2048 & \Xul{0.747} [0.746-0.748] & 0.738 [0.737-0.740] & 0.740 [0.738-0.741] & 0.746 [0.745-0.748] & \textbf{0.753} [0.752-0.754] \\
1024 & \Xul{0.729} [0.728-0.731] & 0.694 [0.693-0.695] & 0.692 [0.690-0.693] & 0.699 [0.697-0.700] & \textbf{0.736} [0.735-0.737] \\
512 & \textbf{0.723} [0.722-0.725] & 0.690 [0.689-0.691] & 0.681 [0.679-0.682] & 0.694 [0.693-0.696] & \Xul{0.720} [0.719-0.721] \\
256 & \textbf{0.701} [0.700-0.702] & 0.684 [0.682-0.685] & 0.671 [0.670-0.673] & 0.682 [0.681-0.684] & \Xul{0.700} [0.698-0.701] \\
\bottomrule
\end{tabular}

\end{adjustbox}
}
\end{table}

\begin{table}[tp]
\centering
\caption{\textbf{Over EchoNext, the clustering algorithm proposed by \citet{zhu_interpretable_2025} to derive prototypes may not converge at low data scales, where the number of positive patches for a given class is similar to the number of prototype slots being discovered for each class.} As before, we use 14 prototypes per label for 12 labels in EchoNext for 168 total prototype slots.}
\label{tab:ecg_sk_ot_converge}

\footnotesize
\begin{tabular}{lrrrrrrrrr}
\toprule
& \multicolumn{9}{c}{EchoNext Train Subset Size} \\
\cmidrule(lr){2-10}
& 72475 & 32768 & 16384 & 8192 & 4096 & 2048 & 1024 & 512 & 256 \\
\midrule
SK-OT Convergence & \cmark & \cmark & \cmark & \cmark & \cmark & \cmark & \cmark & \xmark & \xmark \\
\bottomrule
\end{tabular}

\end{table}

In the modern paradigm of foundation models, pretrained embeddings provide a strong basis for modeling in downstream applications. \citet{zhu_interpretable_2025} introduce a framework for adapting FMs to prototype models by discovering prototypes (de novo in the target dataset) as cluster means of latent embeddings derived by the FM. The authors propose a non-parametric clustering approach based on optimal transport and solved via the Sinkhorn-Knopp algorithm \citep{knight_sinkhorn-knopp_2008}. Briefly, the method consists of deriving \(M\) clusters for each label \(l\), clustering only patches belonging to the label. This introduces the constraint that there must be more label-associated patches than clusters. We hypothesize that such a clustering approach may fail to converge at the extremes of label scarcity (when there are potentially only a single positive sample per label), even if there are technically sufficiently many patches that can be derived from the few positive samples.

To understand whether this method generalizes to our current study, which we refer to as SK-OT, we implement the prototype-discovery clustering approach over patch embeddings of ECGFounder \citep{li_electrocardiogram_2025}. ECGFounder produces 20 patches for each 10-second ECG and we use \(M=14\) prototypes per label, as done elsewhere in our study and satisfying the technical constraint of the clustering algorithm. One modification we make to the method, originally developed for vision tasks, is we do not do background removal of patches and simply treat all patches from an ECG as foreground patches, as each patch reasonably captures some portion of a heartbeat. To be comparable to our closest setting (non-parametric prototype assignment and frozen probing of ProtoSSL) we additionally project and ground prototypes derived by SK-OT onto real training samples and train a linear classifier, as in \autoref{sec:projection}. We find that SK-OT does far worse than ProtoSSL and even worse than directly probing ECGFounder embeddings (\autoref{tab:ecg_protos_from_fms}). Furthermore, we find that the SK-OT clustering does not converge on the smallest data subsets of EchoNext, where the equipartition condition is not met. This leaves many prototypes in their randomly initialized state or otherwise updated far fewer times relative to the theoretical number of updates if the equipartition constraint was met in every iteration. The data subsets where SK-OT fails to converge are summarized in \autoref{tab:ecg_sk_ot_converge}.

To contextualize these results, we compare to two additional approaches for constructing prototypes over ECGFounder embeddings. After initializing a pool of 1000 random prototypes (following our setting for \(K\) elsewhere), we assign these random prototypes to labels either according to ProtoSSL's LAP reduction or randomly. During either assignment, we similarly use \(M=14\) prototypes per label. The assigned, latent prototypes remain randomly initialized before we do label-supervised projection onto the nearest patch embedding derived from ECGFounder. We refer to these variations as ECGFounder LAP and Rand Assign, respectively. All three of SK-OT, LAP, and Rand Assign rely on random initialization of prototypes in the target domain, only varying in the way latent prototypes are updated with target-domain label semantics: either by clustering (SK-OT) or via assignment (LAP vs Rand Assign). We find that both assignment baselines perform comparably to SK-OT (\autoref{tab:ecg_protos_from_fms}). Together, these results suggest that while FM-derived embeddings, such as ECGFounder's, may drive excellent classification performance (\autoref{tab:ecg_blackbox}), they may not be readily adaptable to all prototype settings; more work is needed to expand upon these results.


\subsection{Final-Layer Coefficients Support Assigned Prototype Evidence Without Strict Masking}
\label{app:classifier_explanation}

\begin{table}[tp]
\centering
\caption{\textbf{Over EchoNext, we analyze the coefficients of the final logistic regression for frozen probing of ProtoSSL HEEDB.} Given the ``\textit{this} looks like \textit{that}'' paradigm of prototype part models, in a model that predicts a given binary label, the coefficients of activations (similarity scores) for prototypes belonging to the positive class (\(\beta^+\)) should be larger and more positive than for prototypes which do not belong to the target label (\(\beta^-\)). When examined as odds ratios, we would expect \(\exp(\beta^+) > 1\) and \(\exp(\beta^+) > \exp(\beta^-)\), or expressed as a ratio \(\exp(\beta^+)/\exp(\beta^-) > 1\). We indeed find our final classification models exhibit this property across all data scales and all seeds (\(p=2.84e-14\) by sign test assessing if ratios \(> 1\)). As we do \emph{not} follow \citet{chen_this_2019} in the classifier initialization or regularization masking, the emergence of positive prototype weighting is remarkable, given the L2 regularization applied to all features and substantial label cooccurrence in EchoNext \citep{poterucha_detecting_2025}, discussed further in \autoref{app:classifier_explanation}.}
\label{tab:ecg_proto_coef}
{
\setlength{\tabcolsep}{4pt}
\begin{adjustbox}{max width=\textwidth}

\begin{tabular}{lrrrrrrrrr}
\toprule
\makecell[l]{EchoNext\\Train Size} & 72475 & 32768 & 16384 & 8192 & 4096 & 2048 & 1024 & 512 & 256 \\
\midrule
Pos-Prototype OR > 1 & \cmark & \cmark & \cmark & \cmark & \cmark & \cmark & \cmark & \cmark & \cmark \\
Pos/Neg Ratio > 1 & \cmark & \cmark & \cmark & \cmark & \cmark & \cmark & \cmark & \cmark & \cmark \\
Pos/Neg-Prototype Ratio & 1.029 & 1.027 & 1.019 & 1.016 & 1.011 & 1.009 & 1.007 & 1.006 & 1.004 \\
\bottomrule
\end{tabular}

\end{adjustbox}
}
\end{table}

\paragraph{Conceptual motivation for simplified classifier over prototype activations:} The original ProtoPNet \citep{chen_this_2019} classifier training utilizes L1 regularization that is applied only to the weights corresponding to the positive-class prototypes. However, this masking encodes a single-label assumption (i.e. not multilabel) that is not generally appropriate for multilabel prediction. In the original multiclass setting, suppressing off-class prototype connections supports a clean class-specific scoring rule because each example belongs to exactly one class. In multilabel settings, however, labels can share motifs, co-occur systematically, or provide meaningful counter-evidence. In ECG interpretation, for example, a morphology prototype reflecting ST-segment elevation may provide evidence for multiple infarction-related labels, while dissimilarity to prototypes representing normal sinus rhythm may support atrial- or ventricular-paced rhythm labels. This is closely related to the motivation behind ProtoECGNet's co-occurrence loss \citep{sethi_protoecgnet_2025}, which explicitly recognized that clinically related labels should not be treated as independent or mutually exclusive when learning prototype structure. We therefore do not impose a hard final-layer L1 regularization with masking; instead, we use an L2 regularized logistic-regression classifier over projected prototype activations, allowing the model to learn balanced positive and negative relationships among prototypes while preserving projection-grounded prototype evidence.

In this setting, target-assigned prototypes should receive systematically more positive weight for their assigned label than non-target prototypes, while cross-label connections need not vanish. In fact, the weight initialization in the original ProtoPNet formulation (where positive class connections are initialized with weight of 1 and $-0.5$ otherwise) would support this without the masked regularization. However, such initializations may require extra steps to implement in practice, impeding using out-of-the-box implementations of linear classifiers provided by standard packages. Yet, these relative weights are additionally meaningful to the interpretability of prototype-based models. Semantically, we want a prediction to be made, not only because \enquote{\textit{this} looks like \textit{that}}, but also because \enquote{\textit{this} looks more like \textit{that} positive example than \textit{those} other examples}.

\paragraph{Empirical evidence for meaningful classifier connections:} To assess this, we analyzed the learned logistic-regression coefficients for frozen probing of ProtoSSL HEEDB on EchoNext (\autoref{tab:ecg_proto_coef}). For each binary label, we partitioned prototypes into those assigned to the target label and those assigned to other labels. Let $\beta^+$ denote coefficients for target-assigned prototypes and $\beta^-$ denote coefficients for non-target prototypes. Since each feature is a prototype activation score, positive coefficients correspond to prototype similarity increasing the odds of the target label. Thus, a classifier using assigned prototypes as intended should satisfy $\exp(\beta^+) > 1$ and should weight target-assigned prototypes more positively than non-target prototypes, summarized by $\exp(\beta^+) / \exp(\beta^-) > 1$. We observe both properties across every EchoNext training scale, from the full training set down to 256 examples. The reported ratios are close to 1 because the final classifier is strongly L2 regularized (see \autoref{sec:projection}). As L2 regularization shrinks all coefficients toward zero, this correspondingly drives odds ratios toward 1. In addition, EchoNext is multilabel with substantial label co-occurrence (see Supplemental Figure 4 of the original EchoNext publication by \citet{poterucha_detecting_2025}), so non-target prototypes can still carry legitimate positive information for related labels. Despite these conservative factors, the direction of the effect for target-assigned prototypes being weighted more heavily than non-target prototypes is remarkably consistent across all data scales and seeds ($p=2.84 \times 10^{-14}$ by sign test for ratios greater than 1). As we again note that we use no special initialization or masked regularization (and we do L2 regularization instead of L1), this property observed in \autoref{tab:ecg_proto_coef} is emergent, indicating that prototype-label assignments are reflected in the learned classifier's weighting of prototype evidence.

\section{Human Evaluation Study Details}
\label{app:user_study}

\subsection{Participants}
\label{app:user_participants}
We recruited seven fourth-year MD students from a single U.S. medical school. All participants had completed core clinical rotations, including internal medicine and cardiology, and had received formal instruction in ECG interpretation. Participants provided informed consent electronically before participation. The study was reviewed by the
University of Chicago
Institutional Review Board (IRB) and was determined to be exempt from further IRB review [IRB25-0730], under the Federal Regulations 45 CFR 46.104(d)(3)(i)(A-B).

\subsection{Study design}
We compared explanation quality for two prototype-based ECG models on PTB-XL \citep{wagner_ptb-xl_2020}: ProtoSSL, which used frozen self-supervised prototypes learned from HEEDB \citep{koscova_harvard-emory_2026} followed by label-supervised assignment and projection on PTB-XL, and SupProto Direct, which was trained end-to-end from scratch on PTB-XL. Of note, as in all the prototype models we compare in this study, we use 1-second 2D partial prototypes for all labels, rather than the task-specific three branch structure in \citet{sethi_protoecgnet_2025}. We focused on four diagnoses which were within the original ProtoECGNet's 2D partial prototype branch: anterior myocardial infarction (AMI), right bundle branch block (RBBB), left bundle branch block (LBBB), and premature ventricular contraction (PVC). We selected five cases per diagnosis, for a total of 20 cases. To isolate explanation quality from classification accuracy, we restricted evaluation to cases for which both models made the correct target prediction. Each participant evaluated all 20 cases. 

Case selection was designed to preserve a single dominant morphology and reduce confounding co-occurring findings. For AMI, we retained anterior infarction patterns; in PTB-XL, this corresponds primarily to anteroseptal and anterolateral myocardial infarction labels, while inferior, inferoposterior, posterior, and other non-anterior infarction patterns were excluded. For RBBB, we included complete right bundle branch block only and excluded incomplete right bundle branch block, left bundle branch block, and nonspecific intraventricular conduction delay. For LBBB, we included complete left bundle branch block only and excluded incomplete left bundle branch block, right bundle branch block, and nonspecific intraventricular conduction delay. For PVC, we excluded co-occurring myocardial infarction, major conduction abnormalities, pacing, and other findings that would obscure interpretation of a single dominant PVC morphology.

Each case consisted of two sequential tasks. In the first, participants evaluated prototype quality in isolation. In the second, they evaluated full prototype-test ECG explanations. The evaluation protocol was adapted from prior work on prototype interpretability \citep{davoodi_interpretability_2023}.

\subsection{Visualization interface}
All ECGs were displayed in a standard clinical 12-lead layout, with three rows of 2.5-second segments and a 10-second rhythm strip (lead II). Prototype visualization and highlighting followed the protocol introduced in \citet{sethi_protoecgnet_2025}.

For each model, the interface highlighted the 1-second interval used for matching. As the standard 12-lead display presents different leads at different time offsets, each highlighted interval was accompanied by a synchronized full-lead pop-out view showing all 12 leads over the same 1-second window. This allowed participants to inspect waveform morphology across leads within the exact interval used by the model.

In the prototype task, participants were shown each prototype ECG with its highlighted interval and synchronized full-lead view. In the explanation task, participants were shown paired prototype and test ECGs for each model, with corresponding highlighted intervals and synchronized full-lead views, together with the cosine similarity score reported by that model. Participants were instructed to judge explanation quality based on ECG morphology rather than score magnitude.

\subsection{Prompts and responses}
\label{app:user_prompts}
For each case, participants completed two tasks.

\textbf{Prototype evaluation task.}
Participants were shown two prototype ECGs (Model A and Model B) with highlighted intervals. They were asked:
\begin{itemize}
    \item \textit{Comparative:} ``Which prototype is a better example of the predicted diagnosis?'' (A / B / Both / Neither)
    \item \textit{Absolute (Model A):} ``Is this a good example of the predicted diagnosis?'' (Yes / No)
    \item \textit{Absolute (Model B):} ``Is this a good example of the predicted diagnosis?'' (Yes / No)
\end{itemize}

\textbf{Explanation evaluation task.}
Participants were shown paired prototype-test ECG explanations for each model, with corresponding highlighted intervals and cosine similarity scores. They were asked:
\begin{itemize}
    \item \textit{Comparative:} ``Which explanation provides a better match between the prototype and the test ECG?'' (A / B / Both / Neither)
    \item \textit{Absolute (Model A):} ``Is this a convincing match between the prototype and the test ECG?'' (Yes / No)
    \item \textit{Absolute (Model B):} ``Is this a convincing match between the prototype and the test ECG?'' (Yes / No)
\end{itemize}

In both tasks, participants were instructed to base their judgments on whether the waveform morphology represented a clear and clinically appropriate example of the diagnosis. Prompt wording was adapted from \citet{davoodi_interpretability_2023}.

\subsection{Procedure and quality control}
Participants attended a single instructional session before the study, during which the interface, task structure, and example cases were reviewed. A qualification example was included in which one model clearly outperformed the other; all participants correctly identified the superior model. Participants then completed all evaluations in a single REDCap session within five days of the instructional session.

Model identity (A/B) was randomized independently for each case, and participants were blinded to model type and study hypotheses. Case order was fixed across participants. Within each case, both models were evaluated on the same held-out test ECG, so comparative judgments reflected differences in prototype selection and matching rather than differences in the underlying example.

\subsection{Statistical analysis}
The primary analysis focused on the binary absolute judgments. For each participant, we computed the proportion of responses rated as \enquote{good} for each model separately for the prototype task and the explanation task, treating the participant as the unit of analysis. For each task, we report the participant-level mean acceptability rate for each model, the paired mean difference in percentage points, and a 95\% CI computed using a $t$-interval over participant-level paired differences. Model differences were assessed using exact two-sided Wilcoxon signed-rank tests on participant-level paired acceptability rates. Inter-rater agreement for the binary yes/no judgments was summarized using Fleiss' $\kappa$ ($\kappa = 0.29$ overall). We additionally report descriptive preference distributions (ProtoSSL / SupProto Direct / Both / Neither) for the comparative questions. All analyses were performed in Python using SciPy and statsmodels.

\section{Analysis of Human Evaluation Study}
\label{app:user_study_analysis}

\begin{table}[tp]
\centering
\caption{\textbf{Comparative preference (\%).} Percentages are computed over all evaluations.}
\label{tab:user_study_pref}

\begin{tabular}{lcccc}
\toprule
Task & ProtoSSL HEEDB & SupProto Direct & Both & Neither \\
\midrule
Global quality & 45.7 & 17.9 & 36.4 & 0.0 \\
Case explanations & 45.7 & 25.7 & 24.3 & 4.3 \\
\bottomrule
\end{tabular}

\end{table}

\paragraph{Overview of results.}
ProtoSSL was consistently preferred over SupProto Direct in both prototype quality and prototype-case explanation tasks (\autoref{tab:user_study}, \autoref{tab:user_study_pref}). While inter-rater agreement was modest (Fleiss' $\kappa = 0.288$), preferences were directionally consistent across participants. Importantly, this preference arises despite SupProto Direct achieving higher predictive performance on PTB-XL (macro AUROC: 0.900 [0.899-0.902] vs.\ 0.881 [0.879-0.883]). As all evaluated cases were restricted to correct predictions for both models, the differences in \autoref{tab:user_study} and \autoref{tab:user_study_pref} reflect properties of the explanations themselves rather than classification accuracy.

\paragraph{Interpreting the preference.}
A possible explanation for this pattern is that supervision encourages prototypes to capture features that improve class discrimination, even when those features correspond to less distinct or less representative instances of a diagnosis. In contrast, ProtoSSL learns prototypes prior to label alignment, which may favor patterns that more closely reflect recurring structure in the data distribution. This interpretation is consistent with prior observations that prototype quality depends on evaluator expertise \citep{sethi_protoecgnet_2025}. In the present study, participants were final-year medical students, whose judgments are likely more aligned with clear and representative examples of each diagnosis. It is possible that ProtoSSL may better preserve such examples, whereas SupProto Direct may retain patterns that are discriminative but less consistently perceived as prototypical.

\paragraph{Qualitative comparison.}
To better understand the differences reflected in the user study, we examine a representative example case.

\begin{figure}[tp]
    \centering
    \fbox{
        \includegraphics[width=0.63\linewidth]{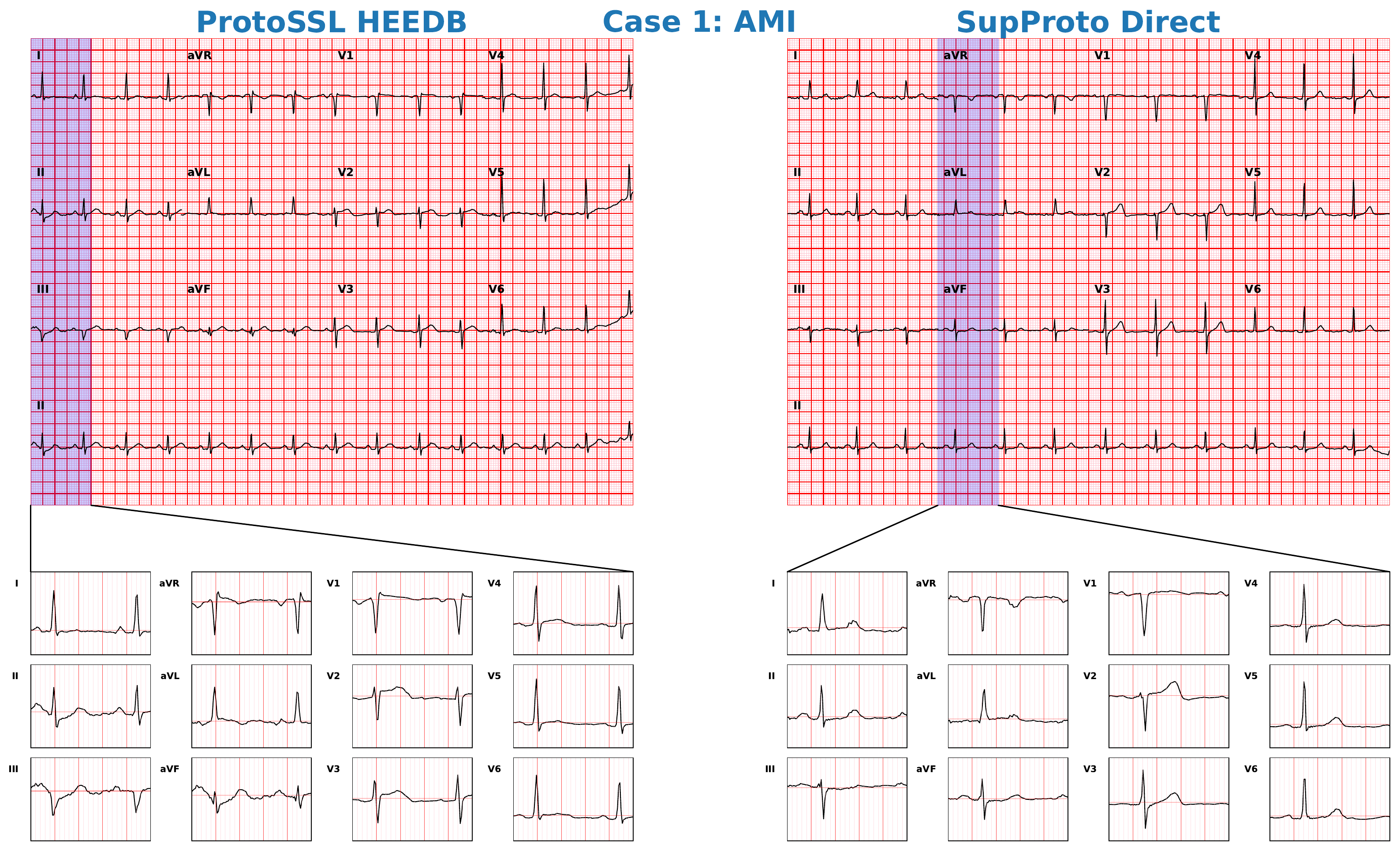}
    }
    \vspace{0.5em}
    \fbox{
        \includegraphics[width=0.63\linewidth]{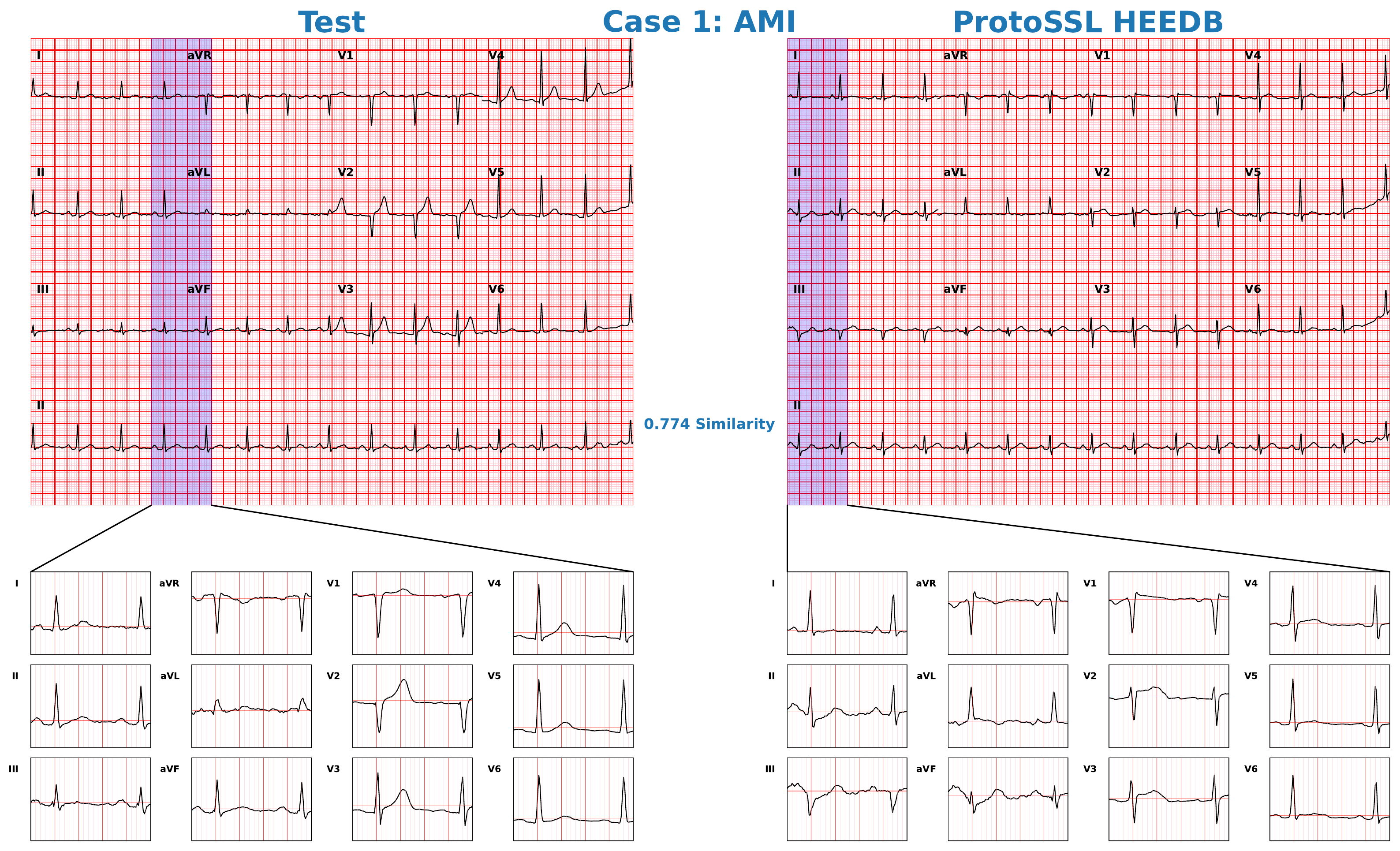}
    }
    \vspace{0.5em}
    \fbox{
        \includegraphics[width=0.63\linewidth]{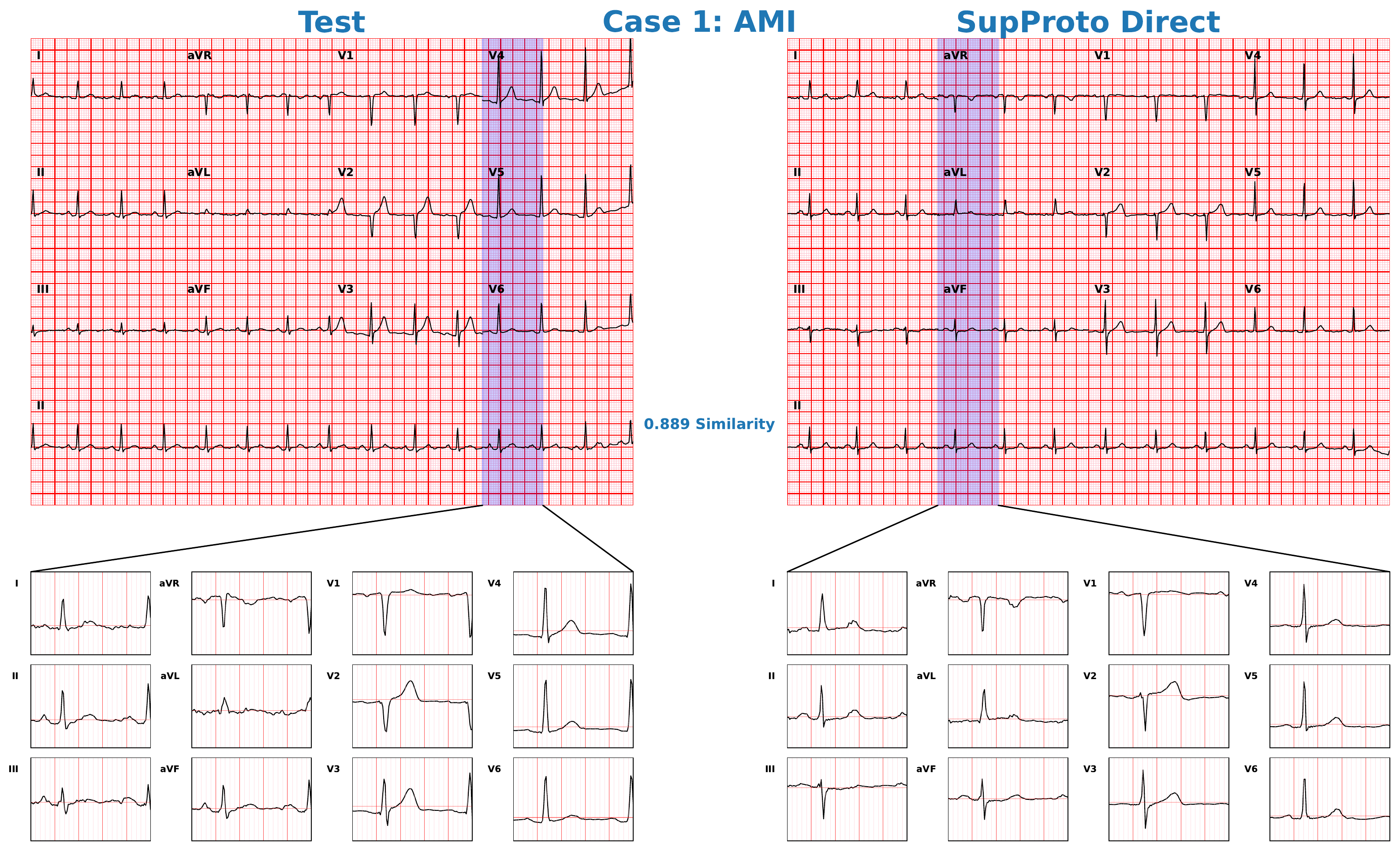}
    }
    \caption{\textbf{Representative example of anterior myocardial infarction (AMI).} Top: prototype visualization. Middle: ProtoSSL explanation for the test ECG. Bottom: SupProto Direct explanation. In all panels, highlighted regions correspond to the 1-second interval used for matching, with synchronized full-lead views showing all 12 leads over the selected window.}
    \label{fig:qualitative_prototypes}
\end{figure}

\autoref{fig:qualitative_prototypes} shows a case of anterior myocardial infarction (AMI). Both models produce prototypes that exhibit ST-segment elevation in the anteroseptal precordial leads (V1-V3), consistent with an anterior infarction pattern. The highlighted interval indicates the segment used for matching, and the corresponding full-lead view allows inspection of morphology across all 12 leads.

The two prototypes capture the same diagnostic pattern but differ in how distinctly it is expressed. The ProtoSSL prototype shows clearer ST-segment elevation across the anterior leads, whereas the SupProto Direct prototype presents a less pronounced instance of the same morphology.

This difference is reflected in participant responses. All participants (7/7) judged the ProtoSSL prototype to be a good example, compared to 5/7 for SupProto Direct; in direct comparison, 5/7 selected \enquote{both} and 2/7 selected ProtoSSL. For paired explanations, both models were generally judged favorably (ProtoSSL: 6/7; SupProto Direct: 7/7), with most participants selecting \enquote{both} (4/7), and the remaining responses split between ProtoSSL (2/7) and SupProto Direct (1/7).

This example reflects the distinction observed across the full study. Both models identify and align the relevant morphology, and in this case both produce paired explanations that are judged favorably. The primary difference appears at the level of the standalone prototypes, where the ProtoSSL example is more consistently judged to be a clear representation of the diagnosis. In aggregate, this distinction is reflected in higher acceptability for ProtoSSL across both prototype quality and paired explanation tasks. Additional representative cases for PVC, RBBB, and LBBB are provided in Supplementary Figs. \ref{fig:pvc}, \ref{fig:rbbb}, and \ref{fig:lbbb}, respectively.

\paragraph{Summary.}
Taken together, these results highlight a distinction between discriminative performance and perceived interpretability. While supervised prototype models optimize for classification accuracy, self-supervised prototype learning appears to preserve patterns that are more consistently recognized as representative by human evaluators. This provides a plausible explanation for why ProtoSSL is preferred in human evaluation despite slightly lower predictive performance.

\section{Extended Limitations}
\label{app:limitations}
ProtoSSL reduces dependence on task-specific label supervision, but foundational prototype learning is still shaped by the unlabeled pretraining distribution and the self-supervised view construction. This differs from label-supervised prototype pretraining, where prototypes are explicitly optimized around a fixed source label space; however, ProtoSSL's latent prototype bank can only reuse structure that is sufficiently represented and learnable in the pretraining corpus, consistent with the known limitations of self-supervised methods \citep{chen_intriguing_2021, tian_what_2020}. As a result, frozen transfer is expected to be strongest when downstream tasks are well aligned with the structure captured during pretraining, while fine-tuning may be needed when the downstream task emphasizes different factors. This distinction is reflected in our experiments: ECG transfer is comparatively consistent across downstream tasks (\autoref{tab:ecg_full}), likely reflecting that all ECG labels have a physiologic basis, whereas audio transfer is more heterogeneous across environmental sound, emotion, and speaker-identification tasks (\autoref{tab:audio_results}). However, even where frozen transfer is less effective, fine-tuning ProtoSSL provides gains over SupProto Direct, suggesting that the pretrained prototype bank remains a useful initialization for downstream adaptation. Accordingly, while frozen ProtoSSL prototypes cannot transfer to waveform morphologies that are novel relative to the pretraining corpus (e.g., rare morphologies that are absent or severely underrepresented in HEEDB), we hypothesize that expanding the pretraining prototype bank may better capture less well represented morphologies (\autoref{app:num_proto_upstream}). Further characterizing transfer to such underrepresented morphologies is an important direction for future work.

ProtoSSL also requires labeled target data for assignment, projection, and classifier training. The method avoids learning a new prototype bank from scratch, but label-supervised projection still depends on the availability and quality of positive examples for each downstream label. This dependence becomes especially important in extreme low-data regimes: although the pretrained prototypes are learned from a large unlabeled corpus, the final projected exemplar must be selected from the finite labeled target set. When few positive examples are available, the nearest label-consistent projection target may be atypical, noisy, or representative of only one manifestation of a heterogeneous label. This is a general tradeoff of projection-grounded prototype methods: grounding prototypes in real examples makes them inspectable (by ensuring that exemplars actually belong to a target class), but the quality and coverage of those explanations depend on the labeled examples available for downstream alignment \citep{barnett_case-based_2021, davoodi_interpretability_2023, carloni_applicability_2023}. In ProtoSSL, label-supervised projection mitigates this risk by requiring projected exemplars to be positive for the assigned label, and multiple prototypes per label can capture distinct motifs when sufficient examples are available; however, projected explanations in severe label-scarce settings should still be interpreted with caution.

ProtoSSL also imposes a capacity tradeoff when transferring to downstream tasks with many labels. Given a pretrained prototype bank of size $K$, assigning $M$ prototypes to each of $L$ downstream labels requires $K \geq L \times M$ (as discussed in \autoref{sec:assign_main}). For tasks with a modest number of labels, this constraint is not limiting, but it becomes important for many-class problems. For example, in VoxCeleb1 speaker identification, the downstream task contains 1251 classes; to conduct our audio experiments with 5 prototypes per label, we had to expand our ProtoSSL AudioSet-pretrained bank to 6324 prototypes. Thus, deploying ProtoSSL on tasks with very large label spaces requires either increasing the pretrained prototype-bank size, or reducing the number of prototypes allocated to each label, which may limit the diversity and coverage of class-specific explanations. While relaxing the assignment constraint (that a pretrained prototype can only be assigned to one label) may help address this tradeoff, such a variant of the assignment problem does not admit an efficient algorithm for solving the LP. Further work is needed to investigate relaxing the assignment constraint while preserving assignment efficiency.

Relatedly, the assignment LP has no formal ``reject'' option: every label-slot must be assigned a prototype, even when all prototype-label associations for that label are weak (e.g., when the label corresponds to structure that was not captured during pretraining). Several aspects of our approach mitigate this risk. First, each prototype–label score is a standardized difference between prototype activation in label-positive and label-negative examples. LAP then finds the globally optimal one-to-one allocation across all label slots, while allowing most members of the pretrained bank to remain unused. Thus, a prototype is selected because its activation separates positive from negative examples for that label relative to the available alternatives, not merely because it has no association with other labels. Label-supervised projection subsequently grounds the selected member in a real target example carrying the assigned label. Third, the final classifier is trained over all projected prototype activations, and our coefficient analysis (\autoref{app:classifier_explanation}) shows that target-assigned prototypes receive systematically more positive weight than non-target prototypes, indicating that assigned prototypes are used as intended by the classifier. Nonetheless, we have not systematically characterized whether any of our downstream labels are so weakly associated that assignment would yield a projected exemplar that is label-consistent but only loosely related to the label's defining features. Formal investigation of this setting and incorporating an explicit abstention mechanism is a natural extension.

In multilabel settings, prototype-label assignment should be interpreted as downstream alignment rather than exclusive semantic ownership. Related labels may share motifs, and prototypes assigned to one label may still provide useful evidence or counter-evidence for another \citep{sethi_protoecgnet_2025}. We therefore use a regularized logistic-regression classifier over projected prototype activations rather than a strict ProtoPNet-style class mask (further discussion in \autoref{app:classifier_explanation}). This preserves a prototype-based prediction mechanism, but individual displayed prototypes should be understood as case-based evidence strongly contributing to a prediction, not as the only evidence used by the classifier.

Our ECG implementation also has limited temporal modeling. Following the partial prototype design described in \autoref{app:ecg_details}, each prototype is matched against one-second windows of the input, and prototype activations are max pooled across windows. This design captures localized waveform morphology well, but it does not explicitly model rhythm, the relative timing of events across beats, or transitions between windows. As a result, labels defined primarily by rhythm or by temporal relationships spanning more than one second (e.g. AFib) can only be captured to the extent that they manifest within a single window. ProtoECGNet \citep{sethi_protoecgnet_2025} addresses this with a multi-branch architecture that pairs time-localized morphology prototypes with global prototypes for rhythm and diffuse abnormalities. In this work, we use a single 1-second 2D partial prototype branch for all labels, in order to isolate the effect of self-supervised prototype discovery. Extending ProtoSSL to a multi-branch design, with self-supervised prototype banks learned at multiple temporal scales and assigned to the branches appropriate for each downstream label, is a promising direction for future work.

Finally, our human evaluation was designed to isolate explanation quality rather than clinical utility under deployment. For projection-based prototype models, whether a human judges a prototype to be similar to an inference target is an empirical property that must be measured rather than assumed from the model's architecture \citep{davoodi_interpretability_2023}. Our estimate of this property is also specific to the models, raters, and diagnoses we studied: seven final-year medical students from a single institution evaluated four morphology-based diagnoses on PTB-XL, and results may differ for other models, more experienced clinicians, or other diagnoses.

We further restricted evaluation to cases correctly predicted by both models. This allowed a paired comparison of prototype quality and prototype-test matching without confounding explanation judgments by prediction errors: had we included cases where only one model was correct, differences in ratings would reflect a mix of explanation quality and predictive accuracy, and raters could penalize an explanation for supporting an incorrect diagnosis even when it faithfully reflected the model's reasoning. This design estimates explanation quality conditional on correct predictions and may overestimate the absolute quality of explanations encountered in practice. Incorrect predictions represent a distinct setting: a prototype may faithfully reveal the model's erroneous reasoning while still being clinically inappropriate for the true diagnosis. Evaluating whether users can recognize and appropriately discount such explanations would require a separate study focused on model errors and clinical decision support.

Importantly, acceptability as judged by raters is not clinical utility. Establishing clinical utility would require a decision-support study measuring whether prototype-based explanations improve clinician performance, as in \citet{barnett_improving_2024}. Future work should therefore evaluate ProtoSSL explanations on incorrect and ambiguous cases, and assess whether they support calibrated reliance, i.e., whether clinicians appropriately trust correct predictions and discount incorrect ones. The present study therefore supports the narrower conclusion that, among jointly correct predictions, ProtoSSL produces prototypes and prototype-based explanations that are more often judged acceptable than a model trained via label-supervision by medically trained evaluators.

\clearpage
\begin{figure}[tp]
    \centering
    \fbox{
        \includegraphics[width=0.63\linewidth]{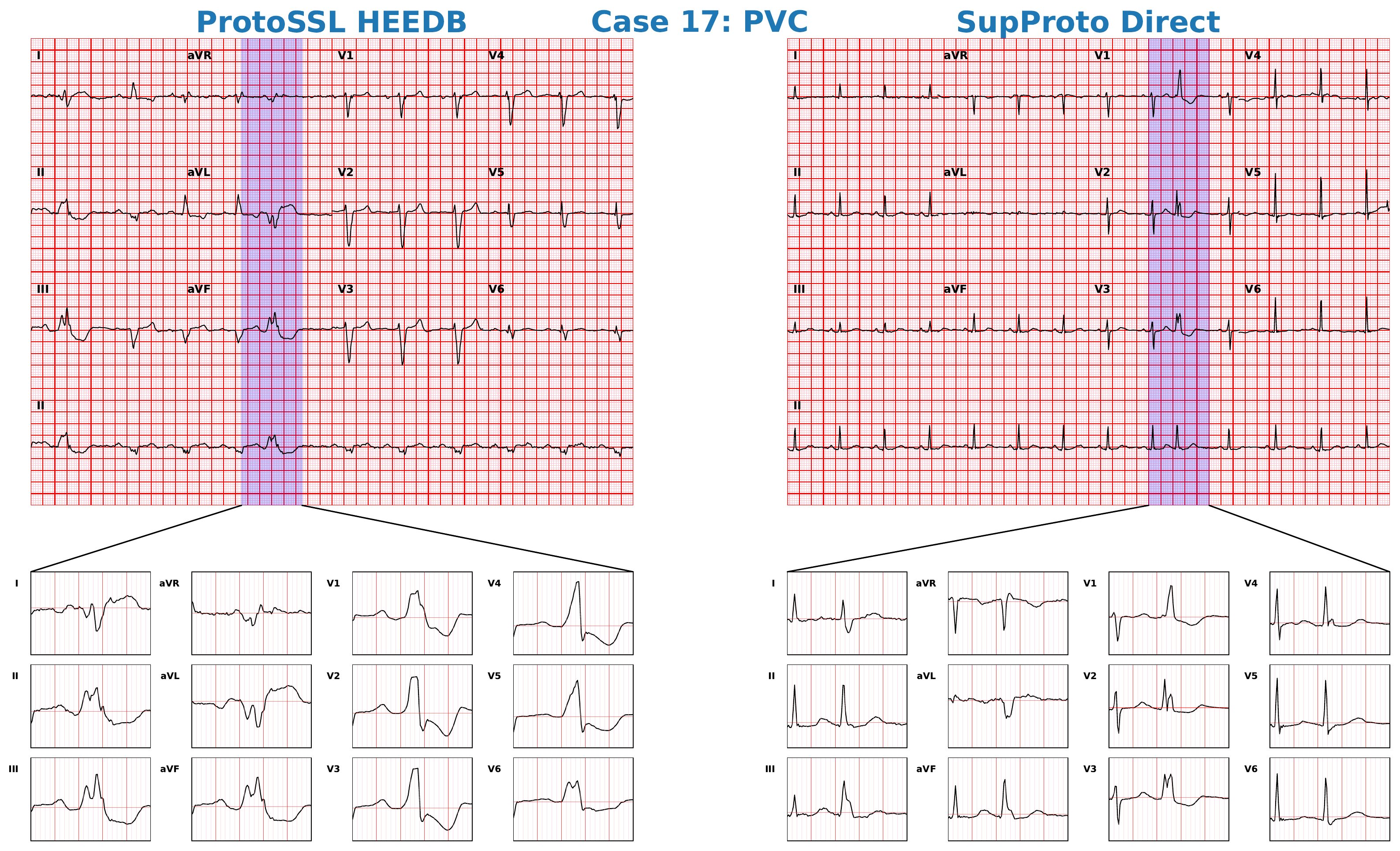}
    }
    \vspace{0.5em}
    \fbox{
        \includegraphics[width=0.63\linewidth]{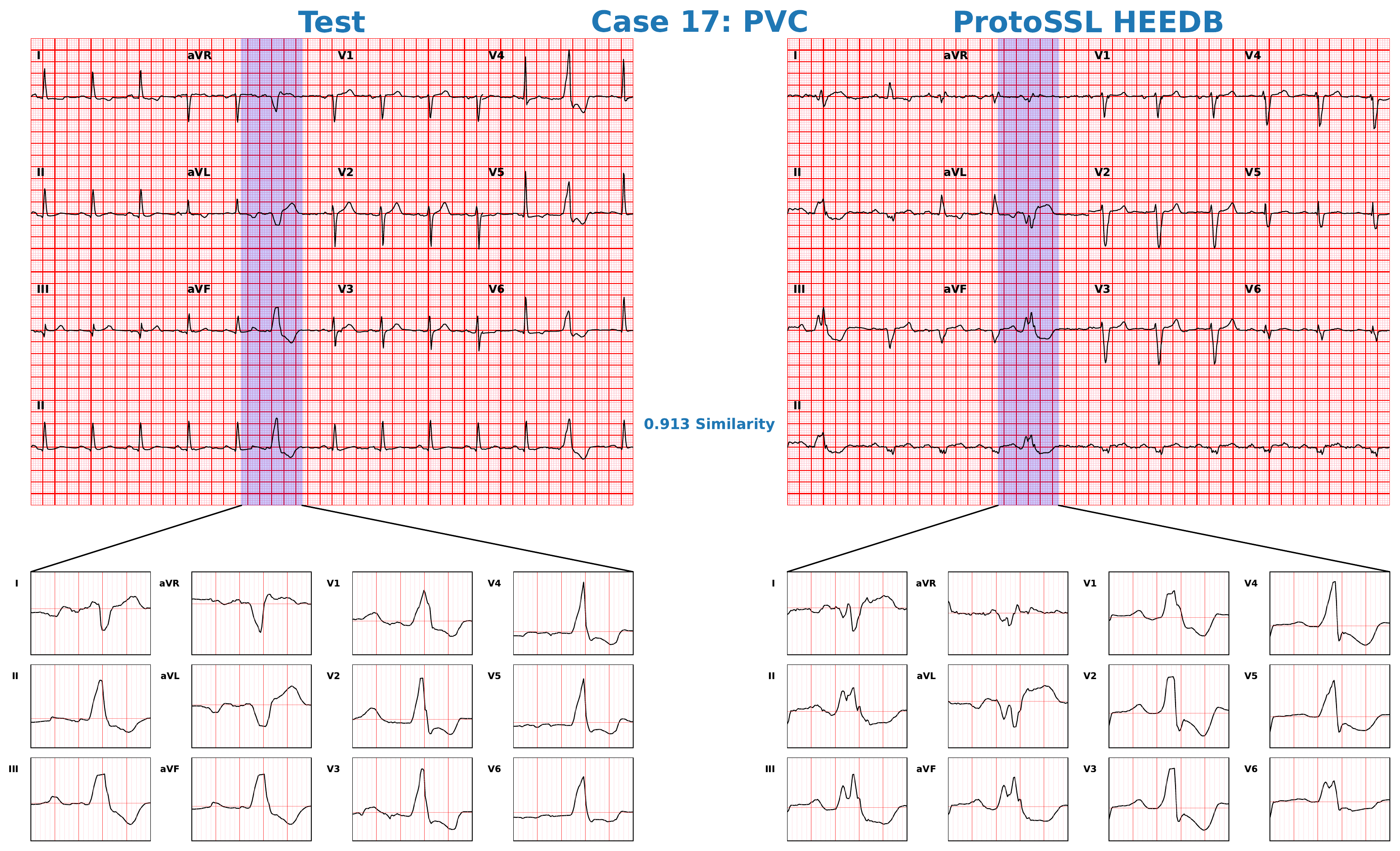}
    }
    \vspace{0.5em}
    \fbox{
        \includegraphics[width=0.63\linewidth]{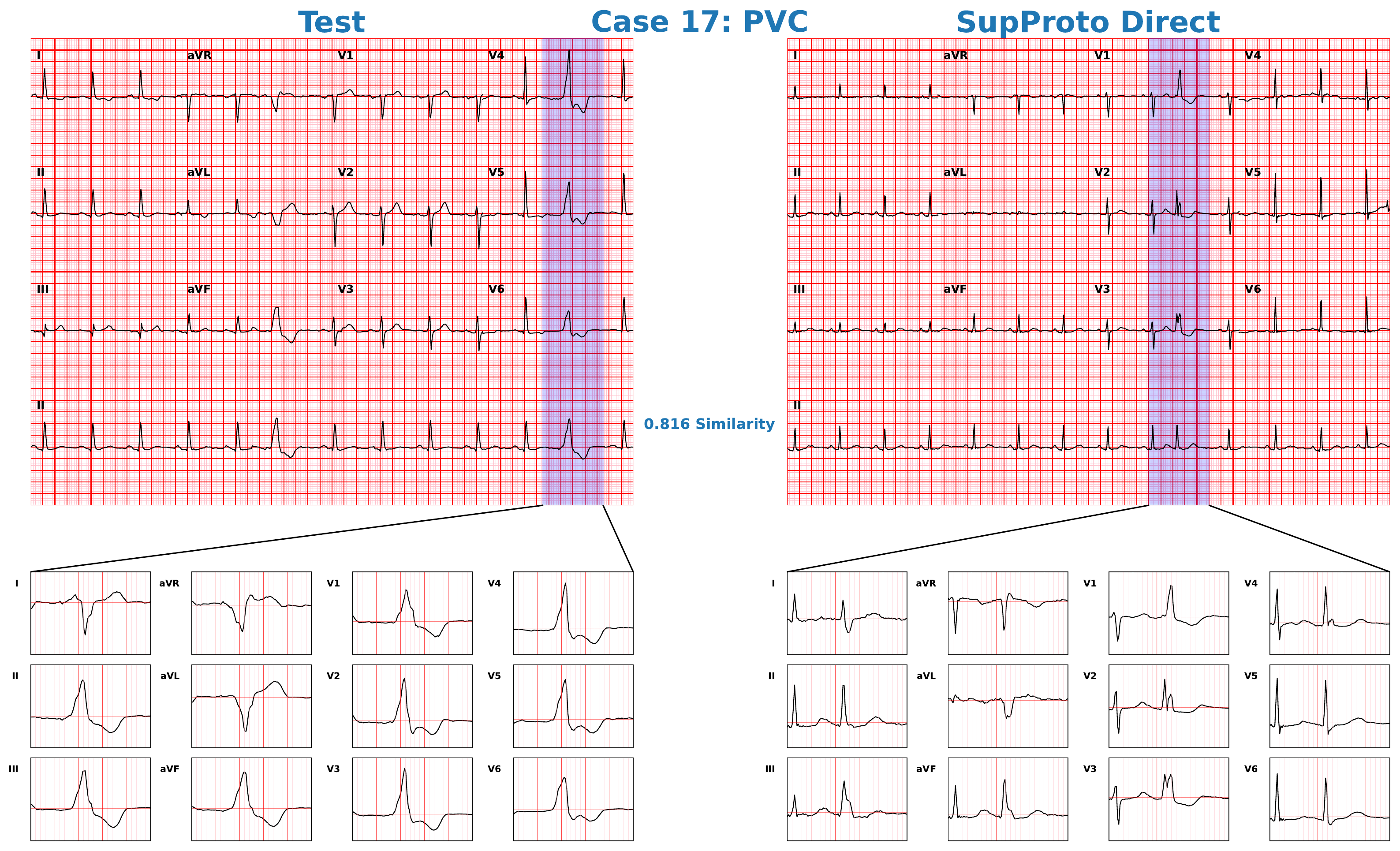}
    }
    \caption{\textbf{Representative example of premature ventricular contraction (PVC).} Top: prototype visualization. Middle: ProtoSSL explanation for the test ECG. Bottom: SupProto Direct explanation. In all panels, highlighted regions correspond to the 1-second interval used for matching, with synchronized full-lead views showing all 12 leads over the selected window.}
    \label{fig:pvc}
\end{figure}
\begin{figure}[tp]
    \centering
    \fbox{
        \includegraphics[width=0.63\linewidth]{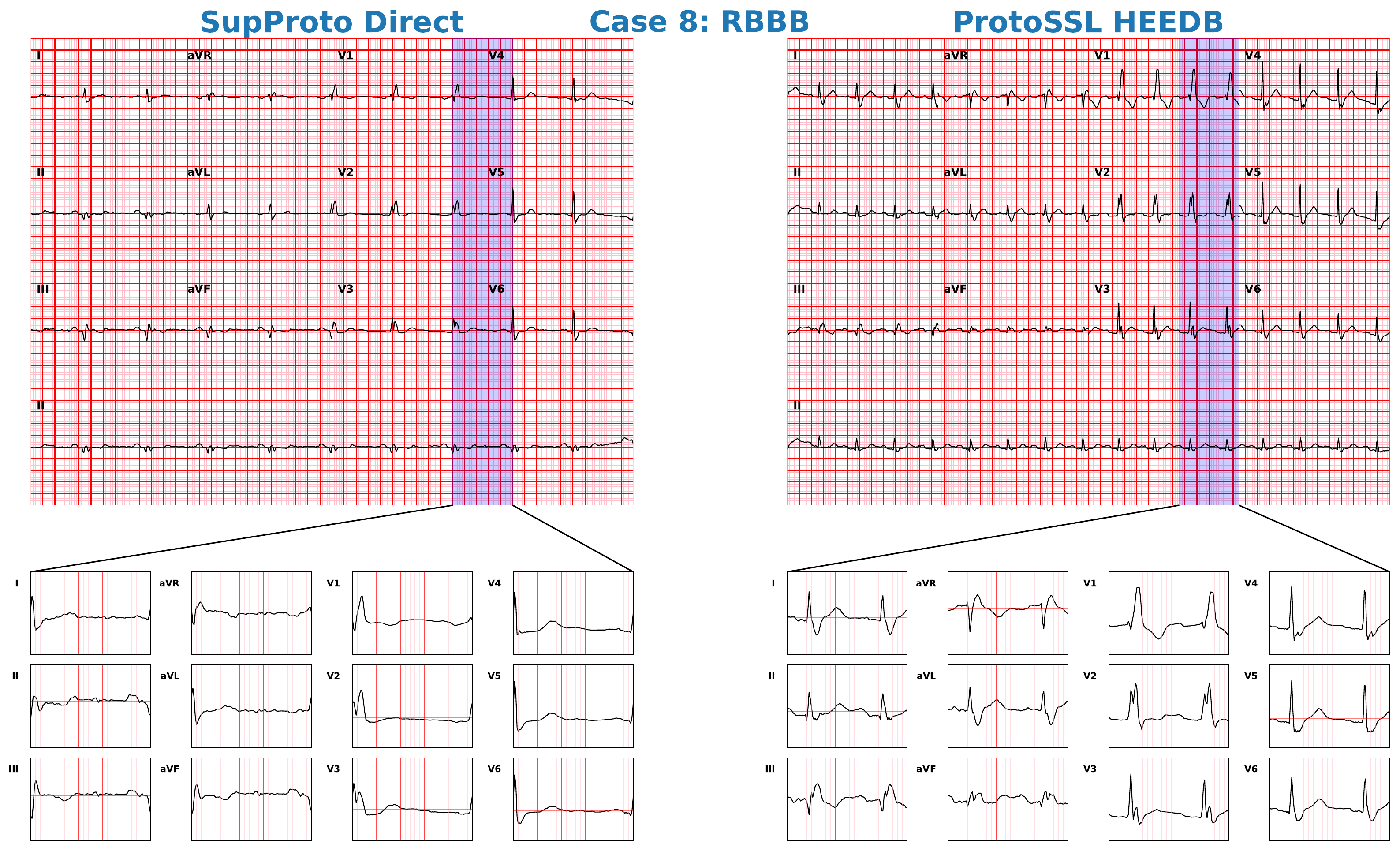}
    }
    \vspace{0.5em}
    \fbox{
        \includegraphics[width=0.63\linewidth]{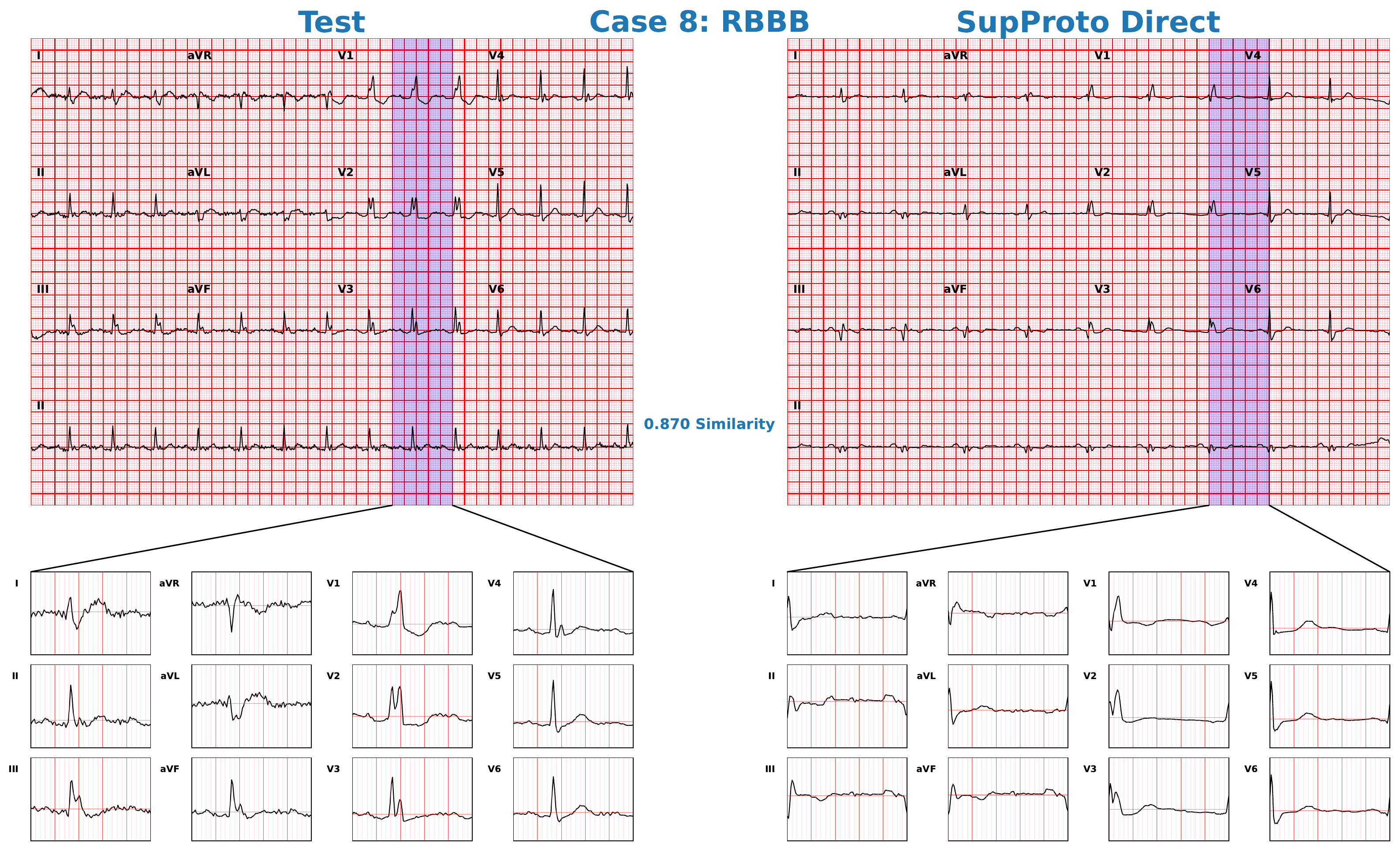}
    }
    \vspace{0.5em}
    \fbox{
        \includegraphics[width=0.63\linewidth]{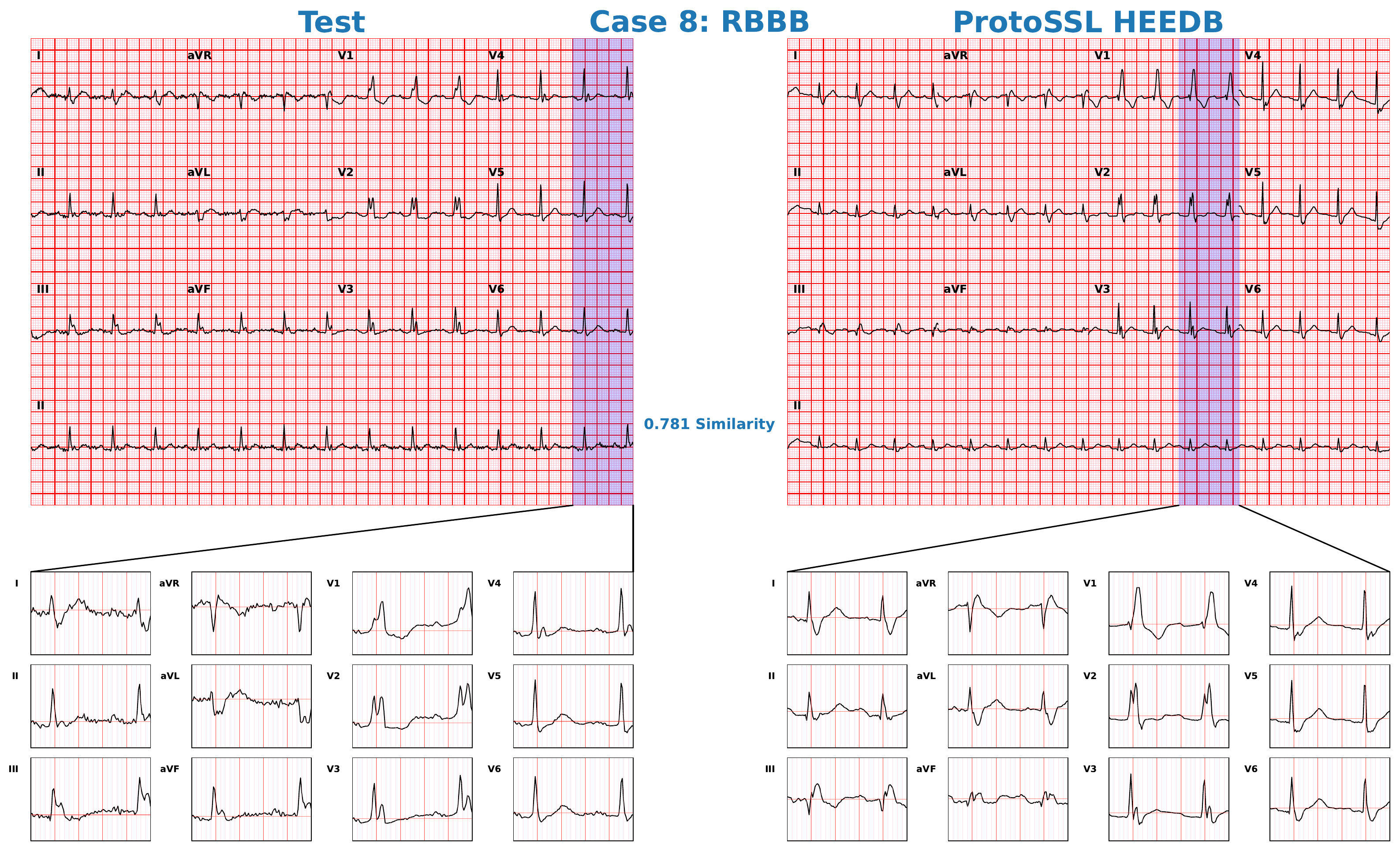}
    }
    \caption{\textbf{Representative example of right bundle branch block (RBBB).} Top: prototype visualization. Middle: SupProto Direct explanation for the test ECG. Bottom: ProtoSSL explanation. In all panels, highlighted regions correspond to the 1-second interval used for matching, with synchronized full-lead views showing all 12 leads over the selected window.}
    \label{fig:rbbb}
\end{figure}
\begin{figure}[tp]
    \centering
    \fbox{
        \includegraphics[width=0.63\linewidth]{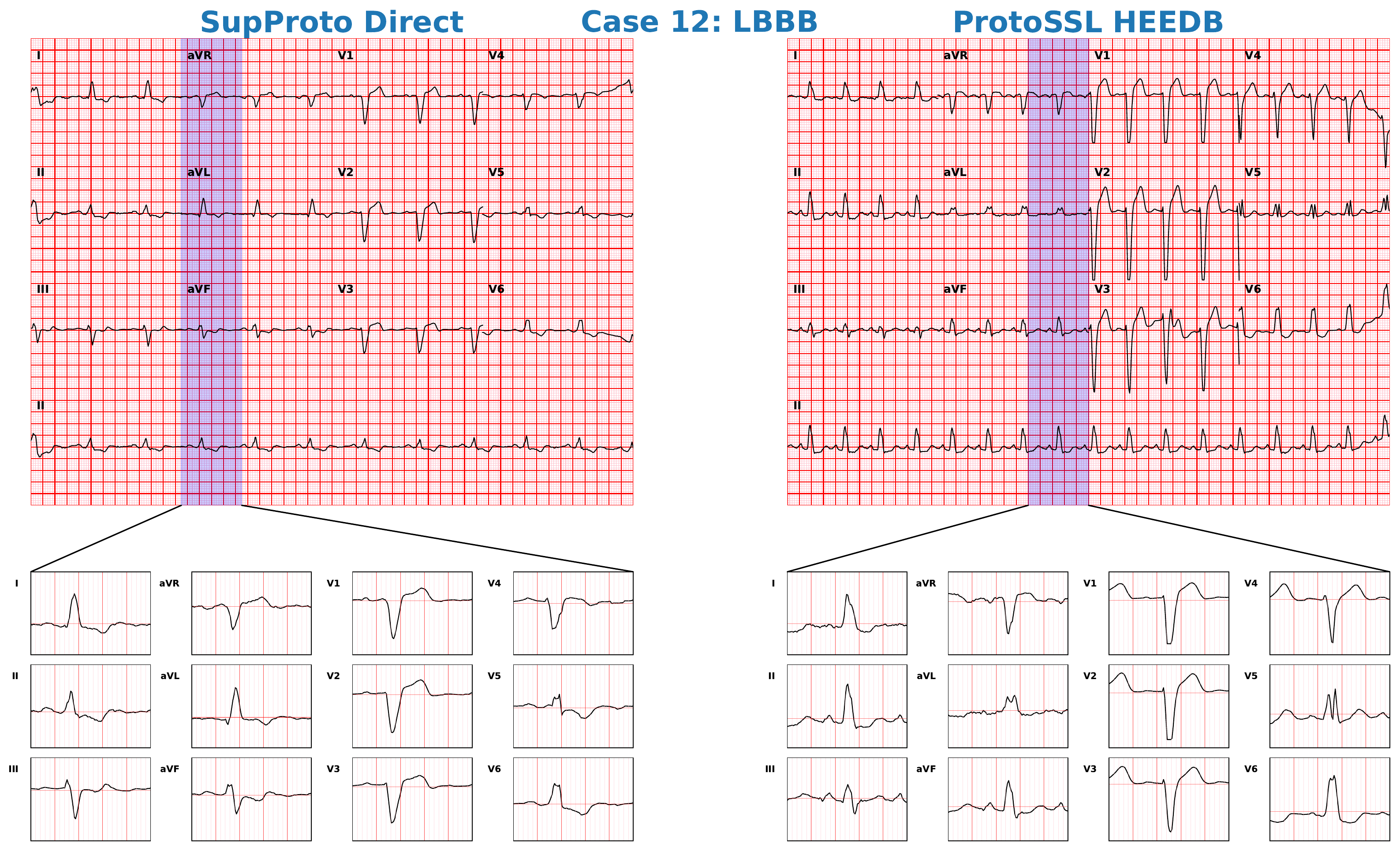}
    }
    \vspace{0.5em}
    \fbox{
        \includegraphics[width=0.63\linewidth]{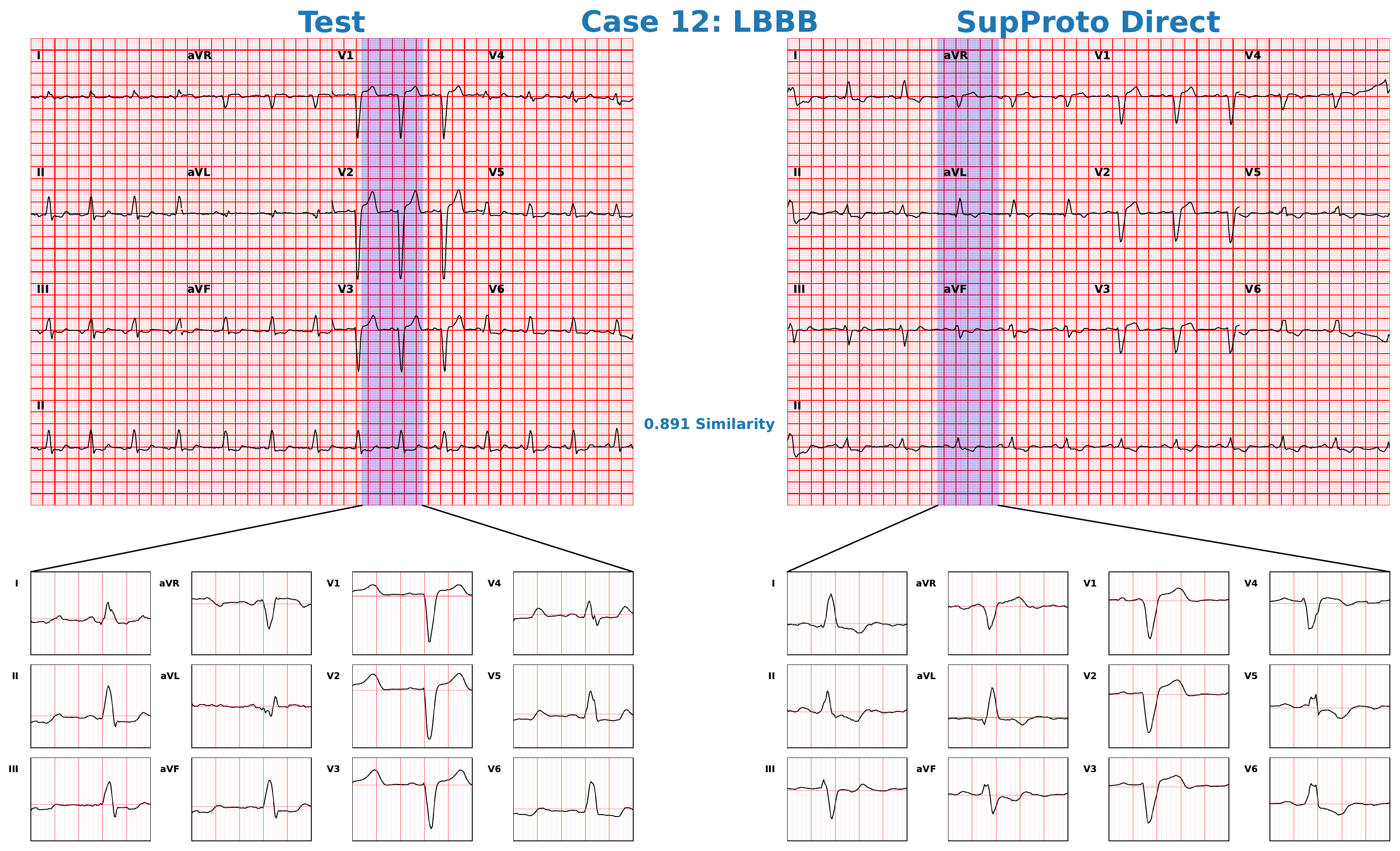}
    }
    \vspace{0.5em}
    \fbox{
        \includegraphics[width=0.63\linewidth]{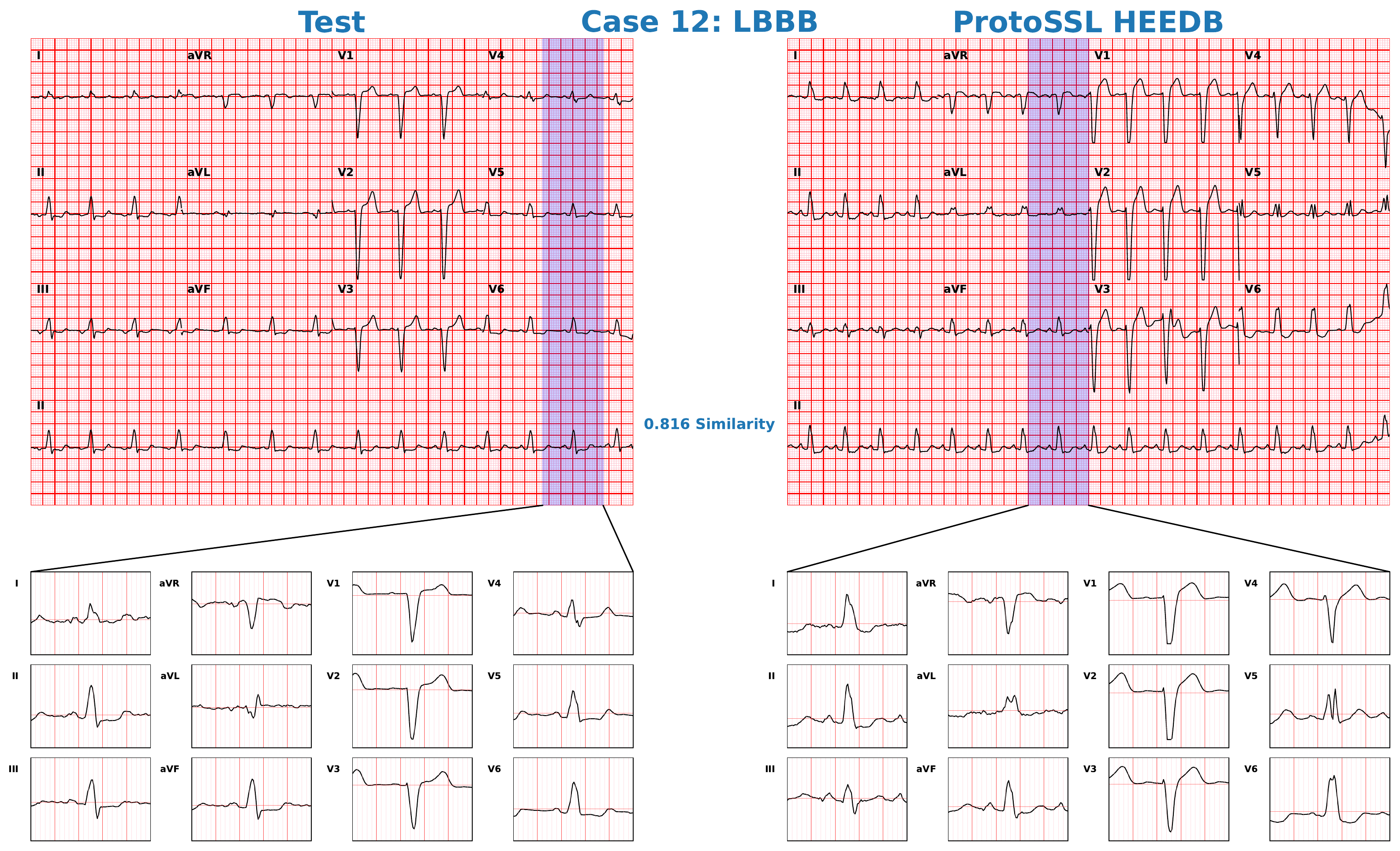}
    }
    \caption{\textbf{Representative example of left bundle branch block (LBBB).} Top: prototype visualization. Middle: SupProto Direct explanation for the test ECG. Bottom: ProtoSSL explanation. In all panels, highlighted regions correspond to the 1-second interval used for matching, with synchronized full-lead views showing all 12 leads over the selected window.}
    \label{fig:lbbb}
\end{figure}

\clearpage
\begin{table}[tp]
\centering
\caption{\textbf{HEEDB ECG dataset characteristics.}}
\label{tab:ecg_dataset_heedb}
{
\setlength{\tabcolsep}{4pt}
\begin{adjustbox}{max totalsize={\textwidth}{0.9\textheight}}
\begin{tabular}{lllll}
\toprule
Split & & Train & Val & Test \\
\midrule
& ECGs, n & 8145485 & 556578 & 368191 \\
& Patients, n & 1566388 & 122408 & 105578 \\
\midrule
& Age, median [Q1, Q3] & 64.5 [52.4, 75.6] & 65.8 [53.4, 75.7] & 65.3 [52.2, 74.9] \\
\midrule
\multirow{2}{*}{Sex} & Female, n (\%) & 3803314 (46.7) & 241867 (43.5) & 183188 (49.8) \\
& Male, n (\%) & 4342171 (53.3) & 314711 (56.5) & 185003 (50.2) \\
\midrule
\multirow{2}{*}{Site} & Emory, n (\%) & 967730 (11.9) & 0 (0.0) & 0 (0.0) \\
& Harvard, n (\%) & 7177755 (88.1) & 556578 (100.0) & 368191 (100.0) \\
\midrule
\multirow{20}{*}{Label} & Anterior Infarct, n (\%) & 100103 (1.2) & 4337 (0.8) & 37 (0.0) \\
& Atrial Fibrillation, n (\%) & 621457 (7.6) & 23311 (4.2) & 35968 (9.8) \\
& Atrial Flutter, n (\%) & 113719 (1.4) & 2671 (0.5) & 7871 (2.1) \\
& Atrial-Paced Rhythm, n (\%) & 30198 (0.4) & 1693 (0.3) & 3662 (1.0) \\
& Incomplete Right Bundle Branch Block, n (\%) & 220539 (2.7) & 8451 (1.5) & 13138 (3.6) \\
& Inferior Infarct, n (\%) & 95826 (1.2) & 399 (0.1) & 47 (0.0) \\
& Lateral Infarct, n (\%) & 86382 (1.1) & 1358 (0.2) & 8430 (2.3) \\
& Left Bundle Branch Block, n (\%) & 188393 (2.3) & 6167 (1.1) & 9058 (2.5) \\
& Normal Sinus Rhythm, n (\%) & 3259334 (40.0) & 111663 (20.1) & 181334 (49.2) \\
& Premature Atrial Complexes, n (\%) & 269759 (3.3) & 9149 (1.6) & 21102 (5.7) \\
& Premature Ventricular Complexes, n (\%) & 373229 (4.6) & 10775 (1.9) & 27406 (7.4) \\
& Right Axis Deviation, n (\%) & 118452 (1.5) & 4548 (0.8) & 14071 (3.8) \\
& Right Bundle Branch Block, n (\%) & 410314 (5.0) & 16640 (3.0) & 24078 (6.5) \\
& Sinus Bradycardia, n (\%) & 827792 (10.2) & 29481 (5.3) & 37519 (10.2) \\
& Sinus Rhythm, n (\%) & 1074780 (13.2) & 59826 (10.7) & 65658 (17.8) \\
& Sinus Tachycardia, n (\%) & 573851 (7.0) & 27260 (4.9) & 39670 (10.8) \\
& Ventricular Tachycardia, n (\%) & 14334 (0.2) & 2467 (0.4) & 154 (0.0) \\
& Ventricular-Paced Rhythm, n (\%) & 54036 (0.7) & 3891 (0.7) & 4744 (1.3) \\
& With 1st Degree AV Block, n (\%) & 311291 (3.8) & 398 (0.1) & 23564 (6.4) \\
& With Sinus Arrhythmia, n (\%) & 220917 (2.7) & 4856 (0.9) & 15162 (4.1) \\
\bottomrule
\end{tabular}
\end{adjustbox}
}
\end{table}


\begin{table}[tp]
\centering
\caption{\textbf{MIMIC-IV ECG dataset characteristics.}}
\label{tab:ecg_dataset_mimic}
{
\setlength{\tabcolsep}{4pt}
\begin{adjustbox}{max totalsize={\textwidth}{0.9\textheight}}
\begin{tabular}{lllllllllllll}
\toprule
Split & & \multicolumn{9}{c}{Train (Nested Subsets)} & Val & Test \\
\midrule
& ECGs, n & 78470 & 32768 & 16384 & 8192 & 4096 & 2048 & 1024 & 512 & 256 & 3900 & 4126 \\
& Patients, n & 46549 & 24229 & 13628 & 7324 & 3860 & 1976 & 1000 & 504 & 253 & 2568 & 2675 \\
\midrule
& \makecell[l]{Age,\\median [Q1, Q3]} & \makecell[l]{65.0\\{}[51.0, 77.0]} & \makecell[l]{65.0\\{}[51.0, 77.0]} & \makecell[l]{65.0\\{}[51.0, 77.0]} & \makecell[l]{65.0\\{}[51.0, 77.0]} & \makecell[l]{65.0\\{}[51.0, 77.0]} & \makecell[l]{65.0\\{}[51.0, 77.0]} & \makecell[l]{65.0\\{}[51.0, 77.0]} & \makecell[l]{65.0\\{}[52.0, 78.0]} & \makecell[l]{65.0\\{}[52.8, 77.2]} & \makecell[l]{64.0\\{}[51.0, 76.0]} & \makecell[l]{65.0\\{}[50.0, 77.0]} \\
\midrule
\multirow{2}{*}{Sex} & Female, n (\%) & 41245 (52.6) & 17222 (52.6) & 8610 (52.6) & 4303 (52.5) & 2150 (52.5) & 1074 (52.4) & 535 (52.2) & 266 (52.0) & 132 (51.6) & 2072 (53.1) & 2152 (52.2) \\
& Male, n (\%) & 37225 (47.4) & 15546 (47.4) & 7774 (47.4) & 3889 (47.5) & 1946 (47.5) & 974 (47.6) & 489 (47.8) & 246 (48.0) & 124 (48.4) & 1828 (46.9) & 1974 (47.8) \\
\midrule
\multirow{46}{*}{Label} & Severe hypoxemia, n (\%) & 486 (0.6) & 215 (0.7) & 112 (0.7) & 53 (0.6) & 30 (0.7) & 15 (0.7) & 10 (1.0) & 8 (1.6) & 2 (0.8) & 21 (0.5) & 24 (0.6) \\
& ECMO, n (\%) & 61 (0.1) & 33 (0.1) & 17 (0.1) & 7 (0.1) & 5 (0.1) & 4 (0.2) & 3 (0.3) & 2 (0.4) & 1 (0.4) & 4 (0.1) & 3 (0.1) \\
& Vasopressors, n (\%) & 619 (0.8) & 256 (0.8) & 135 (0.8) & 67 (0.8) & 28 (0.7) & 11 (0.5) & 7 (0.7) & 4 (0.8) & 4 (1.6) & 28 (0.7) & 31 (0.8) \\
& Inotropes, n (\%) & 163 (0.2) & 75 (0.2) & 37 (0.2) & 19 (0.2) & 8 (0.2) & 5 (0.2) & 5 (0.5) & 3 (0.6) & 3 (1.2) & 10 (0.3) & 7 (0.2) \\
& Mechanical ventilation, n (\%) & 1907 (2.4) & 814 (2.5) & 394 (2.4) & 184 (2.2) & 94 (2.3) & 50 (2.4) & 28 (2.7) & 21 (4.1) & 12 (4.7) & 84 (2.2) & 92 (2.2) \\
& Cardiac arrest, n (\%) & 120 (0.2) & 58 (0.2) & 31 (0.2) & 15 (0.2) & 9 (0.2) & 4 (0.2) & 2 (0.2) & 2 (0.4) & 2 (0.8) & 3 (0.1) & 4 (0.1) \\
& ICU admit within 24 hours, n (\%) & 9582 (12.2) & 4056 (12.4) & 2064 (12.6) & 1012 (12.4) & 520 (12.7) & 266 (13.0) & 146 (14.3) & 83 (16.2) & 46 (18.0) & 439 (11.3) & 497 (12.0) \\
& ICU admit, n (\%) & 11702 (14.9) & 4974 (15.2) & 2518 (15.4) & 1237 (15.1) & 624 (15.2) & 322 (15.7) & 177 (17.3) & 95 (18.6) & 52 (20.3) & 540 (13.8) & 599 (14.5) \\
& Mortality 24 hours, n (\%) & 373 (0.5) & 165 (0.5) & 90 (0.5) & 43 (0.5) & 25 (0.6) & 13 (0.6) & 6 (0.6) & 5 (1.0) & 4 (1.6) & 19 (0.5) & 18 (0.4) \\
& Mortality 7 days, n (\%) & 1211 (1.5) & 509 (1.6) & 262 (1.6) & 124 (1.5) & 65 (1.6) & 35 (1.7) & 18 (1.8) & 11 (2.1) & 8 (3.1) & 53 (1.4) & 57 (1.4) \\
& Mortality 28 days, n (\%) & 3038 (3.9) & 1286 (3.9) & 651 (4.0) & 320 (3.9) & 163 (4.0) & 92 (4.5) & 47 (4.6) & 28 (5.5) & 19 (7.4) & 133 (3.4) & 145 (3.5) \\
& Mortality 90 days, n (\%) & 5836 (7.4) & 2450 (7.5) & 1222 (7.5) & 610 (7.4) & 312 (7.6) & 172 (8.4) & 85 (8.3) & 52 (10.2) & 32 (12.5) & 262 (6.7) & 291 (7.1) \\
& Mortality 180 days, n (\%) & 8176 (10.4) & 3437 (10.5) & 1724 (10.5) & 862 (10.5) & 447 (10.9) & 241 (11.8) & 119 (11.6) & 74 (14.5) & 46 (18.0) & 368 (9.4) & 408 (9.9) \\
& Mortality 365 days, n (\%) & 11360 (14.5) & 4711 (14.4) & 2322 (14.2) & 1163 (14.2) & 609 (14.9) & 325 (15.9) & 163 (15.9) & 95 (18.6) & 55 (21.5) & 552 (14.2) & 590 (14.3) \\
& Mortality stay, n (\%) & 50 (0.1) & 21 (0.1) & 14 (0.1) & 6 (0.1) & 3 (0.1) & 2 (0.1) & 1 (0.1) & 1 (0.2) & 1 (0.4) & 3 (0.1) & 1 (0.0) \\
& Cardiac diagnosis, n (\%) & 52827 (67.3) & 22048 (67.3) & 10988 (67.1) & 5460 (66.7) & 2744 (67.0) & 1376 (67.2) & 700 (68.4) & 358 (69.9) & 180 (70.3) & 2553 (65.5) & 2697 (65.4) \\
& Non-cardiac diagnosis, n (\%) & 77637 (98.9) & 32411 (98.9) & 16214 (99.0) & 8108 (99.0) & 4052 (98.9) & 2026 (98.9) & 1012 (98.8) & 506 (98.8) & 253 (98.8) & 3875 (99.4) & 4084 (99.0) \\
& Septicemia, n (\%) & 3366 (4.3) & 1455 (4.4) & 710 (4.3) & 337 (4.1) & 183 (4.5) & 88 (4.3) & 45 (4.4) & 26 (5.1) & 12 (4.7) & 137 (3.5) & 201 (4.9) \\
& Gram negative septicemia, n (\%) & 959 (1.2) & 422 (1.3) & 219 (1.3) & 105 (1.3) & 55 (1.3) & 31 (1.5) & 12 (1.2) & 7 (1.4) & 5 (2.0) & 45 (1.2) & 52 (1.3) \\
& Protein-calorie malnutrition, n (\%) & 1971 (2.5) & 853 (2.6) & 448 (2.7) & 222 (2.7) & 113 (2.8) & 68 (3.3) & 34 (3.3) & 20 (3.9) & 12 (4.7) & 89 (2.3) & 94 (2.3) \\
& Severe protein-calorie malnutrition, n (\%) & 1119 (1.4) & 473 (1.4) & 254 (1.6) & 118 (1.4) & 60 (1.5) & 35 (1.7) & 23 (2.2) & 14 (2.7) & 9 (3.5) & 54 (1.4) & 54 (1.3) \\
& Disorders of phosphorus metabolism, n (\%) & 1136 (1.4) & 479 (1.5) & 241 (1.5) & 136 (1.7) & 60 (1.5) & 24 (1.2) & 15 (1.5) & 8 (1.6) & 6 (2.3) & 43 (1.1) & 52 (1.3) \\
& Hyperosmolality and/or hypernatremia, n (\%) & 1651 (2.1) & 704 (2.1) & 344 (2.1) & 165 (2.0) & 86 (2.1) & 46 (2.2) & 30 (2.9) & 13 (2.5) & 5 (2.0) & 80 (2.1) & 73 (1.8) \\
& Fluid overload, n (\%) & 837 (1.1) & 348 (1.1) & 164 (1.0) & 82 (1.0) & 41 (1.0) & 22 (1.1) & 11 (1.1) & 5 (1.0) & 2 (0.8) & 45 (1.2) & 55 (1.3) \\
& Pancytopenia, n (\%) & 1000 (1.3) & 436 (1.3) & 219 (1.3) & 116 (1.4) & 56 (1.4) & 28 (1.4) & 15 (1.5) & 12 (2.3) & 9 (3.5) & 61 (1.6) & 48 (1.2) \\
& Other and unspecified coagulation defects, n (\%) & 2063 (2.6) & 927 (2.8) & 473 (2.9) & 226 (2.8) & 117 (2.9) & 61 (3.0) & 36 (3.5) & 18 (3.5) & 9 (3.5) & 114 (2.9) & 112 (2.7) \\
& Delirium due to conditions classified elsewhere, n (\%) & 1278 (1.6) & 507 (1.5) & 242 (1.5) & 114 (1.4) & 61 (1.5) & 36 (1.8) & 19 (1.9) & 11 (2.1) & 9 (3.5) & 62 (1.6) & 64 (1.6) \\
& Alcoholic liver damage, n (\%) & 1328 (1.7) & 568 (1.7) & 282 (1.7) & 145 (1.8) & 72 (1.8) & 38 (1.9) & 22 (2.1) & 13 (2.5) & 9 (3.5) & 84 (2.2) & 80 (1.9) \\
& Nonrheumatic aortic valve disorders, n (\%) & 1753 (2.2) & 718 (2.2) & 351 (2.1) & 181 (2.2) & 85 (2.1) & 47 (2.3) & 32 (3.1) & 22 (4.3) & 13 (5.1) & 95 (2.4) & 82 (2.0) \\
& Heart valve replaced, n (\%) & 1119 (1.4) & 492 (1.5) & 239 (1.5) & 134 (1.6) & 64 (1.6) & 38 (1.9) & 17 (1.7) & 10 (2.0) & 3 (1.2) & 53 (1.4) & 63 (1.5) \\
& Myocardial infarction, n (\%) & 8193 (10.4) & 3362 (10.3) & 1691 (10.3) & 845 (10.3) & 422 (10.3) & 227 (11.1) & 127 (12.4) & 70 (13.7) & 38 (14.8) & 399 (10.2) & 392 (9.5) \\
& Other chronic ischemic heart disease, unspecified, n (\%) & 18035 (23.0) & 7448 (22.7) & 3716 (22.7) & 1854 (22.6) & 924 (22.6) & 478 (23.3) & 252 (24.6) & 131 (25.6) & 71 (27.7) & 888 (22.8) & 871 (21.1) \\
& Chronic pulmonary heart disease, n (\%) & 3124 (4.0) & 1297 (4.0) & 674 (4.1) & 326 (4.0) & 163 (4.0) & 87 (4.2) & 43 (4.2) & 21 (4.1) & 14 (5.5) & 155 (4.0) & 140 (3.4) \\
& Primary/intrinsic cardiomyopathies, n (\%) & 1935 (2.5) & 857 (2.6) & 431 (2.6) & 222 (2.7) & 103 (2.5) & 61 (3.0) & 32 (3.1) & 16 (3.1) & 13 (5.1) & 96 (2.5) & 85 (2.1) \\
& Paroxysmal ventricular tachycardia, n (\%) & 1004 (1.3) & 407 (1.2) & 210 (1.3) & 95 (1.2) & 51 (1.2) & 29 (1.4) & 16 (1.6) & 9 (1.8) & 7 (2.7) & 49 (1.3) & 42 (1.0) \\
& Atrial fibrillation, n (\%) & 13326 (17.0) & 5506 (16.8) & 2709 (16.5) & 1337 (16.3) & 658 (16.1) & 337 (16.5) & 154 (15.0) & 83 (16.2) & 44 (17.2) & 634 (16.3) & 633 (15.3) \\
& Atrial flutter, n (\%) & 1660 (2.1) & 710 (2.2) & 357 (2.2) & 180 (2.2) & 85 (2.1) & 46 (2.2) & 27 (2.6) & 16 (3.1) & 9 (3.5) & 80 (2.1) & 83 (2.0) \\
& Congestive heart failure (chf) nos, n (\%) & 7720 (9.8) & 3215 (9.8) & 1645 (10.0) & 806 (9.8) & 402 (9.8) & 204 (10.0) & 97 (9.5) & 48 (9.4) & 29 (11.3) & 350 (9.0) & 389 (9.4) \\
& Heart failure with reduced ef [systolic or combined heart failure], n (\%) & 5719 (7.3) & 2403 (7.3) & 1194 (7.3) & 593 (7.2) & 288 (7.0) & 151 (7.4) & 74 (7.2) & 45 (8.8) & 31 (12.1) & 275 (7.1) & 253 (6.1) \\
& Heart failure with preserved ef [diastolic heart failure], n (\%) & 6814 (8.7) & 2845 (8.7) & 1449 (8.8) & 729 (8.9) & 360 (8.8) & 181 (8.8) & 95 (9.3) & 44 (8.6) & 17 (6.6) & 329 (8.4) & 320 (7.8) \\
& Pneumonitis due to inhalation of food or vomitus, n (\%) & 1357 (1.7) & 572 (1.7) & 291 (1.8) & 143 (1.7) & 61 (1.5) & 29 (1.4) & 17 (1.7) & 8 (1.6) & 6 (2.3) & 59 (1.5) & 59 (1.4) \\
& Pleurisy; pleural effusion, n (\%) & 2111 (2.7) & 896 (2.7) & 442 (2.7) & 206 (2.5) & 109 (2.7) & 55 (2.7) & 32 (3.1) & 19 (3.7) & 12 (4.7) & 109 (2.8) & 103 (2.5) \\
& Respiratory failure, n (\%) & 4518 (5.8) & 1946 (5.9) & 983 (6.0) & 476 (5.8) & 225 (5.5) & 116 (5.7) & 65 (6.3) & 34 (6.6) & 19 (7.4) & 183 (4.7) & 190 (4.6) \\
& Ascites (non malignant), n (\%) & 1137 (1.4) & 477 (1.5) & 230 (1.4) & 115 (1.4) & 59 (1.4) & 38 (1.9) & 20 (2.0) & 11 (2.1) & 8 (3.1) & 55 (1.4) & 54 (1.3) \\
& End stage renal disease, n (\%) & 3157 (4.0) & 1294 (3.9) & 628 (3.8) & 328 (4.0) & 172 (4.2) & 91 (4.4) & 49 (4.8) & 30 (5.9) & 18 (7.0) & 141 (3.6) & 159 (3.9) \\
& Sepsis, n (\%) & 4088 (5.2) & 1782 (5.4) & 895 (5.5) & 436 (5.3) & 231 (5.6) & 119 (5.8) & 60 (5.9) & 35 (6.8) & 18 (7.0) & 176 (4.5) & 237 (5.7) \\
\bottomrule
\end{tabular}
\end{adjustbox}
}
\end{table}


\begin{table}[tp]
\centering
\caption{\textbf{PTB-XL ECG dataset characteristics.}}
\label{tab:ecg_dataset_ptbxl}
{
\setlength{\tabcolsep}{4pt}
\begin{adjustbox}{max totalsize={\textwidth}{0.9\textheight}}
\begin{tabular}{lllllllllll}
\toprule
Split & & \multicolumn{7}{c}{Train (Nested Subsets)} & Val & Test \\
\midrule
& ECGs, n & 17418 & 8722 & 4356 & 2175 & 1091 & 547 & 273 & 2183 & 2198 \\
& Patients, n & 15023 & 7497 & 3727 & 1878 & 943 & 472 & 236 & 1942 & 1904 \\
\midrule
& \makecell[l]{Age,\\median [Q1, Q3]} & \makecell[l]{61.0\\{}[50.0, 72.0]} & \makecell[l]{61.0\\{}[50.0, 72.0]} & \makecell[l]{61.0\\{}[50.0, 71.0]} & \makecell[l]{61.0\\{}[51.0, 72.0]} & \makecell[l]{61.0\\{}[50.0, 73.0]} & \makecell[l]{61.0\\{}[51.5, 72.0]} & \makecell[l]{61.0\\{}[49.0, 73.0]} & \makecell[l]{63.0\\{}[50.0, 73.0]} & \makecell[l]{63.0\\{}[50.0, 75.0]} \\
\midrule
\multirow{2}{*}{Sex} & Female, n (\%) & 8329 (47.8) & 4177 (47.9) & 2084 (47.8) & 1039 (47.8) & 519 (47.6) & 261 (47.7) & 134 (49.1) & 1050 (48.1) & 1066 (48.5) \\
& Male, n (\%) & 9089 (52.2) & 4545 (52.1) & 2272 (52.2) & 1136 (52.2) & 572 (52.4) & 286 (52.3) & 139 (50.9) & 1133 (51.9) & 1132 (51.5) \\
\midrule
\multirow{71}{*}{Label} & NDT, n (\%) & 1461 (8.4) & 730 (8.4) & 365 (8.4) & 183 (8.4) & 99 (9.1) & 50 (9.1) & 28 (10.3) & 182 (8.3) & 182 (8.3) \\
& NST\_, n (\%) & 615 (3.5) & 308 (3.5) & 154 (3.5) & 77 (3.5) & 38 (3.5) & 20 (3.7) & 11 (4.0) & 75 (3.4) & 77 (3.5) \\
& DIG, n (\%) & 145 (0.8) & 73 (0.8) & 36 (0.8) & 18 (0.8) & 9 (0.8) & 4 (0.7) & 2 (0.7) & 18 (0.8) & 18 (0.8) \\
& LNGQT, n (\%) & 94 (0.5) & 47 (0.5) & 24 (0.6) & 12 (0.6) & 8 (0.7) & 5 (0.9) & 2 (0.7) & 12 (0.5) & 11 (0.5) \\
& NORM, n (\%) & 7596 (43.6) & 3825 (43.9) & 1907 (43.8) & 940 (43.2) & 477 (43.7) & 244 (44.6) & 128 (46.9) & 955 (43.7) & 963 (43.8) \\
& IMI, n (\%) & 2143 (12.3) & 1070 (12.3) & 535 (12.3) & 267 (12.3) & 129 (11.8) & 69 (12.6) & 34 (12.5) & 266 (12.2) & 267 (12.1) \\
& ASMI, n (\%) & 1887 (10.8) & 944 (10.8) & 472 (10.8) & 236 (10.9) & 116 (10.6) & 59 (10.8) & 30 (11.0) & 236 (10.8) & 234 (10.6) \\
& LVH, n (\%) & 1708 (9.8) & 854 (9.8) & 427 (9.8) & 213 (9.8) & 108 (9.9) & 53 (9.7) & 27 (9.9) & 210 (9.6) & 214 (9.7) \\
& LAFB, n (\%) & 1298 (7.5) & 649 (7.4) & 325 (7.5) & 162 (7.4) & 84 (7.7) & 43 (7.9) & 19 (7.0) & 163 (7.5) & 162 (7.4) \\
& ISC\_, n (\%) & 1019 (5.9) & 509 (5.8) & 255 (5.9) & 127 (5.8) & 64 (5.9) & 28 (5.1) & 14 (5.1) & 125 (5.7) & 128 (5.8) \\
& IRBBB, n (\%) & 894 (5.1) & 447 (5.1) & 224 (5.1) & 112 (5.1) & 52 (4.8) & 26 (4.8) & 13 (4.8) & 112 (5.1) & 112 (5.1) \\
& 1AVB, n (\%) & 634 (3.6) & 316 (3.6) & 157 (3.6) & 79 (3.6) & 42 (3.8) & 21 (3.8) & 11 (4.0) & 80 (3.7) & 79 (3.6) \\
& IVCD, n (\%) & 630 (3.6) & 314 (3.6) & 158 (3.6) & 79 (3.6) & 42 (3.8) & 21 (3.8) & 9 (3.3) & 78 (3.6) & 79 (3.6) \\
& ISCAL, n (\%) & 527 (3.0) & 264 (3.0) & 132 (3.0) & 66 (3.0) & 35 (3.2) & 19 (3.5) & 8 (2.9) & 66 (3.0) & 66 (3.0) \\
& CRBBB, n (\%) & 432 (2.5) & 217 (2.5) & 109 (2.5) & 54 (2.5) & 23 (2.1) & 13 (2.4) & 7 (2.6) & 55 (2.5) & 54 (2.5) \\
& CLBBB, n (\%) & 428 (2.5) & 214 (2.5) & 107 (2.5) & 53 (2.4) & 25 (2.3) & 14 (2.6) & 7 (2.6) & 54 (2.5) & 54 (2.5) \\
& ILMI, n (\%) & 383 (2.2) & 191 (2.2) & 95 (2.2) & 47 (2.2) & 23 (2.1) & 12 (2.2) & 6 (2.2) & 47 (2.2) & 48 (2.2) \\
& LAO/LAE, n (\%) & 341 (2.0) & 170 (1.9) & 85 (2.0) & 42 (1.9) & 21 (1.9) & 11 (2.0) & 6 (2.2) & 43 (2.0) & 42 (1.9) \\
& AMI, n (\%) & 282 (1.6) & 141 (1.6) & 70 (1.6) & 35 (1.6) & 18 (1.6) & 10 (1.8) & 5 (1.8) & 36 (1.6) & 35 (1.6) \\
& ALMI, n (\%) & 232 (1.3) & 116 (1.3) & 58 (1.3) & 29 (1.3) & 14 (1.3) & 8 (1.5) & 4 (1.5) & 29 (1.3) & 27 (1.2) \\
& ISCIN, n (\%) & 175 (1.0) & 87 (1.0) & 43 (1.0) & 21 (1.0) & 12 (1.1) & 6 (1.1) & 2 (0.7) & 21 (1.0) & 22 (1.0) \\
& INJAS, n (\%) & 171 (1.0) & 87 (1.0) & 43 (1.0) & 21 (1.0) & 11 (1.0) & 5 (0.9) & 2 (0.7) & 21 (1.0) & 22 (1.0) \\
& LMI, n (\%) & 161 (0.9) & 80 (0.9) & 40 (0.9) & 20 (0.9) & 10 (0.9) & 5 (0.9) & 2 (0.7) & 20 (0.9) & 20 (0.9) \\
& ISCIL, n (\%) & 143 (0.8) & 71 (0.8) & 35 (0.8) & 18 (0.8) & 10 (0.9) & 3 (0.5) & 1 (0.4) & 18 (0.8) & 18 (0.8) \\
& LPFB, n (\%) & 141 (0.8) & 70 (0.8) & 34 (0.8) & 17 (0.8) & 8 (0.7) & 4 (0.7) & 2 (0.7) & 18 (0.8) & 18 (0.8) \\
& ISCAS, n (\%) & 135 (0.8) & 68 (0.8) & 34 (0.8) & 17 (0.8) & 8 (0.7) & 4 (0.7) & 2 (0.7) & 17 (0.8) & 17 (0.8) \\
& INJAL, n (\%) & 116 (0.7) & 59 (0.7) & 30 (0.7) & 15 (0.7) & 8 (0.7) & 4 (0.7) & 2 (0.7) & 15 (0.7) & 14 (0.6) \\
& ISCLA, n (\%) & 113 (0.6) & 57 (0.7) & 28 (0.6) & 14 (0.6) & 7 (0.6) & 4 (0.7) & 2 (0.7) & 14 (0.6) & 13 (0.6) \\
& RVH, n (\%) & 102 (0.6) & 52 (0.6) & 26 (0.6) & 13 (0.6) & 5 (0.5) & 3 (0.5) & 2 (0.7) & 12 (0.5) & 12 (0.5) \\
& ANEUR, n (\%) & 84 (0.5) & 42 (0.5) & 21 (0.5) & 10 (0.5) & 5 (0.5) & 2 (0.4) & 1 (0.4) & 10 (0.5) & 10 (0.5) \\
& RAO/RAE, n (\%) & 79 (0.5) & 40 (0.5) & 20 (0.5) & 10 (0.5) & 4 (0.4) & 2 (0.4) & 1 (0.4) & 10 (0.5) & 10 (0.5) \\
& EL, n (\%) & 77 (0.4) & 39 (0.4) & 20 (0.5) & 10 (0.5) & 4 (0.4) & 2 (0.4) & 1 (0.4) & 10 (0.5) & 9 (0.4) \\
& WPW, n (\%) & 64 (0.4) & 32 (0.4) & 16 (0.4) & 8 (0.4) & 2 (0.2) & 1 (0.2) & 1 (0.4) & 7 (0.3) & 8 (0.4) \\
& ILBBB, n (\%) & 62 (0.4) & 31 (0.4) & 15 (0.3) & 7 (0.3) & 3 (0.3) & 2 (0.4) & 1 (0.4) & 7 (0.3) & 8 (0.4) \\
& IPLMI, n (\%) & 41 (0.2) & 20 (0.2) & 10 (0.2) & 5 (0.2) & 3 (0.3) & 2 (0.4) & 1 (0.4) & 5 (0.2) & 5 (0.2) \\
& ISCAN, n (\%) & 35 (0.2) & 17 (0.2) & 9 (0.2) & 5 (0.2) & 3 (0.3) & 1 (0.2) & 1 (0.4) & 5 (0.2) & 4 (0.2) \\
& IPMI, n (\%) & 26 (0.1) & 13 (0.1) & 7 (0.2) & 4 (0.2) & 3 (0.3) & 2 (0.4) & 2 (0.7) & 4 (0.2) & 3 (0.1) \\
& SEHYP, n (\%) & 24 (0.1) & 12 (0.1) & 6 (0.1) & 3 (0.1) & 2 (0.2) & 2 (0.4) & 1 (0.4) & 3 (0.1) & 2 (0.1) \\
& INJIN, n (\%) & 14 (0.1) & 7 (0.1) & 4 (0.1) & 2 (0.1) & 2 (0.2) & 1 (0.2) & 1 (0.4) & 2 (0.1) & 2 (0.1) \\
& INJLA, n (\%) & 13 (0.1) & 7 (0.1) & 3 (0.1) & 2 (0.1) & 2 (0.2) & 2 (0.4) & 1 (0.4) & 2 (0.1) & 2 (0.1) \\
& PMI, n (\%) & 13 (0.1) & 6 (0.1) & 3 (0.1) & 2 (0.1) & 1 (0.1) & 1 (0.2) & 1 (0.4) & 2 (0.1) & 2 (0.1) \\
& 3AVB, n (\%) & 12 (0.1) & 6 (0.1) & 2 (0.0) & 1 (0.0) & 1 (0.1) & 1 (0.2) & 1 (0.4) & 2 (0.1) & 2 (0.1) \\
& INJIL, n (\%) & 11 (0.1) & 6 (0.1) & 3 (0.1) & 2 (0.1) & 2 (0.2) & 1 (0.2) & 1 (0.4) & 2 (0.1) & 2 (0.1) \\
& 2AVB, n (\%) & 12 (0.1) & 6 (0.1) & 3 (0.1) & 2 (0.1) & 2 (0.2) & 2 (0.4) & 2 (0.7) & 1 (0.0) & 1 (0.0) \\
& ABQRS, n (\%) & 2683 (15.4) & 1341 (15.4) & 673 (15.4) & 335 (15.4) & 159 (14.6) & 85 (15.5) & 45 (16.5) & 322 (14.8) & 322 (14.6) \\
& PVC, n (\%) & 915 (5.3) & 456 (5.2) & 228 (5.2) & 114 (5.2) & 63 (5.8) & 29 (5.3) & 14 (5.1) & 114 (5.2) & 114 (5.2) \\
& STD\_, n (\%) & 807 (4.6) & 403 (4.6) & 201 (4.6) & 100 (4.6) & 49 (4.5) & 24 (4.4) & 12 (4.4) & 101 (4.6) & 101 (4.6) \\
& VCLVH, n (\%) & 701 (4.0) & 350 (4.0) & 176 (4.0) & 88 (4.0) & 43 (3.9) & 23 (4.2) & 10 (3.7) & 87 (4.0) & 87 (4.0) \\
& QWAVE, n (\%) & 438 (2.5) & 218 (2.5) & 109 (2.5) & 54 (2.5) & 27 (2.5) & 12 (2.2) & 6 (2.2) & 55 (2.5) & 55 (2.5) \\
& LOWT, n (\%) & 350 (2.0) & 175 (2.0) & 87 (2.0) & 43 (2.0) & 21 (1.9) & 11 (2.0) & 5 (1.8) & 44 (2.0) & 44 (2.0) \\
& NT\_, n (\%) & 340 (2.0) & 170 (1.9) & 85 (2.0) & 42 (1.9) & 21 (1.9) & 10 (1.8) & 5 (1.8) & 41 (1.9) & 42 (1.9) \\
& PAC, n (\%) & 318 (1.8) & 158 (1.8) & 79 (1.8) & 39 (1.8) & 21 (1.9) & 9 (1.6) & 4 (1.5) & 40 (1.8) & 40 (1.8) \\
& LPR, n (\%) & 272 (1.6) & 136 (1.6) & 68 (1.6) & 34 (1.6) & 16 (1.5) & 9 (1.6) & 4 (1.5) & 34 (1.6) & 34 (1.5) \\
& INVT, n (\%) & 235 (1.3) & 118 (1.4) & 59 (1.4) & 29 (1.3) & 14 (1.3) & 7 (1.3) & 4 (1.5) & 30 (1.4) & 29 (1.3) \\
& LVOLT, n (\%) & 145 (0.8) & 72 (0.8) & 36 (0.8) & 18 (0.8) & 10 (0.9) & 5 (0.9) & 2 (0.7) & 19 (0.9) & 18 (0.8) \\
& HVOLT, n (\%) & 49 (0.3) & 24 (0.3) & 12 (0.3) & 6 (0.3) & 3 (0.3) & 2 (0.4) & 1 (0.4) & 7 (0.3) & 6 (0.3) \\
& TAB\_, n (\%) & 28 (0.2) & 14 (0.2) & 7 (0.2) & 4 (0.2) & 2 (0.2) & 2 (0.4) & 1 (0.4) & 4 (0.2) & 3 (0.1) \\
& STE\_, n (\%) & 22 (0.1) & 11 (0.1) & 5 (0.1) & 3 (0.1) & 1 (0.1) & 1 (0.2) & 1 (0.4) & 3 (0.1) & 3 (0.1) \\
& PRC(S), n (\%) & 8 (0.0) & 4 (0.0) & 2 (0.0) & 1 (0.0) & 1 (0.1) & 1 (0.2) & 1 (0.4) & 1 (0.0) & 1 (0.0) \\
& SR, n (\%) & 13404 (77.0) & 6705 (76.9) & 3351 (76.9) & 1676 (77.1) & 842 (77.2) & 417 (76.2) & 210 (76.9) & 1670 (76.5) & 1674 (76.2) \\
& AFIB, n (\%) & 1211 (7.0) & 604 (6.9) & 302 (6.9) & 151 (6.9) & 71 (6.5) & 36 (6.6) & 20 (7.3) & 151 (6.9) & 152 (6.9) \\
& STACH, n (\%) & 661 (3.8) & 330 (3.8) & 165 (3.8) & 82 (3.8) & 40 (3.7) & 20 (3.7) & 10 (3.7) & 83 (3.8) & 82 (3.7) \\
& SARRH, n (\%) & 618 (3.5) & 308 (3.5) & 154 (3.5) & 77 (3.5) & 39 (3.6) & 21 (3.8) & 11 (4.0) & 77 (3.5) & 77 (3.5) \\
& SBRAD, n (\%) & 509 (2.9) & 254 (2.9) & 127 (2.9) & 63 (2.9) & 31 (2.8) & 16 (2.9) & 8 (2.9) & 64 (2.9) & 64 (2.9) \\
& PACE, n (\%) & 237 (1.4) & 118 (1.4) & 59 (1.4) & 29 (1.3) & 12 (1.1) & 5 (0.9) & 2 (0.7) & 29 (1.3) & 28 (1.3) \\
& SVARR, n (\%) & 128 (0.7) & 64 (0.7) & 32 (0.7) & 16 (0.7) & 9 (0.8) & 4 (0.7) & 2 (0.7) & 15 (0.7) & 14 (0.6) \\
& BIGU, n (\%) & 66 (0.4) & 33 (0.4) & 17 (0.4) & 9 (0.4) & 5 (0.5) & 2 (0.4) & 1 (0.4) & 8 (0.4) & 8 (0.4) \\
& AFLT, n (\%) & 59 (0.3) & 29 (0.3) & 14 (0.3) & 7 (0.3) & 4 (0.4) & 2 (0.4) & 1 (0.4) & 7 (0.3) & 7 (0.3) \\
& SVTAC, n (\%) & 21 (0.1) & 10 (0.1) & 5 (0.1) & 3 (0.1) & 2 (0.2) & 1 (0.2) & 1 (0.4) & 3 (0.1) & 3 (0.1) \\
& PSVT, n (\%) & 19 (0.1) & 10 (0.1) & 5 (0.1) & 3 (0.1) & 1 (0.1) & 1 (0.2) & 1 (0.4) & 3 (0.1) & 2 (0.1) \\
& TRIGU, n (\%) & 16 (0.1) & 8 (0.1) & 4 (0.1) & 2 (0.1) & 1 (0.1) & 1 (0.2) & 1 (0.4) & 2 (0.1) & 2 (0.1) \\
\bottomrule
\end{tabular}
\end{adjustbox}
}
\end{table}


\begin{table}[tp]
\centering
\caption{\textbf{CinC Georgia ECG dataset characteristics.}}
\label{tab:ecg_dataset_cinc}
{
\setlength{\tabcolsep}{4pt}
\begin{adjustbox}{max totalsize={\textwidth}{0.9\textheight}}
\begin{tabular}{llllllllll}
\toprule
Split & & \multicolumn{6}{c}{Train (Nested Subsets)} & Val & Test \\
\midrule
& ECGs, n & 8192 & 4096 & 2048 & 1024 & 512 & 256 & 1009 & 1008 \\
& Patients, n & 8192 & 4096 & 2048 & 1024 & 512 & 256 & 1009 & 1008 \\
\midrule
& Age, median [Q1, Q3] & 62.0 [51.0, 72.0] & 62.0 [51.0, 72.0] & 62.0 [50.0, 71.0] & 62.0 [50.0, 72.0] & 62.0 [50.8, 71.2] & 61.0 [50.0, 71.0] & 62.0 [50.0, 73.0] & 62.0 [51.0, 72.0] \\
\midrule
\multirow{2}{*}{Sex} & Female, n (\%) & 3791 (46.3) & 1893 (46.2) & 942 (46.0) & 471 (46.0) & 235 (45.9) & 117 (45.7) & 467 (46.3) & 466 (46.2) \\
& Male, n (\%) & 4401 (53.7) & 2203 (53.8) & 1106 (54.0) & 553 (54.0) & 277 (54.1) & 139 (54.3) & 542 (53.7) & 542 (53.8) \\
\midrule
\multirow{50}{*}{Label} & AF, n (\%) & 454 (5.5) & 227 (5.5) & 114 (5.6) & 57 (5.6) & 29 (5.7) & 15 (5.9) & 56 (5.6) & 56 (5.6) \\
& AFL, n (\%) & 149 (1.8) & 76 (1.9) & 38 (1.9) & 18 (1.8) & 9 (1.8) & 5 (2.0) & 18 (1.8) & 18 (1.8) \\
& AH, n (\%) & 44 (0.5) & 22 (0.5) & 10 (0.5) & 4 (0.4) & 2 (0.4) & 1 (0.4) & 5 (0.5) & 6 (0.6) \\
& AJR, n (\%) & 15 (0.2) & 9 (0.2) & 5 (0.2) & 3 (0.3) & 2 (0.4) & 1 (0.4) & 2 (0.2) & 2 (0.2) \\
& ALR, n (\%) & 10 (0.1) & 5 (0.1) & 2 (0.1) & 1 (0.1) & 1 (0.2) & 1 (0.4) & 1 (0.1) & 1 (0.1) \\
& AP, n (\%) & 42 (0.5) & 23 (0.6) & 12 (0.6) & 5 (0.5) & 2 (0.4) & 1 (0.4) & 5 (0.5) & 5 (0.5) \\
& ATach, n (\%) & 22 (0.3) & 11 (0.3) & 4 (0.2) & 2 (0.2) & 1 (0.2) & 1 (0.4) & 3 (0.3) & 3 (0.3) \\
& AVB, n (\%) & 60 (0.7) & 30 (0.7) & 14 (0.7) & 6 (0.6) & 3 (0.6) & 1 (0.4) & 7 (0.7) & 7 (0.7) \\
& AnMIs, n (\%) & 220 (2.7) & 109 (2.7) & 54 (2.6) & 27 (2.6) & 13 (2.5) & 6 (2.3) & 27 (2.7) & 27 (2.7) \\
& BBB, n (\%) & 92 (1.1) & 45 (1.1) & 23 (1.1) & 11 (1.1) & 5 (1.0) & 3 (1.2) & 11 (1.1) & 12 (1.2) \\
& CRBBB, n (\%) & 22 (0.3) & 12 (0.3) & 7 (0.3) & 3 (0.3) & 2 (0.4) & 1 (0.4) & 3 (0.3) & 3 (0.3) \\
& ERe, n (\%) & 112 (1.4) & 56 (1.4) & 29 (1.4) & 13 (1.3) & 7 (1.4) & 3 (1.2) & 14 (1.4) & 14 (1.4) \\
& IAVB, n (\%) & 613 (7.5) & 306 (7.5) & 152 (7.4) & 76 (7.4) & 38 (7.4) & 19 (7.4) & 76 (7.5) & 75 (7.4) \\
& IIAVB, n (\%) & 19 (0.2) & 9 (0.2) & 7 (0.3) & 4 (0.4) & 2 (0.4) & 1 (0.4) & 2 (0.2) & 2 (0.2) \\
& IIs, n (\%) & 357 (4.4) & 178 (4.3) & 88 (4.3) & 44 (4.3) & 22 (4.3) & 11 (4.3) & 44 (4.4) & 44 (4.4) \\
& ILBBB, n (\%) & 69 (0.8) & 35 (0.9) & 17 (0.8) & 8 (0.8) & 4 (0.8) & 2 (0.8) & 8 (0.8) & 9 (0.9) \\
& IRBBB, n (\%) & 321 (3.9) & 160 (3.9) & 80 (3.9) & 40 (3.9) & 20 (3.9) & 10 (3.9) & 39 (3.9) & 40 (4.0) \\
& LAA, n (\%) & 58 (0.7) & 29 (0.7) & 13 (0.6) & 7 (0.7) & 4 (0.8) & 2 (0.8) & 7 (0.7) & 7 (0.7) \\
& LAD, n (\%) & 750 (9.2) & 372 (9.1) & 185 (9.0) & 93 (9.1) & 47 (9.2) & 24 (9.4) & 92 (9.1) & 92 (9.1) \\
& LAE, n (\%) & 688 (8.4) & 340 (8.3) & 171 (8.3) & 85 (8.3) & 43 (8.4) & 22 (8.6) & 85 (8.4) & 85 (8.4) \\
& LAnFB, n (\%) & 142 (1.7) & 71 (1.7) & 34 (1.7) & 17 (1.7) & 8 (1.6) & 4 (1.6) & 17 (1.7) & 18 (1.8) \\
& LBBB, n (\%) & 185 (2.3) & 90 (2.2) & 44 (2.1) & 21 (2.1) & 10 (2.0) & 5 (2.0) & 23 (2.3) & 23 (2.3) \\
& LIs, n (\%) & 721 (8.8) & 359 (8.8) & 180 (8.8) & 90 (8.8) & 46 (9.0) & 23 (9.0) & 89 (8.8) & 89 (8.8) \\
& LPFB, n (\%) & 20 (0.2) & 10 (0.2) & 4 (0.2) & 2 (0.2) & 1 (0.2) & 1 (0.4) & 2 (0.2) & 3 (0.3) \\
& LQRSV, n (\%) & 295 (3.6) & 147 (3.6) & 73 (3.6) & 36 (3.5) & 18 (3.5) & 9 (3.5) & 37 (3.7) & 36 (3.6) \\
& LQT, n (\%) & 1083 (13.2) & 541 (13.2) & 271 (13.2) & 135 (13.2) & 68 (13.3) & 34 (13.3) & 148 (14.7) & 140 (13.9) \\
& LVH, n (\%) & 982 (12.0) & 489 (11.9) & 245 (12.0) & 122 (11.9) & 62 (12.1) & 31 (12.1) & 121 (12.0) & 121 (12.0) \\
& NSIVCB, n (\%) & 163 (2.0) & 81 (2.0) & 41 (2.0) & 21 (2.1) & 10 (2.0) & 5 (2.0) & 20 (2.0) & 20 (2.0) \\
& NSR, n (\%) & 1464 (17.9) & 723 (17.7) & 372 (18.2) & 194 (18.9) & 100 (19.5) & 50 (19.5) & 132 (13.1) & 118 (11.7) \\
& NSSTTA, n (\%) & 1486 (18.1) & 743 (18.1) & 371 (18.1) & 186 (18.2) & 94 (18.4) & 47 (18.4) & 183 (18.1) & 183 (18.2) \\
& PAC, n (\%) & 507 (6.2) & 253 (6.2) & 127 (6.2) & 63 (6.2) & 31 (6.1) & 16 (6.2) & 62 (6.1) & 63 (6.2) \\
& QAb, n (\%) & 364 (4.4) & 182 (4.4) & 90 (4.4) & 45 (4.4) & 22 (4.3) & 11 (4.3) & 45 (4.5) & 45 (4.5) \\
& RAAb, n (\%) & 11 (0.1) & 4 (0.1) & 2 (0.1) & 2 (0.2) & 1 (0.2) & 1 (0.4) & 2 (0.2) & 1 (0.1) \\
& RAD, n (\%) & 64 (0.8) & 32 (0.8) & 16 (0.8) & 8 (0.8) & 4 (0.8) & 2 (0.8) & 8 (0.8) & 8 (0.8) \\
& RAb, n (\%) & 8 (0.1) & 6 (0.1) & 5 (0.2) & 3 (0.3) & 2 (0.4) & 1 (0.4) & 1 (0.1) & 1 (0.1) \\
& RBBB, n (\%) & 430 (5.2) & 215 (5.2) & 108 (5.3) & 55 (5.4) & 27 (5.3) & 14 (5.5) & 53 (5.3) & 53 (5.3) \\
& RVH, n (\%) & 64 (0.8) & 32 (0.8) & 16 (0.8) & 8 (0.8) & 4 (0.8) & 2 (0.8) & 8 (0.8) & 8 (0.8) \\
& SA, n (\%) & 363 (4.4) & 182 (4.4) & 91 (4.4) & 46 (4.5) & 22 (4.3) & 11 (4.3) & 45 (4.5) & 44 (4.4) \\
& SB, n (\%) & 1334 (16.3) & 665 (16.2) & 334 (16.3) & 168 (16.4) & 84 (16.4) & 42 (16.4) & 164 (16.3) & 164 (16.3) \\
& STD, n (\%) & 30 (0.4) & 14 (0.3) & 8 (0.4) & 4 (0.4) & 2 (0.4) & 1 (0.4) & 4 (0.4) & 4 (0.4) \\
& STE, n (\%) & 108 (1.3) & 54 (1.3) & 27 (1.3) & 13 (1.3) & 6 (1.2) & 3 (1.2) & 13 (1.3) & 13 (1.3) \\
& STIAb, n (\%) & 788 (9.6) & 394 (9.6) & 196 (9.6) & 97 (9.5) & 49 (9.6) & 25 (9.8) & 97 (9.6) & 97 (9.6) \\
& STach, n (\%) & 1003 (12.2) & 500 (12.2) & 250 (12.2) & 125 (12.2) & 63 (12.3) & 32 (12.5) & 124 (12.3) & 123 (12.2) \\
& SVT, n (\%) & 26 (0.3) & 11 (0.3) & 5 (0.2) & 3 (0.3) & 2 (0.4) & 1 (0.4) & 3 (0.3) & 3 (0.3) \\
& TAb, n (\%) & 1832 (22.4) & 924 (22.6) & 459 (22.4) & 228 (22.3) & 115 (22.5) & 58 (22.7) & 226 (22.4) & 225 (22.3) \\
& TInv, n (\%) & 643 (7.8) & 320 (7.8) & 160 (7.8) & 80 (7.8) & 40 (7.8) & 20 (7.8) & 79 (7.8) & 79 (7.8) \\
& VEB, n (\%) & 33 (0.4) & 16 (0.4) & 8 (0.4) & 4 (0.4) & 2 (0.4) & 1 (0.4) & 4 (0.4) & 4 (0.4) \\
& VH, n (\%) & 55 (0.7) & 29 (0.7) & 14 (0.7) & 6 (0.6) & 3 (0.6) & 1 (0.4) & 7 (0.7) & 7 (0.7) \\
& VPB, n (\%) & 284 (3.5) & 142 (3.5) & 71 (3.5) & 36 (3.5) & 18 (3.5) & 9 (3.5) & 35 (3.5) & 35 (3.5) \\
& VPP, n (\%) & 35 (0.4) & 18 (0.4) & 9 (0.4) & 5 (0.5) & 3 (0.6) & 2 (0.8) & 5 (0.5) & 4 (0.4) \\
\bottomrule
\end{tabular}
\end{adjustbox}
}
\end{table}

\clearpage

\begin{table}[tp]
\centering
\caption{\textbf{EchoNext ECG dataset characteristics.}}
\label{tab:ecg_dataset_echonext}
{
\setlength{\tabcolsep}{4pt}
\begin{adjustbox}{max totalsize={\textwidth}{0.9\textheight}}
\begin{tabular}{lllllllllllll}
\toprule
Split & & \multicolumn{9}{c}{Train (Nested Subsets)} & Val & Test \\
\midrule
& ECGs, n & 72475 & 32768 & 16384 & 8192 & 4096 & 2048 & 1024 & 512 & 256 & 4626 & 5442 \\
& Patients, n & 26218 & 16370 & 10282 & 6072 & 3405 & 1846 & 965 & 496 & 255 & 4626 & 5442 \\
\midrule
& \makecell[l]{Age,\\median [Q1, Q3]} & \makecell[l]{63.0\\{}[52.0, 73.0]} & \makecell[l]{63.0\\{}[52.0, 73.0]} & \makecell[l]{63.0\\{}[51.0, 73.0]} & \makecell[l]{63.0\\{}[51.0, 73.0]} & \makecell[l]{63.0\\{}[51.0, 73.0]} & \makecell[l]{63.0\\{}[51.0, 73.0]} & \makecell[l]{63.0\\{}[51.0, 73.0]} & \makecell[l]{63.0\\{}[51.0, 73.0]} & \makecell[l]{63.0\\{}[51.0, 73.0]} & \makecell[l]{64.0\\{}[52.0, 75.0]} & \makecell[l]{64.0\\{}[52.0, 74.0]} \\
\midrule
\multirow{2}{*}{Sex} & Female, n (\%) & 33524 (46.3) & 15157 (46.3) & 7580 (46.3) & 3790 (46.3) & 1895 (46.3) & 949 (46.3) & 475 (46.4) & 236 (46.1) & 120 (46.9) & 2356 (50.9) & 2731 (50.2) \\
& Male, n (\%) & 38951 (53.7) & 17611 (53.7) & 8804 (53.7) & 4402 (53.7) & 2201 (53.7) & 1099 (53.7) & 549 (53.6) & 276 (53.9) & 136 (53.1) & 2270 (49.1) & 2711 (49.8) \\
\midrule
\multirow{12}{*}{Label} & LVEF Lo, n (\%) & 16962 (23.4) & 7690 (23.5) & 3845 (23.5) & 1908 (23.3) & 953 (23.3) & 481 (23.5) & 243 (23.7) & 121 (23.6) & 64 (25.0) & 866 (18.7) & 962 (17.7) \\
& LVWT Hi, n (\%) & 17667 (24.4) & 7975 (24.3) & 4003 (24.4) & 1989 (24.3) & 1001 (24.4) & 503 (24.6) & 243 (23.7) & 118 (23.0) & 66 (25.8) & 877 (19.0) & 1061 (19.5) \\
& AS, n (\%) & 2919 (4.0) & 1318 (4.0) & 642 (3.9) & 315 (3.8) & 155 (3.8) & 83 (4.1) & 46 (4.5) & 20 (3.9) & 10 (3.9) & 252 (5.4) & 286 (5.3) \\
& AR, n (\%) & 878 (1.2) & 405 (1.2) & 208 (1.3) & 104 (1.3) & 43 (1.0) & 20 (1.0) & 11 (1.1) & 6 (1.2) & 3 (1.2) & 62 (1.3) & 66 (1.2) \\
& MR, n (\%) & 6137 (8.5) & 2760 (8.4) & 1379 (8.4) & 687 (8.4) & 332 (8.1) & 156 (7.6) & 75 (7.3) & 43 (8.4) & 16 (6.2) & 282 (6.1) & 337 (6.2) \\
& TR, n (\%) & 7707 (10.6) & 3460 (10.6) & 1772 (10.8) & 859 (10.5) & 453 (11.1) & 213 (10.4) & 93 (9.1) & 41 (8.0) & 23 (9.0) & 305 (6.6) & 353 (6.5) \\
& PR, n (\%) & 603 (0.8) & 273 (0.8) & 132 (0.8) & 64 (0.8) & 36 (0.9) & 13 (0.6) & 7 (0.7) & 3 (0.6) & 2 (0.8) & 21 (0.5) & 20 (0.4) \\
& RVD, n (\%) & 9597 (13.2) & 4349 (13.3) & 2178 (13.3) & 1075 (13.1) & 535 (13.1) & 260 (12.7) & 126 (12.3) & 52 (10.2) & 29 (11.3) & 368 (8.0) & 419 (7.7) \\
& PEff, n (\%) & 2079 (2.9) & 972 (3.0) & 490 (3.0) & 254 (3.1) & 134 (3.3) & 68 (3.3) & 40 (3.9) & 19 (3.7) & 11 (4.3) & 52 (1.1) & 69 (1.3) \\
& PASP Hi, n (\%) & 13727 (18.9) & 6185 (18.9) & 3068 (18.7) & 1536 (18.8) & 757 (18.5) & 360 (17.6) & 175 (17.1) & 85 (16.6) & 38 (14.8) & 581 (12.6) & 699 (12.8) \\
& TRV Hi, n (\%) & 7492 (10.3) & 3393 (10.4) & 1699 (10.4) & 834 (10.2) & 410 (10.0) & 198 (9.7) & 82 (8.0) & 37 (7.2) & 16 (6.2) & 267 (5.8) & 375 (6.9) \\
& SHD, n (\%) & 37958 (52.4) & 17163 (52.4) & 8582 (52.4) & 4291 (52.4) & 2146 (52.4) & 1072 (52.3) & 536 (52.3) & 268 (52.3) & 134 (52.3) & 1990 (43.0) & 2318 (42.6) \\
\bottomrule
\end{tabular}
\end{adjustbox}
}
\end{table}


\begin{table}[tp]
\centering
\caption{\textbf{ZZU pECG dataset characteristics.}}
\label{tab:ecg_dataset_zzu}
{
\setlength{\tabcolsep}{4pt}
\begin{adjustbox}{max totalsize={\textwidth}{0.9\textheight}}
\begin{tabular}{llllllllll}
\toprule
Split & & \multicolumn{6}{c}{Train (Nested Subsets)} & Val & Test \\
\midrule
& ECGs, n & 8658 & 4096 & 2048 & 1024 & 512 & 256 & 1033 & 2079 \\
& Patients, n & 7243 & 3694 & 1921 & 994 & 505 & 254 & 1033 & 2079 \\
\midrule
& \makecell[l]{Age (in days),\\median [Q1, Q3]} & \makecell[l]{3540.0\\{}[2145.2, 4696.8]} & \makecell[l]{3537.0\\{}[2140.5, 4694.2]} & \makecell[l]{3535.0\\{}[2138.2, 4684.2]} & \makecell[l]{3534.0\\{}[2133.5, 4673.2]} & \makecell[l]{3523.5\\{}[2131.2, 4667.8]} & \makecell[l]{3523.5\\{}[2124.2, 4660.5]} & \makecell[l]{3618.0\\{}[2273.0, 4734.0]} & \makecell[l]{3354.0\\{}[2109.5, 4542.5]} \\
\midrule
\multirow{2}{*}{Sex} & Female, n (\%) & 3645 (42.1) & 1725 (42.1) & 863 (42.1) & 431 (42.1) & 215 (42.0) & 108 (42.2) & 453 (43.9) & 840 (40.4) \\
& Male, n (\%) & 5013 (57.9) & 2371 (57.9) & 1185 (57.9) & 593 (57.9) & 297 (58.0) & 148 (57.8) & 580 (56.1) & 1239 (59.6) \\
\midrule
\multirow{4}{*}{Label} & Myocarditis, n (\%) & 412 (4.8) & 197 (4.8) & 101 (4.9) & 47 (4.6) & 20 (3.9) & 14 (5.5) & 30 (2.9) & 61 (2.9) \\
& Cardiomyopathy, n (\%) & 91 (1.1) & 41 (1.0) & 23 (1.1) & 16 (1.6) & 10 (2.0) & 7 (2.7) & 7 (0.7) & 17 (0.8) \\
& Kawasaki disease, n (\%) & 98 (1.1) & 42 (1.0) & 22 (1.1) & 8 (0.8) & 6 (1.2) & 4 (1.6) & 12 (1.2) & 25 (1.2) \\
& Congenital heart disease, n (\%) & 1353 (15.6) & 635 (15.5) & 350 (17.1) & 172 (16.8) & 92 (18.0) & 45 (17.6) & 152 (14.7) & 314 (15.1) \\
\bottomrule
\end{tabular}
\end{adjustbox}
}
\end{table}


\begin{table}[tp]
\centering
\caption{\textbf{CODE-15\% ECG dataset characteristics.}}
\label{tab:ecg_dataset_code15}
{
\setlength{\tabcolsep}{4pt}
\begin{adjustbox}{max totalsize={\textwidth}{0.9\textheight}}
\begin{tabular}{lllllllllllll}
\toprule
Split & & \multicolumn{9}{c}{Train (Nested Subsets)} & Val & Test \\
\midrule
& ECGs, n & 74112 & 32768 & 16384 & 8192 & 4096 & 2048 & 1024 & 512 & 256 & 24704 & 24705 \\
& Patients, n & 74112 & 32768 & 16384 & 8192 & 4096 & 2048 & 1024 & 512 & 256 & 24704 & 24705 \\
\midrule
& \makecell[l]{Age,\\median [Q1, Q3]} & \makecell[l]{52.0\\{}[35.0, 69.0]} & \makecell[l]{52.0\\{}[35.0, 69.0]} & \makecell[l]{53.0\\{}[35.0, 69.0]} & \makecell[l]{52.0\\{}[35.0, 70.0]} & \makecell[l]{53.0\\{}[35.0, 70.0]} & \makecell[l]{52.0\\{}[35.0, 70.0]} & \makecell[l]{52.0\\{}[35.0, 70.0]} & \makecell[l]{52.0\\{}[35.0, 70.0]} & \makecell[l]{53.0\\{}[35.0, 72.0]} & \makecell[l]{52.0\\{}[35.0, 69.0]} & \makecell[l]{53.0\\{}[35.0, 69.0]} \\
\midrule
\multirow{2}{*}{Sex} & Female, n (\%) & 44167 (59.6) & 19527 (59.6) & 9762 (59.6) & 4880 (59.6) & 2439 (59.5) & 1218 (59.5) & 608 (59.4) & 303 (59.2) & 150 (58.6) & 14723 (59.6) & 14723 (59.6) \\
& Male, n (\%) & 29945 (40.4) & 13241 (40.4) & 6622 (40.4) & 3312 (40.4) & 1657 (40.5) & 830 (40.5) & 416 (40.6) & 209 (40.8) & 106 (41.4) & 9981 (40.4) & 9982 (40.4) \\
\midrule
\multirow{7}{*}{Label} & 1dAVb, n (\%) & 1231 (1.7) & 537 (1.6) & 275 (1.7) & 144 (1.8) & 70 (1.7) & 33 (1.6) & 19 (1.9) & 13 (2.5) & 9 (3.5) & 421 (1.7) & 379 (1.5) \\
& RBBB, n (\%) & 2227 (3.0) & 976 (3.0) & 504 (3.1) & 260 (3.2) & 114 (2.8) & 62 (3.0) & 32 (3.1) & 14 (2.7) & 8 (3.1) & 734 (3.0) & 749 (3.0) \\
& LBBB, n (\%) & 1206 (1.6) & 552 (1.7) & 278 (1.7) & 143 (1.7) & 78 (1.9) & 42 (2.1) & 24 (2.3) & 9 (1.8) & 6 (2.3) & 417 (1.7) & 409 (1.7) \\
& SB, n (\%) & 1321 (1.8) & 604 (1.8) & 305 (1.9) & 135 (1.6) & 67 (1.6) & 35 (1.7) & 14 (1.4) & 5 (1.0) & 3 (1.2) & 401 (1.6) & 388 (1.6) \\
& ST, n (\%) & 1709 (2.3) & 740 (2.3) & 388 (2.4) & 198 (2.4) & 91 (2.2) & 50 (2.4) & 22 (2.1) & 9 (1.8) & 7 (2.7) & 541 (2.2) & 531 (2.1) \\
& AF, n (\%) & 1361 (1.8) & 590 (1.8) & 297 (1.8) & 150 (1.8) & 83 (2.0) & 45 (2.2) & 23 (2.2) & 13 (2.5) & 3 (1.2) & 458 (1.9) & 463 (1.9) \\
& mortality, n (\%) & 3989 (5.4) & 1782 (5.4) & 880 (5.4) & 442 (5.4) & 223 (5.4) & 109 (5.3) & 56 (5.5) & 28 (5.5) & 17 (6.6) & 1331 (5.4) & 1329 (5.4) \\
\bottomrule
\end{tabular}
\end{adjustbox}
}
\end{table}

\end{document}